\documentclass{article} 
\usepackage{iclr2027_conference,times}

\usepackage{amsmath,amsfonts,bm}

\def\eqref#1{equation~\ref{#1}}

\def\1{\bm{1}}

\DeclareMathAlphabet{\mathsfit}{\encodingdefault}{\sfdefault}{m}{sl}
\SetMathAlphabet{\mathsfit}{bold}{\encodingdefault}{\sfdefault}{bx}{n}

\iclrfinalcopy

\usepackage{hyperref}
\usepackage{url}

\usepackage{graphicx}

\usepackage{graphicx}
\usepackage{subcaption}

\usepackage{amssymb} 
\usepackage{amsmath}
\usepackage{amsmath}
\usepackage{enumitem}

\usepackage{wrapfig}
\usepackage{booktabs}

\usepackage{caption}

\usepackage{algorithm}
\usepackage{algorithmic}
\usepackage{tabularx}

\usepackage[most]{tcolorbox}

\newtcolorbox{datasetbox}[1]{
    enhanced,
    breakable,
    colback=black!2,
    colframe=black!25,
    colbacktitle=black!7,
    coltitle=black,
    title=#1,
    fonttitle=\bfseries\small,
    fontupper=\small,
    boxrule=0.5pt,
    arc=1mm,
    left=2mm,
    right=2mm,
    top=1.5mm,
    bottom=1.5mm
}

\title{From Geographic Memory to Spatial 

Generalization in Large Language Models}

\author{
Chen Chu\textsuperscript{1},
Zhaonan Wang\textsuperscript{2,*},
Cyrus Shahabi\textsuperscript{1,*},
Hengcai Zhang\textsuperscript{3},
Feng Lu\textsuperscript{3} \\
\textsuperscript{1}University of Southern California \\
\textsuperscript{2}NYU Shanghai \\
\textsuperscript{3}University of Chinese Academy of Sciences \\
\texttt{\{chenchu,shahabi\}@usc.edu} \\
\texttt{zhaonan.wang@nyu.edu} \\
\textsuperscript{*}Corresponding author
}

\begin{document}

\maketitle

\fancyhead{}

\begin{abstract}
Large language models (LLMs) can associate places with geographic descriptions, but recovering an observed association does not establish an ability to generalize across geographic space.
We distinguish geospatial memory, the recovery of location--description associations observed during training, from spatial generalization, the prediction of descriptions for locations in unobserved regions.
We investigate whether continuous location representations improve this generalization and encourage geographic knowledge to become spatially organized within an LLM.
We frame this organization as a latent spatial field: a learned mapping from locations to descriptions connected through spatial dependencies.
Building on a continuous spatial encoder, we project location representations into the model's embedding space and supply them as spatial tokens.
We evaluate the resulting models on two newly curated geo-text benchmarks for Los Angeles and London using two complementary assessments.
Geospatial captioning tests both memory and spatial generalization by generating descriptions at observed locations and at locations beyond the spatial coverage of the training observations.
A separate geospatial grounding model tests whether generated location--description pairs provide useful additional supervision for mapping text to geographic space.
Across multiple LLM families, continuous location representations improve captioning at unobserved locations relative to textual coordinates, while their generated descriptions yield modest but consistent improvements in downstream grounding.
Layer-wise probing and representation clustering provide additional evidence consistent with spatially organized geographic representations.
Together, these findings show the value of evaluating spatial generalization separately from memory and of incorporating continuous spatial representations into language models. Data and codes can be found at: \url{https://github.com/chuchen2017/GeospatialReasoning}
\end{abstract}

\section{Introduction}
\label{sec:introduction}

Large language models (LLMs) can answer questions about geographic places, but correct answers alone do not reveal how their geographic knowledge is organized~\citep{geollama,bias,citybench}.
Consider a model trained on descriptions of several locations near a beach.
At a training location, generating ``a short walk from the beach'' may simply recover an observed location--description association.
Generating an appropriate description for a location in a nearby region not covered during training tests whether the model can generalize beyond those associations.
This distinction motivates our central question: \emph{Can LLMs organize geographic knowledge spatially enough to generalize to locations without direct training observations, and how does the representation of location affect this ability?}

This question matters because geographic observations are sparse and unevenly distributed samples of continuous space~\citep{ros,socio-semantic}.
Well-documented locations provide abundant information, while other areas have limited coverage~\citep{citgpt}.
Geographic knowledge nevertheless contains dependencies that can support inference across these gaps: nearby locations often share environmental characteristics, and descriptions of places depend on their relationships to landmarks, transportation networks, and surrounding areas~\citep{song2026}.
We distinguish two capabilities.
\textbf{Geospatial memory} is the recovery of location--description associations observed during training.
\textbf{Geospatial reasoning} is spatial generalization to regions beyond the spatial coverage of the training observations.
This second setting tests whether learned geographic relationships support predictions beyond the observed pairs.

How locations are represented may influence both capabilities.
Place names and coordinates represented as character strings fit naturally into an LLM's text interface, but their token representations do not explicitly encode geographic proximity~\citep{slm}.
Existing location encoders~\citep{space2vec,geo2vec} provide continuous representations designed to introduce spatial structure.
These methods provide the foundation for our approach.
We investigate whether incorporating such representations into an LLM improves its ability to learn geographic dependencies, generate in unobserved regions, and organize geographic information within its hidden states.

We develop a \textbf{spatially explicit LLM} by projecting a continuous location encoding into the model's input embedding space and supplying it as a spatial token.
The model learns to generate geographic descriptions conditioned on this representation.
We interpret the resulting location-to-description mapping as a \textbf{latent spatial field}: geographic knowledge associated with different locations is connected through learned spatial dependencies.
This perspective provides a hypothesis about how geographic knowledge can be organized, without requiring every geographic property to vary smoothly.
Our experiments examine whether continuous location encodings encourage this organization more effectively than textual coordinates, discrete identifiers and hierarchical spatial indexes.

We evaluate the approach using coordinate--description datasets constructed from Airbnb listings in Los Angeles and London. \textbf{Geospatial captioning} asks the model to generate a geographic description from a location.
Captioning at observed training locations evaluates memory; captioning in withheld spatial regions evaluates reasoning under our definition.
Using the same task for both settings allows us to distinguish the ability to recover observed associations from the ability to generalize across gaps in the training data.
Comparisons with nearest-neighbor retrieval further assess whether generated descriptions provide benefits beyond reusing nearby observations.

We then test whether these descriptions contain useful geographic information through \textbf{geospatial grounding}, a separate task that maps a description to a distribution over possible locations.
We use the captioning model to generate additional location--description pairs in data-sparse areas and evaluate whether those pairs improve an independently trained grounding model.
Grounding therefore provides a downstream utility test for the generated knowledge, complementing direct comparisons between generated captions and withheld reference descriptions.
Finally, layer-wise probing and clustering examine how geographic information is represented within the captioning model.

Across the tested LLM families and both cities, textual coordinates support strong recovery of observed associations, while continuous location representations improve captioning in withheld regions.
Generated descriptions can also provide modest improvements in downstream grounding.
Probing and clustering reveal patterns consistent with more spatially organized hidden representations.
Together, these findings support the value of continuous location encodings for geographic generalization, while providing complementary behavioral and representational evidence for the latent-spatial-field perspective. Our contributions are threefold:
\begin{itemize}[leftmargin=*]
    \item \textbf{A framework for evaluating geographic generalization in LLMs.}
    We distinguish recovery of observed associations from prediction in spatially uncovered regions, using captioning to evaluate both and independent grounding to assess downstream utility.
    \item \textbf{An integration of continuous location representations into LLMs.}
    We systematically compare continuous encodings with textual, grid-based, and hierarchical representations.
    \item \textbf{An empirical investigation of geographic knowledge organization.}
    Experiments on Los Angeles and London connect location representation, generalization performance, downstream usefulness, and the spatial structure of hidden representations.
\end{itemize}

\section{Related Work}

\paragraph{Geospatial Reasoning in LLMs} 
Spatial reasoning is typically formulated as inference from explicitly provided relations, such as deriving the relation between $A$ and $C$ from `$A$ is north of $B$'' and `$B$ is north of $C$''~\citep{symbolic,spatialrgpt}, or decomposed into atomic functions~\citep{slm,spatialagent}. Real-world geographic dependencies, however, are too complex to enumerate symbolically~\citep{gwe}, while supplying all relevant observations is impractical. We study geospatial reasoning as inference from knowledge internalized by an LLM.

\paragraph{Geospatial Knowledge in LLMs}

Prior work shows that LLMs encode geographic knowledge useful for downstream prediction tasks~\citep{citybench,geollm}, while \cite{spacetime} reveal decodable location representations across multiple spatial scales. These studies primarily examine entities and places assumed to be represented in the training data. We instead investigate whether learned geographic knowledge reflects surface-level text--location associations encountered during training or a coherent internal spatial organization that supports generalization to unobserved locations.

\paragraph{Geospatial Representation}
Geospatial representation learning maps locations or entities into a latent space while preserving properties such as distance, direction, periodicity, and multi-scale structure~\citep{geo2vec,poly2vec}. It introduces inductive bias of location into various downstream models~\citep{satclip,csp,space2vec}. We integrate continuous location encodings directly into an LLM and investigate whether they encourage geographic knowledge to form a coherent latent spatial field that supports both memory and reasoning.

\section{Geospatial Memory and Reasoning over Latent Spatial Field}
\label{sec:formulation}

Given the largely persistent and continuous nature of geospatial knowledge, we propose to view its internal organization within a language model as a latent spatial field. Under this view, geographic knowledge is not merely stored as a collection of independent location--value associations. Instead, these associations form a structured field in which the knowledge associated with different locations is connected through implicit spatial dependencies.

\subsection{Geospatial Knowledge as a Latent Spatial Field}
\label{sec:spatial_field}

Let $\Omega \subseteq \mathbb{R}^{2}$ denote a continuous geographic domain, and let $\mathcal{V}$ denote a general space of geographic values. A value $v \in \mathcal{V}$ may represent a natural-language description, a land-use category, or any other information associated with a location. We represent the observed geospatial data as $\mathcal{D}=\bigl\{(\mathbf{x}_{i},v_{i})\bigr\}_{i=1}^{N}$, where $\mathbf{x}_{i} \in \Omega$ is a geographic location and $v_{i} \in \mathcal{V}$ is its observed value. These observations can be viewed as sparse samples from an underlying spatial field $F:\Omega\rightarrow\mathcal{V}$, where $F(\mathbf{x})$ represents the geographic knowledge associated with location $\mathbf{x}$. In practice, these observations usually are not collected uniformly. We express the data-generating process as
\begin{equation}
    (\mathbf{x},v)
    \sim
    \pi_{\mathrm{obs}}(\mathbf{x})
    p
    \bigl(
        v \mid \mathbf{x}
    \bigr),
\end{equation}
where $\pi_{\mathrm{obs}}(\mathbf{x})$ denotes the spatial observation density and $p(v \mid \mathbf{x})$ describes the distribution of geographic values at location $\mathbf{x}$. Geographic values are also not independent across locations. Consequently, inferring $F(\mathbf{x})$ at an unobserved location requires more than applying an explicitly given symbolic rule: the model must first internalize the latent dependencies $F$ between space and geographic values from the observed data.

\subsection{Memory and Reasoning}
\label{sec:memory_reasoning}

Given the training observations in $\mathcal{D}_{\mathrm{train}}$, a model approximates the underlying field $F$ by learning
\(
\theta^{*}
=
\underset{\theta}{\arg\min}
\sum_{(\mathbf{x}_{i},v_{i})\in\mathcal{D}_{\mathrm{train}}}
\mathcal{L}\bigl(F_{\theta}(\mathbf{x}_{i}),v_{i}\bigr),
\)
where $\mathcal{L}$ denotes the training loss. The learned field is $F_{\theta^{*}}$, and inference at location $\mathbf{x}$ is given by $\hat{v}=F_{\theta^{*}}(\mathbf{x})$. We distinguish geospatial memory from geospatial reasoning according to whether the target location--value association was observed during training. Let $\mathcal{X}_{\mathrm{obs}}=\bigl\{\mathbf{x}_{i}:(\mathbf{x}_{i},v_{i})\in\mathcal{D}_{\mathrm{train}}\bigr\}$ denote the set of observed training locations. We define \textbf{geospatial memory} as the ability to retain values associated with locations in $\mathcal{X}_{\mathrm{obs}}$:
\begin{equation}
    F_{\theta^{*}}
    \bigl(
        \mathbf{x}_{i}
    \bigr)
    \approx
    v_{i},
    \qquad
    \mathbf{x}_{i}
    \in
    \mathcal{X}_{\mathrm{obs}}.
\end{equation}
Geospatial memory therefore measures whether the model has internalized the observed geographic knowledge so that it can be directly retrieved from its parameters.

To define geospatial reasoning, let $\mathcal{H} \subseteq \Omega$ denote a collection of regions, not covered by the training observations. We define the corresponding dataset as $\mathcal{D}_{\mathrm{h}}=\bigl\{(\mathbf{x}_{i},v_{i})\in\mathcal{D}:\mathbf{x}_{i}\in\mathcal{H}\bigr\}$, where $\mathcal{H}\cap\mathcal{X}_{\mathrm{obs}}=\varnothing$. The values in $\mathcal{D}_{\mathrm{h}}$ remain unavailable during training. We define \textbf{geospatial reasoning} as the ability to infer the geographic value $v_{*}$ at an unobserved location $\mathbf{x}_{*} \in \mathcal{H}$:
\begin{equation}
    \hat{v}_{*}
    =
    F_{\theta^{*}}
    \bigl(
        \mathbf{x}_{*}
    \bigr)
    \approx
    v_{*},
    \qquad
    \mathbf{x}_{*}
    \in
    \mathcal{H}.
\end{equation}

Equivalently, geospatial reasoning estimates the posterior predictive distribution $p\bigl(v_{*}\mid\mathbf{x}_{*},\mathcal{D}_{\mathrm{train}}\bigr)$ for a location whose value was not observed. Although $\mathcal{D}_{\mathrm{train}}$ is not explicitly supplied as context at inference time, its spatial dependencies have been internalized in the learned parameters $\theta^{*}$. The model must use this learned structure to bridge the missing portion of the geographic field.

\section{Intrinsic and Extrinsic Evaluation of Memory and Reasoning}
\label{sec:evaluation}

We employ two complementary evaluation tasks. Geospatial captioning directly evaluates whether a model can recover geographic knowledge at observed locations and reason about unobserved ones. Geospatial grounding, implemented using a separate model, provides an extrinsic evaluation of reasoning by testing whether knowledge inferred at unobserved locations benefits a downstream task.

\begin{figure*}[h]
    \centering
    \includegraphics[width=1\textwidth]{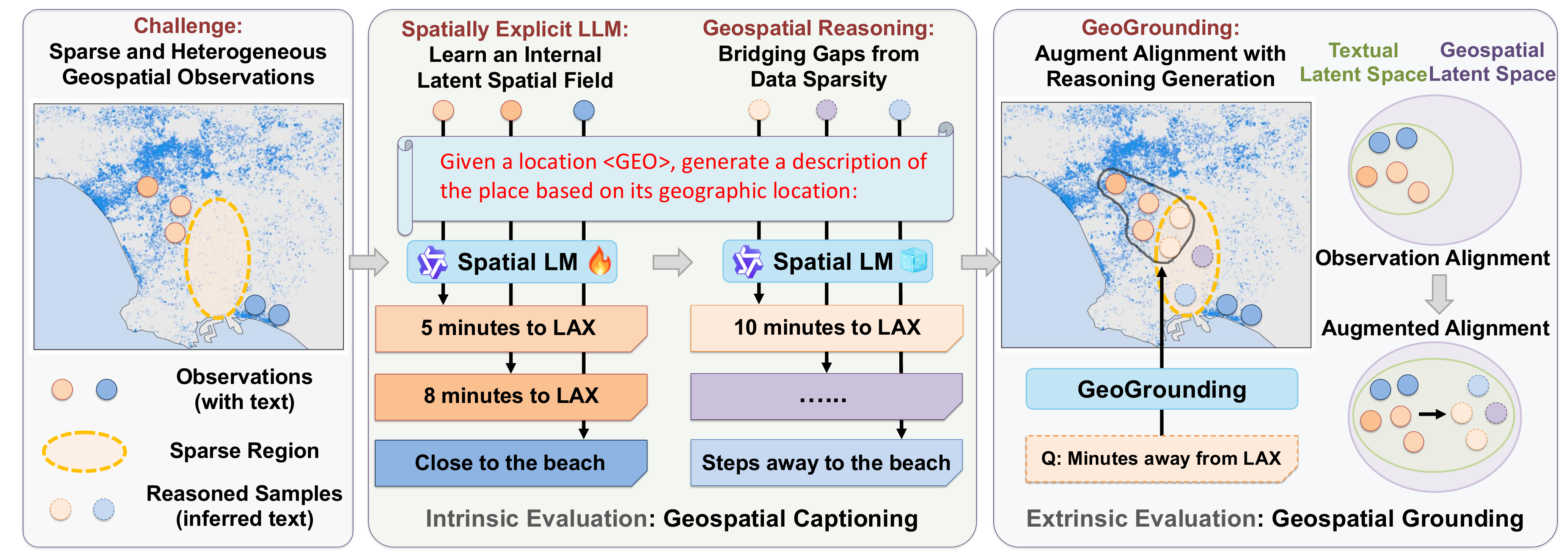} 
    \caption{
    \textbf{Overview of geospatial captioning and grounding.}
    Geo-text observations are spatially heterogeneous. A spatially explicit LLM organizes geographic knowledge as a latent spatial field that supports generalization across space. Through \textbf{geospatial captioning}, it generates geo-text pairs that augment the observed data and improve text--space alignment in \textbf{geospatial grounding}.
    }
    \label{fig:overview}
\end{figure*}

\subsection{Geospatial Captioning}
\label{sec:captioning_formulation}

As shown in Figure~\ref{fig:overview}, geospatial captioning instantiates our general formulation by defining the geographic value space $\mathcal{V}$ as the space of natural-language descriptions of locations $\mathcal{Y}$. A captioning model learns the conditional mapping $c_{\psi}:\Omega\rightarrow\mathcal{P}(\mathcal{Y})$, where $\mathcal{P}(\mathcal{Y})$ denotes the space of probability distributions over possible descriptions. Given a location $\mathbf{x}$, the model generates
\begin{equation}
    \hat{y}
    \sim
    p_{\psi}
    \bigl(
        y \mid \mathbf{x}
    \bigr).
\end{equation}

At an observed location $\mathbf{x}\in\mathcal{X}_{\mathrm{obs}}$, the generated caption $\hat{y}$ evaluates geospatial memory. At an unobserved location $\mathbf{x}\in\mathcal{H}$, it evaluates geospatial reasoning by testing whether the model can infer an unobserved geo-text association. Because multiple descriptions may be valid for the same location, the intrinsic evaluation measures the semantic and lexical alignment between the generated description $\hat{y}$ and the reference description $y$.

\subsection{Extrinsic Evaluation through Geospatial Grounding}
\label{sec:grounding_formulation}

Intrinsic similarity to a reference does not fully establish whether generated descriptions contain useful geographic knowledge. We therefore introduce geospatial grounding as an independent extrinsic evaluation task. Given a natural-language description, a separate grounding model predicts a probability distribution over geographic space:
\begin{equation}
    g_{\phi}
    :
    \mathcal{Y}
    \rightarrow
    \mathcal{P}(\Omega),
    \qquad
    g_{\phi}(y)
    =
    q_{\phi}
    \bigl(
        \cdot \mid y
    \bigr),
\end{equation}
where locations consistent with the query receive higher probabilities. This formulation supports both precise descriptions of individual locations and spatially ambiguous descriptions of regions.

We query the captioning model at augmentation locations $\mathcal{X}_{\mathrm{aug}}$ in data-sparse regions and construct the augmented dataset $\widetilde{\mathcal{D}}_{\mathrm{aug}}=\bigl\{(\tilde{\mathbf{x}}_{j},\tilde{y}_{j}):\tilde{\mathbf{x}}_{j}\in\mathcal{X}_{\mathrm{aug}},\;\tilde{y}_{j}\sim p_{\psi}(y\mid\tilde{\mathbf{x}}_{j})\bigr\}$. We then compare a grounding model trained on $\mathcal{D}_{\mathrm{train}}$ with one trained on $\mathcal{D}_{\mathrm{train}}\cup\widetilde{\mathcal{D}}_{\mathrm{aug}}$. Improved grounding performance indicates that the LLM's generalization provides useful and geographically consistent supervision rather than merely plausible text. By filling gaps in the observed geo-text distribution, such improvements provide evidence that the captioning model has learned a coherent latent spatial field and can infer meaningful geographic knowledge at previously uncovered locations.

\section{Spatially Explicit LLMs}
\label{sec:method}

Having formulated geospatial reasoning as reconstructing an unobserved portion of a latent spatial field, we next consider how locations should be represented within an LLM. The input representation determines the geometry through which the model observes space and, consequently, the spatial field it can learn. When a coordinate such as \texttt{(34.0° N, 118.2° W)} is represented as text, it is decomposed into a sequence of discrete tokens whose embedding similarity does not necessarily preserve geographic proximity: nearby coordinates may differ in multiple tokens, while distant coordinates may share many. The LLM must therefore recover the continuous structure of space while simultaneously learning how geographic values vary over it.

We instead encode each location continuously and inject it into the LLM as a spatial token, providing an explicit inductive bias for the continuity and multi-scale structure of geographic space. Let $PE_{\mathrm{geo}}:\Omega\rightarrow\mathbb{R}^{d_{\mathrm{geo}}}$ denote a spatial encoder whose representation may capture absolute location, direction, and multi-scale spatial patterns. To align this representation with the dimension and distribution of the LLM token embeddings, we introduce a learned adapter $A\in\mathbb{R}^{d_{\mathrm{geo}}\times d_{\mathrm{LLM}}}$:
\begin{equation}
    \mathbf{u}_{\mathbf{x}}
    =
    PE_{\mathrm{geo}}(\mathbf{x}),
    \qquad
    \mathbf{Z}_{\mathbf{x}}
    =
    \mathbf{u}_{\mathbf{x}}A
    \in
    \mathbb{R}^{d_{\mathrm{LLM}}}.
\end{equation}
We replace the textual coordinate tokens with $\mathbf{Z}_{\mathbf{x}}$ and refer to the resulting architecture as a \textbf{spatially explicit LLM}, in which location is represented as a continuous modality that preserves spatial relationships. Our central hypothesis is that this representation encourages the LLM to organize geographic knowledge as a smoother and more generalizable latent spatial field.

\section{Experiments}
\label{sec:experiments}

\subsection{Datasets and Settings}
\label{sec:datasets}

We construct paired location--description datasets from Airbnb listings. Each listing contains a human-written property overview that describes its location and other context. We preprocess these overviews to extract self-contained, location-specific descriptions, forming overall data for captioning and grounding. To evaluate spatial generalization, we partition each city into \textbf{disjoint} training and evaluation regions and curate benchmarks for both tasks. The grounding benchmark includes two settings: \textbf{point-level grounding} for descriptions associated with specific coordinates and \textbf{region-level grounding} for descriptions associated with a region. We conduct experiments independently on Los Angeles and London. Further details are provided in Appendices~\ref{appendix:dataset} and~\ref{appendix:benchmark_construction}.

We evaluate captioning using BERTScore F1 and ROUGE-L F1. For point-level grounding, we report Hit Rate@$1\%$ Area and Hit Rate@$5\%$ Area, which measure the percentage of ground-truth locations that fall within the highest-scoring region covering $1\%$ or $5\%$ of the domain. For region-level grounding, coverage rate, Jensen--Shannon divergence (JSD) and Wasserstein distance measure the alignment between predicted and ground-truth distributions. Details are provided in Appendix~\ref{appendix:metric}.

\subsection{Spatial Representations and LLMs}
\label{sec:representations_models}

We compare four spatial representation methods and one hybrid variant:

\begin{itemize}[leftmargin=*, nosep]

    \item \textbf{GRID}: The geographic domain is divided into an $n \times n$ grid. Each location is assigned to a unique grid cell, and the corresponding grid identifier is provided to the LLM~\citep{geollama}.
    
    \item \textbf{HASH}: Each coordinate is encoded as a length-$9$ Base32 geohash, providing a hierarchical spatial index with approximately $2.4$ meters of positional precision~\citep{reageo}.

    \item \textbf{TEXT}: The coordinate pair is directly represented as text and provided to the LLM as part of the input prompt~\citep{yuhan2025}.


    \item \textbf{LATENT}: Each coordinate is encoded as a $144$-dimensional continuous spatial representation using the positional encoder introduced in Geo2Vec~\citep{geo2vec}.

    \item \textbf{HYBRID}: The textual coordinate and latent representation are provided together. The \textbf{TEXT} representation is followed by the corresponding \textbf{LATENT} representation in the input sequence.
\end{itemize}

We evaluate these representations across three LLM families. \textbf{GPT-OSS} denotes \texttt{gpt-oss-20b}\footnote{\url{https://huggingface.co/openai/gpt-oss-20b}}, \textbf{Qwen3.5} denotes \texttt{Qwen3.5-9B}\footnote{\url{https://huggingface.co/Qwen/Qwen3.5-9B}}, and \textbf{Gemma4} denotes \texttt{Gemma-4-12B}\footnote{\url{https://huggingface.co/google/gemma-4-12B}}. Qwen3.5 and Gemma4 natively support multimodal inputs, whereas GPT-OSS is a text-only mixture-of-experts model. We evaluate each LLM using all spatial representation methods. For a fair comparison, all models are trained separately on each dataset using the same training samples, experimental procedures, and random seeds. Further training details are provided in Appendix~\ref{appendix:captioning}.

\subsection{Evaluation of Geospatial Memory}
\label{sec:memory_results}

\begin{wraptable}{r}{0.55\textwidth}
\centering


\caption{\textbf{Geospatial captioning performance on observed locations.} Evaluation of memory. Bold and underlined denote the best and second-best results.}
\label{tab:memory}

\fontsize{8.5pt}{9.5pt}\selectfont
\setlength{\tabcolsep}{1.8pt}
\renewcommand{\arraystretch}{0.95}

\begin{tabular}{@{}lcccc@{}}
\toprule
& \multicolumn{2}{c}{Los Angeles}
& \multicolumn{2}{c}{London} \\
\cmidrule(lr){2-3}
\cmidrule(lr){4-5}
Method
& BERT $\uparrow$
& ROUGE $\uparrow$
& BERT $\uparrow$
& ROUGE $\uparrow$ \\
\midrule

Random
& 0.4410 & 0.2070
& 0.4664 & 0.2282 \\

\midrule
GPT-OSS
& 0.4048 & 0.1155
& 0.4158 & 0.1523 \\
Qwen3.5
& 0.4255 & 0.1477
& 0.4555 & 0.2094 \\
Gemma4
& 0.4392 & 0.2245
& 0.5012 & 0.2654 \\

\midrule
GPT-OSS-GRID
& 0.5125 & 0.2884
& 0.5449 & 0.3299 \\
GPT-OSS-HASH
& 0.5376 & 0.3190
& 0.5568 & 0.3414 \\

GPT-OSS-TEXT
& 0.5582 & 0.3462
& 0.5657 & 0.3548 \\
GPT-OSS-HYBRID
& 0.5511 & 0.3322
& 0.5728 & 0.3627 \\
GPT-OSS-LATENT
& 0.5562 & 0.3409
& 0.5730 & 0.3630 \\

\midrule
Qwen3.5-GRID
& 0.5593 & 0.3566
& 0.5694 & 0.3604 \\
Qwen3.5-HASH
& 0.5588 & 0.3520
& 0.5822 & 0.3793 \\
Qwen3.5-TEXT
& \underline{0.5623} & \underline{0.3585}
& \textbf{0.5849} & \textbf{0.3821} \\
Qwen3.5-HYBRID
& 0.5565 & 0.3444
& 0.5795 & 0.3752 \\
Qwen3.5-LATENT
& 0.5536 & 0.3401
& 0.5771 & 0.3711 \\

\midrule
Gemma4-GRID
& 0.5464 & 0.3363
& 0.5640 & 0.3537 \\
Gemma4-HASH
& 0.5619 & 0.3513
& 0.5812 & 0.3805 \\

Gemma4-TEXT
& 0.5567 & 0.3478
& \underline{0.5833} & \underline{0.3817} \\
Gemma4-HYBRID
& 0.5580 & 0.3500
& 0.5770 & 0.3746 \\
Gemma4-LATENT
& 0.5524 & 0.3393
& 0.5678 & 0.3591 \\

\bottomrule
\end{tabular}
\end{wraptable}

We first evaluate the LLMs' zero-shot performance on the observed locations. In Table~\ref{tab:memory}, rows labeled only with the model name report zero-shot results without tuning. \texttt{Random} is a baseline that assigns each query a description sampled randomly from the training descriptions.

First, zero-shot performance is poor. GPT-OSS and Qwen3.5 underperform random sampling, while Gemma4 performs only slightly better; all three models remain substantially below their fine-tuned variants. These results suggest that the pretrained models contain limited geospatial knowledge about the queried cities. The substantial improvements after fine-tuning can therefore be attributed to the acquisition of geospatial memory.

Among the fine-tuned models, \textbf{TEXT} achieves the strongest overall memory performance, followed by the \textbf{HASH} and \textbf{HYBRID} representations. This result is expected because an exact textual coordinate serves as a unique identifier for an observed location, and LLMs are effective at learning direct token-to-token associations. In contrast, \textbf{GRID} generally performs worse because multiple locations share the same grid identifier, limiting the accurate retrieval of location-specific knowledge.

\subsection{Intrinsic Evaluation of Geospatial Reasoning}
\label{sec:reasoning_results}

We introduce \texttt{k-NN} as a retrieval-based baseline. For each test location, \texttt{k-NN} retrieves the descriptions associated with its $k=4$ nearest training locations as predictions. This baseline tests whether LLMs can infer geographic knowledge beyond simply copying descriptions from nearby observations. Distance-stratified evaluation can be found in Appendix~\ref{appendix:distance_stratified}.

\begin{wraptable}{r}{0.59\textwidth}
\centering
\caption{\textbf{Geospatial captioning performance across unobserved regions.} Evaluation of spatial generalization. }

\label{tab:caption}

\fontsize{8.5pt}{9.5pt}\selectfont
\setlength{\tabcolsep}{1.8pt}
\renewcommand{\arraystretch}{0.95}

\begin{tabular}{@{}lcccc@{}}
\toprule
& \multicolumn{2}{c}{Los Angeles}
& \multicolumn{2}{c}{London} \\
\cmidrule(lr){2-3}
\cmidrule(lr){4-5}
Method
& BERT $\uparrow$
& ROUGE $\uparrow$
& BERT $\uparrow$
& ROUGE $\uparrow$ \\
\midrule

k-NN
& 0.5213 & 0.2931
& 0.5306 & 0.2980 \\

\midrule
GPT-OSS-GRID
& 0.4921 & 0.2595
& 0.5084 & 0.2871 \\
GPT-OSS-HASH
& 0.5320 & 0.3109
& 0.5348 & 0.3162 \\
GPT-OSS-TEXT
& 0.5329 & 0.3078
& 0.5524 & 0.3369 \\
GPT-OSS-HYBRID
& \underline{0.5503} & \underline{0.3289}
& 0.5597 & 0.3447 \\
GPT-OSS-LATENT
& \textbf{0.5524} & \textbf{0.3295}
& 0.5606 & \underline{0.3459} \\

\midrule
Qwen3.5-GRID
& 0.4929 & 0.2663
& 0.5242 & 0.2959 \\
Qwen3.5-HASH
& 0.5385 & 0.3167
& 0.5425 & 0.3245 \\
Qwen3.5-TEXT
& 0.5345 & 0.3243
& 0.5580 & 0.3377 \\


Qwen3.5-HYBRID
& 0.5441 & 0.3240
& \underline{0.5613} & 0.3449 \\
Qwen3.5-LATENT
& 0.5432 & 0.3212
& 0.5607 & 0.3455 \\

\midrule
Gemma4-GRID
& 0.4888 & 0.2602
& 0.5218 & 0.2973 \\
Gemma4-HASH
& 0.5435 & 0.3225
& 0.5425 & 0.3245 \\
Gemma4-TEXT
& 0.5423 & 0.3195
& 0.5599 & 0.3434 \\
Gemma4-HYBRID
& 0.5439 & 0.3237
& 0.5609 & 0.3451 \\
Gemma4-LATENT
& 0.5453 & 0.3287
& \textbf{0.5617} & \textbf{0.3475} \\

\bottomrule
\end{tabular}

\vspace{1pt}
\end{wraptable}

As shown in Table~\ref{tab:caption}, \textbf{GRID} consistently underperforms all methods, including \texttt{k-NN}, because the grid identifiers encode no spatial relationship at all. In contrast, all other fine-tuned representations outperform nearest-neighbor retrieval, indicating that spatially tuned LLMs can learn geographic patterns beyond simply copying nearby observations.


More importantly, the generalization results reverse the pattern observed for geospatial memory, where \textbf{TEXT} achieves the strongest overall performance. For generalization, \textbf{LATENT} and \textbf{HYBRID} consistently outperform \textbf{TEXT} variants across almost all LLM families and both datasets. This contrast empirically supports our distinction between geospatial memory and reasoning. The gains also extend to GPT-OSS, despite its text-only architecture, demonstrating that continuous spatial representations can be effectively integrated into different LLM families. Their consistent improvements suggest that spatially explicit representations provide a broadly applicable foundation for strengthening geospatial awareness in LLMs.

\subsection{Extrinsic Evaluation of Geospatial Reasoning}
\label{sec:reasoning_results_ex}

We further evaluate the inferred geographic knowledge through the geospatial grounding framework. For each augmentation setting, we train a separate CLIP-based grounding model over five runs, using different seeds but otherwise identical settings across representation methods. \texttt{GeoGrounding} denotes the baseline trained only on $\mathcal{D}_{\mathrm{train}}$, whereas \texttt{w/} denotes models trained with augmented dataset generated by the corresponding captioning model. More details are provided in Appendix~\ref{appendix:grounding}.

Tables~\ref{tab:grounding_single} and~\ref{tab:grounding_multi} present the point- and region-level grounding results, respectively. Although \textbf{TEXT} substantially outperforms \texttt{k-NN} on BERTScore and ROUGE-L, its descriptions do not consistently improve either grounding task, showing that reference similarity does not guarantee useful or spatially coherent knowledge. Grounding therefore provides complementary evidence beyond conventional NLP metrics. Meanwhile, \textbf{GRID} augmentation consistently degrades both tasks, since its less accurate descriptions introduce noisy supervision and distort text--space alignment. Thus, augmentation is beneficial only when the generated descriptions preserve meaningful geographic structure.

\begin{table*}[t]
\centering
\caption{\textbf{Point-level geospatial grounding performance under different data augmentations.} Values following $\pm$ denote standard deviations. $\uparrow$ indicates that higher values are better.}
\label{tab:grounding_single}

\begingroup
\fontsize{7.5pt}{9pt}\selectfont
\setlength{\tabcolsep}{1.2pt}
\renewcommand{\arraystretch}{1.05}

\begin{tabular}{@{}lcccc@{}}
\toprule
& \multicolumn{2}{c}{Los Angeles}
& \multicolumn{2}{c}{London} \\
\cmidrule(lr){2-3}
\cmidrule(lr){4-5}
Method
& Hit Rate@1\% $\uparrow$
& Hit Rate@5\% $\uparrow$
& Hit Rate@1\% $\uparrow$
& Hit Rate@5\% $\uparrow$ \\
\midrule

GeoGrounding
& 34.90\% {\scriptsize $\pm$ 0.52\%}
& \underline{54.81\%} {\scriptsize $\pm$ 0.57\%}
& 55.07\% {\scriptsize $\pm$ 0.33\%}
& 69.26\% {\scriptsize $\pm$ 0.70\%} \\

w/ k-NN
& 35.05\% {\scriptsize $\pm$ 0.46\%}
& 54.57\% {\scriptsize $\pm$ 0.41\%}
& 55.52\% {\scriptsize $\pm$ 0.25\%}
& 69.52\% {\scriptsize $\pm$ 0.32\%} \\

\midrule

w/ GPT-OSS-GRID
& 30.70\% {\scriptsize $\pm$ 0.26\%}
& 51.39\% {\scriptsize $\pm$ 0.78\%}
& 53.24\% {\scriptsize $\pm$ 0.56\%}
& 67.41\% {\scriptsize $\pm$ 0.92\%} \\

w/ GPT-OSS-HASH
& 34.79\% {\scriptsize $\pm$ 0.45\%}
& 54.46\% {\scriptsize $\pm$ 0.43\%}
& 55.10\% {\scriptsize $\pm$ 0.66\%}
& 69.04\% {\scriptsize $\pm$ 0.99\%} \\

w/ GPT-OSS-TEXT
& 34.84\% {\scriptsize $\pm$ 0.26\%}
& 54.60\% {\scriptsize $\pm$ 0.31\%}
& 55.34\% {\scriptsize $\pm$ 0.63\%}
& 69.16\% {\scriptsize $\pm$ 0.66\%} \\

w/ GPT-OSS-HYBRID
& 35.09\% {\scriptsize $\pm$ 0.41\%}
& 54.62\% {\scriptsize $\pm$ 0.68\%}
& 55.54\% {\scriptsize $\pm$ 0.37\%}
& 69.43\% {\scriptsize $\pm$ 0.21\%} \\

w/ GPT-OSS-LATENT
& 35.00\% {\scriptsize $\pm$ 0.44\%}
& 54.65\% {\scriptsize $\pm$ 0.24\%}
& 55.52\% {\scriptsize $\pm$ 0.28\%}
& 69.40\% {\scriptsize $\pm$ 0.45\%} \\

\midrule

w/ Qwen3.5-GRID
& 31.92\% {\scriptsize $\pm$ 0.31\%}
& 52.47\% {\scriptsize $\pm$ 0.50\%}
& 53.98\% {\scriptsize $\pm$ 0.73\%}
& 67.99\% {\scriptsize $\pm$ 0.78\%} \\

w/ Qwen3.5-HASH
& 34.96\% {\scriptsize $\pm$ 0.47\%}
& 54.60\% {\scriptsize $\pm$ 0.42\%}
& 55.19\% {\scriptsize $\pm$ 0.80\%}
& 69.17\% {\scriptsize $\pm$ 0.89\%} \\

w/ Qwen3.5-TEXT
& 34.98\% {\scriptsize $\pm$ 0.50\%}
& 54.54\% {\scriptsize $\pm$ 0.49\%}
& 55.43\% {\scriptsize $\pm$ 0.75\%}
& 69.19\% {\scriptsize $\pm$ 0.72\%} \\

w/ Qwen3.5-HYBRID
& 35.07\% {\scriptsize $\pm$ 0.35\%}
& 54.41\% {\scriptsize $\pm$ 0.54\%}
& 55.30\% {\scriptsize $\pm$ 0.41\%}
& 69.25\% {\scriptsize $\pm$ 0.57\%} \\

w/ Qwen3.5-LATENT
& \underline{35.40\%} {\scriptsize $\pm$ 0.47\%}
& 54.76\% {\scriptsize $\pm$ 0.43\%}
& \underline{55.91\%} {\scriptsize $\pm$ 0.40\%}
& \underline{69.88\%} {\scriptsize $\pm$ 0.45\%} \\

\midrule

w/ Gemma4-GRID
& 30.57\% {\scriptsize $\pm$ 0.71\%}
& 51.27\% {\scriptsize $\pm$ 0.97\%}
& 53.42\% {\scriptsize $\pm$ 0.78\%}
& 67.97\% {\scriptsize $\pm$ 1.12\%} \\

w/ Gemma4-HASH
& 34.92\% {\scriptsize $\pm$ 0.23\%}
& 54.77\% {\scriptsize $\pm$ 0.26\%}
& 55.67\% {\scriptsize $\pm$ 0.33\%}
& 69.76\% {\scriptsize $\pm$ 0.48\%} \\

w/ Gemma4-TEXT
& 34.96\% {\scriptsize $\pm$ 0.39\%}
& 54.48\% {\scriptsize $\pm$ 0.40\%}
& 55.58\% {\scriptsize $\pm$ 0.37\%}
& 69.68\% {\scriptsize $\pm$ 0.59\%} \\

w/ Gemma4-HYBRID
& 35.03\% {\scriptsize $\pm$ 0.53\%}
& 54.47\% {\scriptsize $\pm$ 0.60\%}
& 55.28\% {\scriptsize $\pm$ 0.77\%}
& 69.30\% {\scriptsize $\pm$ 0.89\%} \\

w/ Gemma4-LATENT
& \textbf{35.89\%} {\scriptsize $\pm$ 0.39\%}
& \textbf{55.05\%} {\scriptsize $\pm$ 0.52\%}
& \textbf{55.96\%} {\scriptsize $\pm$ 0.25\%}
& \textbf{70.11\%} {\scriptsize $\pm$ 0.43\%} \\

\bottomrule
\end{tabular}
\endgroup
\end{table*}

\begin{table*}[t]

\centering
\caption{\textbf{Region-level geospatial grounding performance under different data augmentations.} $\downarrow$ indicates that lower values are better. }
\label{tab:grounding_multi}

\begingroup
\fontsize{7.5pt}{9pt}\selectfont
\setlength{\tabcolsep}{1.2pt}
\renewcommand{\arraystretch}{1.05}

\begin{tabular}{@{}lllllll@{}}
\toprule
& \multicolumn{3}{c}{Los Angeles}
& \multicolumn{3}{c}{London} \\
\cmidrule(lr){2-4} \cmidrule(lr){5-7}

Method
& \multicolumn{1}{c}{Coverage@1\% $\uparrow$}
& \multicolumn{1}{c}{JSD $\downarrow$}
& \multicolumn{1}{c}{Wasserstein $\downarrow$}
& \multicolumn{1}{c}{Coverage@1\% $\uparrow$}
& \multicolumn{1}{c}{JSD $\downarrow$}
& \multicolumn{1}{c}{Wasserstein $\downarrow$} \\
\midrule

GeoGrounding
& \underline{60.22\%} {\scriptsize $\pm$ 0.86\%}
& 0.4747 {\scriptsize $\pm$ 0.0019}
& 7459.29 {\scriptsize $\pm$ 116.46}
& 74.96\% {\scriptsize $\pm$ 0.31\%}
& 0.4632 {\scriptsize $\pm$ 0.0079}
& 6202.41 {\scriptsize $\pm$ 341.13} \\

w/ k-NN
& 60.06\% {\scriptsize $\pm$ 0.74\%}
& 0.4734 {\scriptsize $\pm$ 0.0029}
& 7394.78 {\scriptsize $\pm$ 71.03}
& 75.09\% {\scriptsize $\pm$ 0.39\%}
& 0.4693 {\scriptsize $\pm$ 0.0046}
& 6120.26 {\scriptsize $\pm$ 239.72} \\

\midrule

w/ GPT-OSS-GRID
& 58.57\% {\scriptsize $\pm$ 0.47\%}
& 0.4889 {\scriptsize $\pm$ 0.0047}
& 7743.80 {\scriptsize $\pm$ 57.20}
& 72.59\% {\scriptsize $\pm$ 0.33\%}
& 0.4895 {\scriptsize $\pm$ 0.0067}
& 6629.12 {\scriptsize $\pm$ 300.62} \\

w/ GPT-OSS-HASH
& 59.72\% {\scriptsize $\pm$ 0.72\%}
& 0.4716 {\scriptsize $\pm$ 0.0027}
& 7386.82 {\scriptsize $\pm$ 62.27}
& 74.82\% {\scriptsize $\pm$ 0.67\%}
& 0.4645 {\scriptsize $\pm$ 0.0098}
& 6075.04 {\scriptsize $\pm$ 334.89} \\

w/ GPT-OSS-TEXT
& 60.07\% {\scriptsize $\pm$ 0.51\%}
& 0.4690 {\scriptsize $\pm$ 0.0048}
& \underline{7327.45} {\scriptsize $\pm$ 115.96}
& 74.81\% {\scriptsize $\pm$ 0.66\%}
& 0.4690 {\scriptsize $\pm$ 0.0064}
& 6031.12 {\scriptsize $\pm$ 249.98} \\

w/ GPT-OSS-HYBRID
& 59.11\% {\scriptsize $\pm$ 0.55\%}
& 0.4709 {\scriptsize $\pm$ 0.0018}
& 7459.86 {\scriptsize $\pm$ 64.31}
& 74.91\% {\scriptsize $\pm$ 0.44\%}
& \underline{0.4536} {\scriptsize $\pm$ 0.0041}
& 5984.08 {\scriptsize $\pm$ 215.11} \\

w/ GPT-OSS-LATENT
& 59.24\% {\scriptsize $\pm$ 0.59\%}
& \underline{0.4672} {\scriptsize $\pm$ 0.0063}
& \textbf{7308.13} {\scriptsize $\pm$ 127.33}
& 74.76\% {\scriptsize $\pm$ 0.32\%}
& 0.4549 {\scriptsize $\pm$ 0.0039}
& 6006.23 {\scriptsize $\pm$ 168.65} \\

\midrule

w/ Qwen3.5-GRID
& 59.04\% {\scriptsize $\pm$ 0.53\%}
& 0.4863 {\scriptsize $\pm$ 0.0049}
& 7756.12 {\scriptsize $\pm$ 120.81}
& 73.46\% {\scriptsize $\pm$ 0.60\%}
& 0.4783 {\scriptsize $\pm$ 0.0071}
& 6707.53 {\scriptsize $\pm$ 275.28} \\

w/ Qwen3.5-HASH
& 60.11\% {\scriptsize $\pm$ 0.50\%}
& 0.4736 {\scriptsize $\pm$ 0.0037}
& 7438.10 {\scriptsize $\pm$ 76.03}
& 74.92\% {\scriptsize $\pm$ 0.87\%}
& 0.4635 {\scriptsize $\pm$ 0.0087}
& 6002.55 {\scriptsize $\pm$ 257.31} \\

w/ Qwen3.5-TEXT
& 60.21\% {\scriptsize $\pm$ 0.58\%}
& 0.4728 {\scriptsize $\pm$ 0.0019}
& 7383.66 {\scriptsize $\pm$ 82.94}
& 75.02\% {\scriptsize $\pm$ 0.71\%}
& 0.4653 {\scriptsize $\pm$ 0.0083}
& 5999.37 {\scriptsize $\pm$ 336.68} \\

w/ Qwen3.5-HYBRID
& 59.79\% {\scriptsize $\pm$ 0.60\%}
& 0.4720 {\scriptsize $\pm$ 0.0073}
& 7490.35 {\scriptsize $\pm$ 104.48}
& 74.50\% {\scriptsize $\pm$ 0.54\%}
& 0.4590 {\scriptsize $\pm$ 0.0095}
& 6165.56 {\scriptsize $\pm$ 288.68} \\

w/ Qwen3.5-LATENT
& 59.88\% {\scriptsize $\pm$ 0.80\%}
& 0.4715 {\scriptsize $\pm$ 0.0009}
& 7394.02 {\scriptsize $\pm$ 37.57}
& 75.12\% {\scriptsize $\pm$ 0.57\%}
& 0.4555 {\scriptsize $\pm$ 0.0067}
& 6069.78 {\scriptsize $\pm$ 254.73} \\

\midrule

w/ Gemma4-GRID
& 57.93\% {\scriptsize $\pm$ 0.65\%}
& 0.4908 {\scriptsize $\pm$ 0.0028}
& 7768.11 {\scriptsize $\pm$ 84.88}
& 73.34\% {\scriptsize $\pm$ 0.67\%}
& 0.4735 {\scriptsize $\pm$ 0.0100}
& 6253.76 {\scriptsize $\pm$ 320.27} \\

w/ Gemma4-HASH
& 60.01\% {\scriptsize $\pm$ 0.20\%}
& 0.4719 {\scriptsize $\pm$ 0.0020}
& 7389.10 {\scriptsize $\pm$ 72.46}
& \textbf{75.34\%} {\scriptsize $\pm$ 0.41\%}
& 0.4610 {\scriptsize $\pm$ 0.0068}
& \underline{5882.12} {\scriptsize $\pm$ 293.23} \\

w/ Gemma4-TEXT
& 59.98\% {\scriptsize $\pm$ 0.37\%}
& 0.4743 {\scriptsize $\pm$ 0.0032}
& 7429.47 {\scriptsize $\pm$ 89.79}
& 74.91\% {\scriptsize $\pm$ 0.24\%}
& 0.4563 {\scriptsize $\pm$ 0.0048}
& 5931.76 {\scriptsize $\pm$ 210.17} \\

w/ Gemma4-HYBRID
& 59.53\% {\scriptsize $\pm$ 0.87\%}
& 0.4707 {\scriptsize $\pm$ 0.0076}
& 7435.79 {\scriptsize $\pm$ 136.56}
& 74.60\% {\scriptsize $\pm$ 0.10\%}
& 0.4589 {\scriptsize $\pm$ 0.0107}
& 6017.19 {\scriptsize $\pm$ 280.55} \\

w/ Gemma4-LATENT
& \textbf{60.40\%} {\scriptsize $\pm$ 0.74\%}
& \textbf{0.4650} {\scriptsize $\pm$ 0.0060}
& 7350.43 {\scriptsize $\pm$ 50.92}
& \underline{75.33\%} {\scriptsize $\pm$ 0.23\%}
& \textbf{0.4513} {\scriptsize $\pm$ 0.0077}
& \textbf{5870.19} {\scriptsize $\pm$ 285.26} \\

\bottomrule
\end{tabular}
\endgroup
\end{table*}

\textbf{LATENT} provides the strongest overall augmentation performance. Gemma4-LATENT achieves the best mean on all four point-level metrics and four of the six region-level metrics. Although modest, these consistent gains, particularly in region-level distributional alignment, provide functional evidence that continuous representations promote a smoother and more structured latent spatial field, enabling more geographically coherent generation at unobserved locations.

In sum, \textbf{HYBRID} generally improves captioning and grounding over \textbf{TEXT} while retaining competitive memory performance, offering a practical balance between memorizing observed textual associations and generalizing over unobserved regions. Overall, the grounding results support our argument that learning a continuous latent spatial field enables stronger geospatial generalization.

\subsection{Evidence of Spatial Latent Field}

\begin{figure}[t]
\centering
\begingroup
\setlength{\tabcolsep}{1pt}
\renewcommand{\arraystretch}{0.85}

\begin{tabular}{@{}ccc@{}}
\includegraphics[width=0.325\columnwidth]{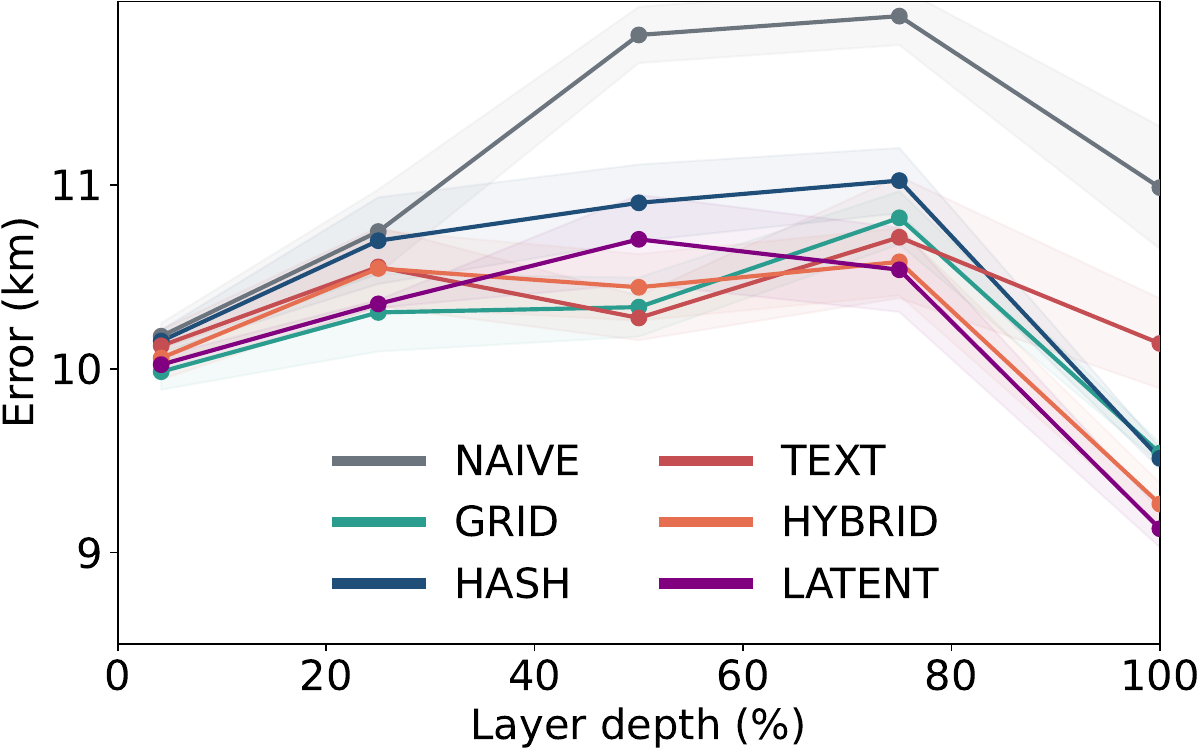}
&
\includegraphics[width=0.325\columnwidth]{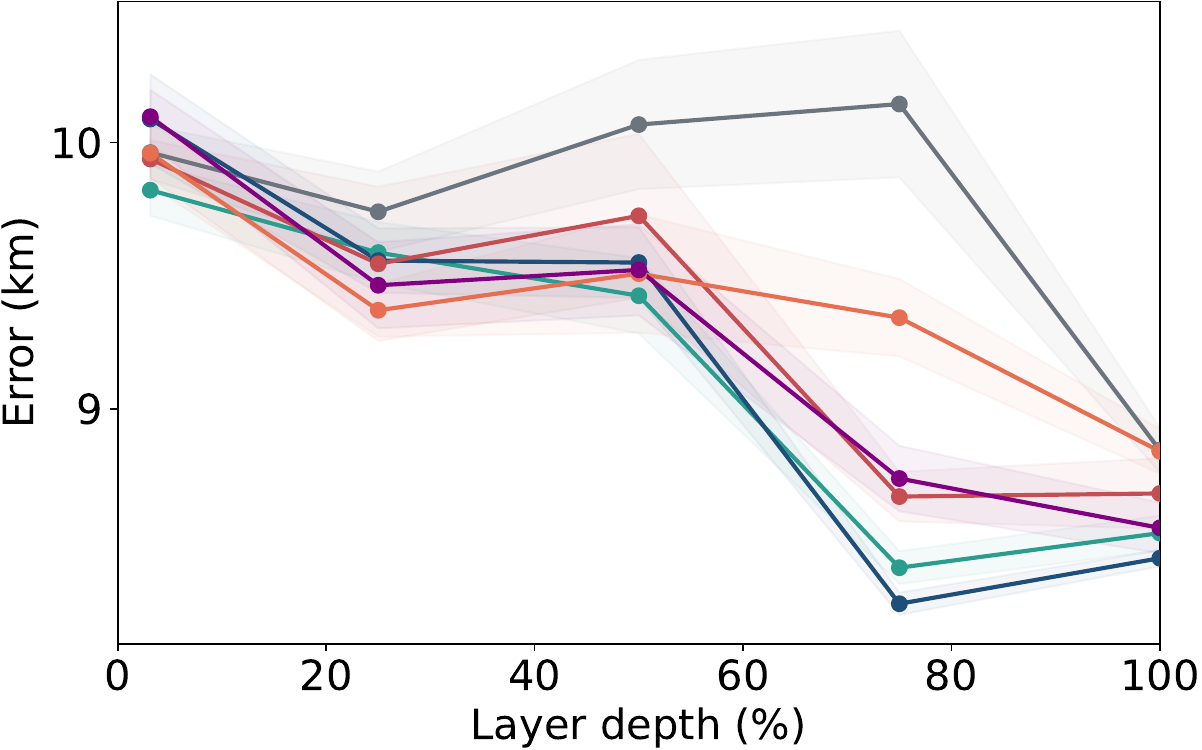}
&
\includegraphics[width=0.325\columnwidth]{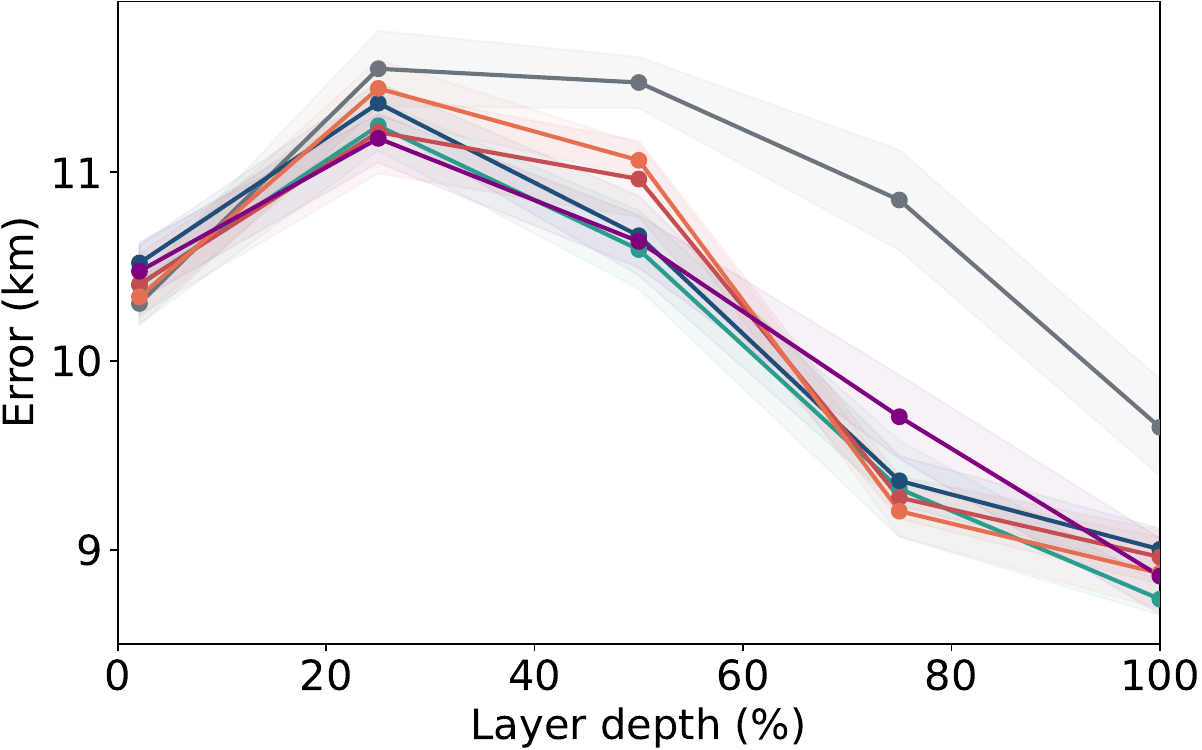}
\\[-1pt]
\scriptsize\textbf{GPT-OSS}
&
\scriptsize\textbf{Qwen3.5}
&
\scriptsize\textbf{Gemma4}
\end{tabular}

\endgroup
\vspace{-1.5mm}
\caption{\textbf{Layer-wise geospatial probing across models on LA.} Lower error indicates that location is more explicitly encoded. Curves and shaded regions show means and standard deviations.}
\label{fig:interlayer_analysis}
\vspace{-2mm}
\end{figure}

\begin{figure*}[t]
\centering
\begingroup
\setlength{\tabcolsep}{0pt}
\renewcommand{\arraystretch}{0}

\newcommand{\modellabel}[1]{%
    \begin{minipage}[c]{0.052\textwidth}
        \centering
        \fontsize{6pt}{6.5pt}\selectfont\bfseries #1
    \end{minipage}%
}

\newcommand{\clusterpanel}[1]{%
    \begin{minipage}[c]{0.157\textwidth}
        \centering
        \includegraphics[
            width=\linewidth
        ]{#1}
    \end{minipage}%
}

\begin{tabular}{@{}ccccccc@{}}
&
\scriptsize\textbf{NAIVE}
&
\scriptsize\textbf{GRID}
&
\scriptsize\textbf{HASH}
&
\scriptsize\textbf{TEXT}
&
\scriptsize\textbf{HYBRID}
&
\scriptsize\textbf{LATENT}
\\[1pt]

\modellabel{GPT}
&
\clusterpanel{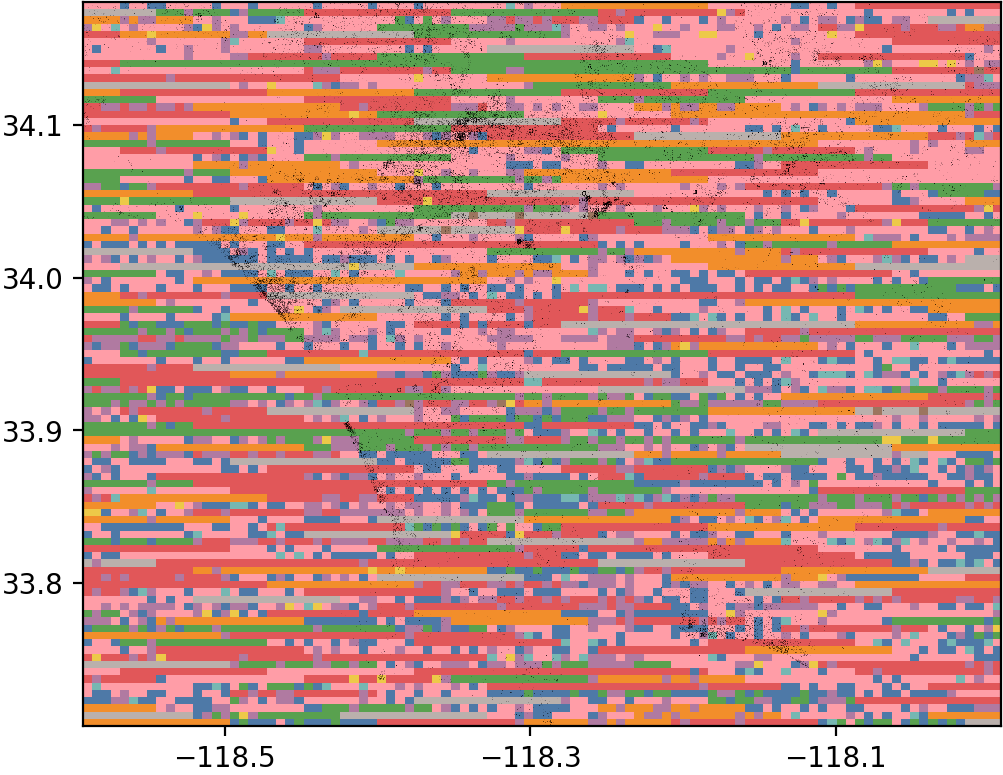}
&
\clusterpanel{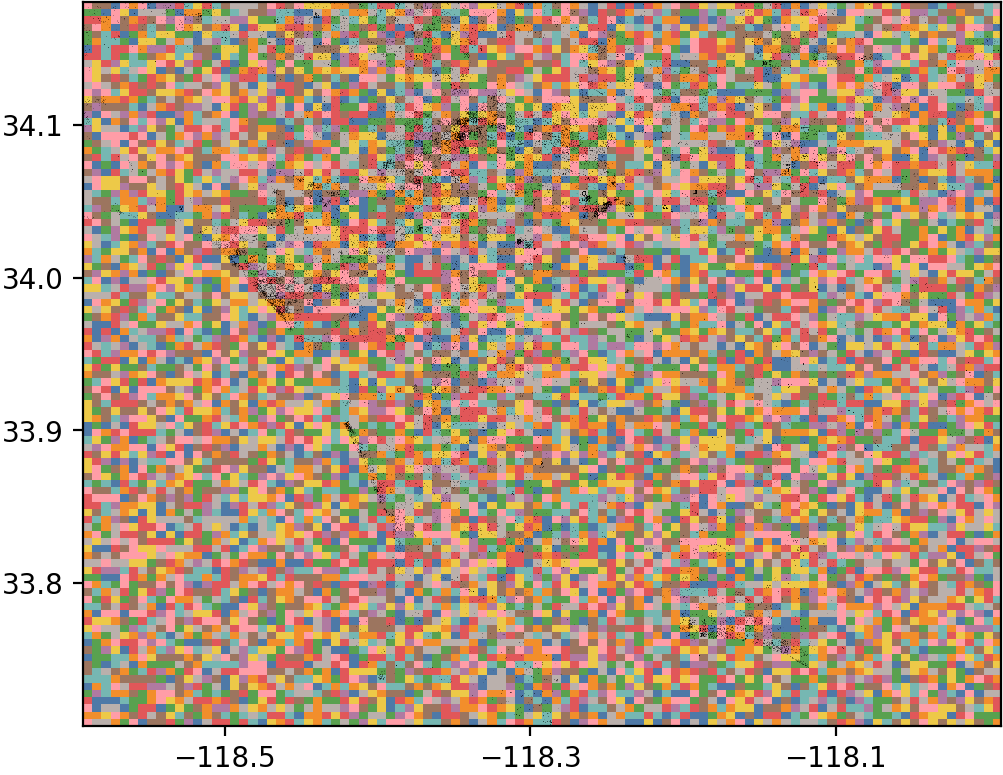}
&
\clusterpanel{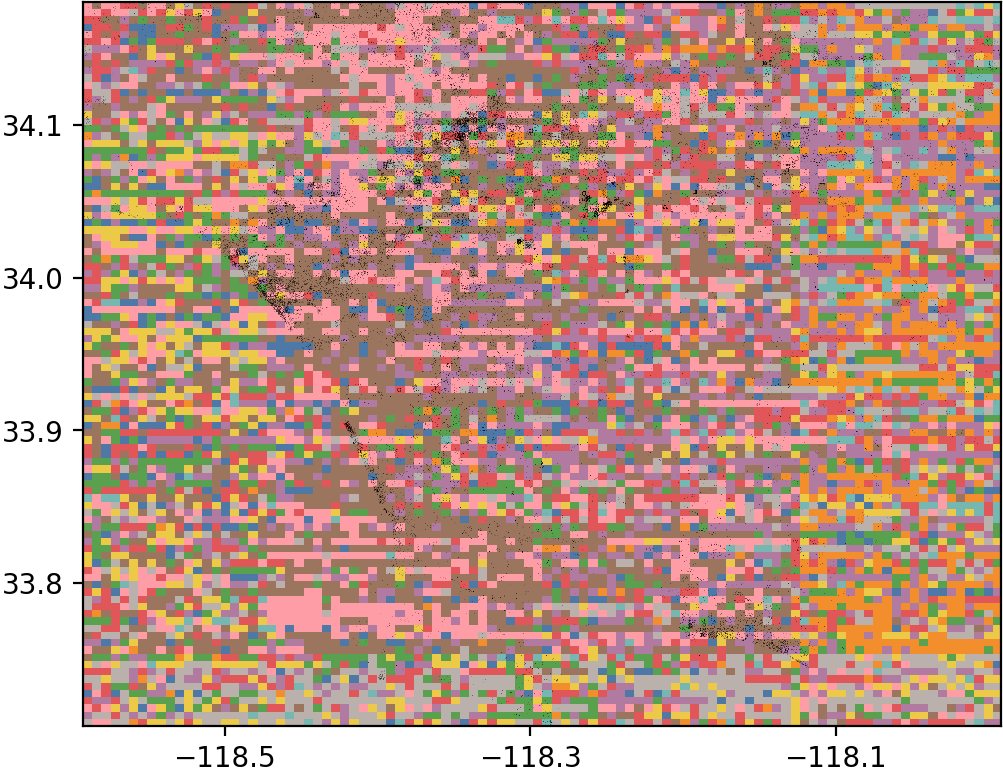}
&
\clusterpanel{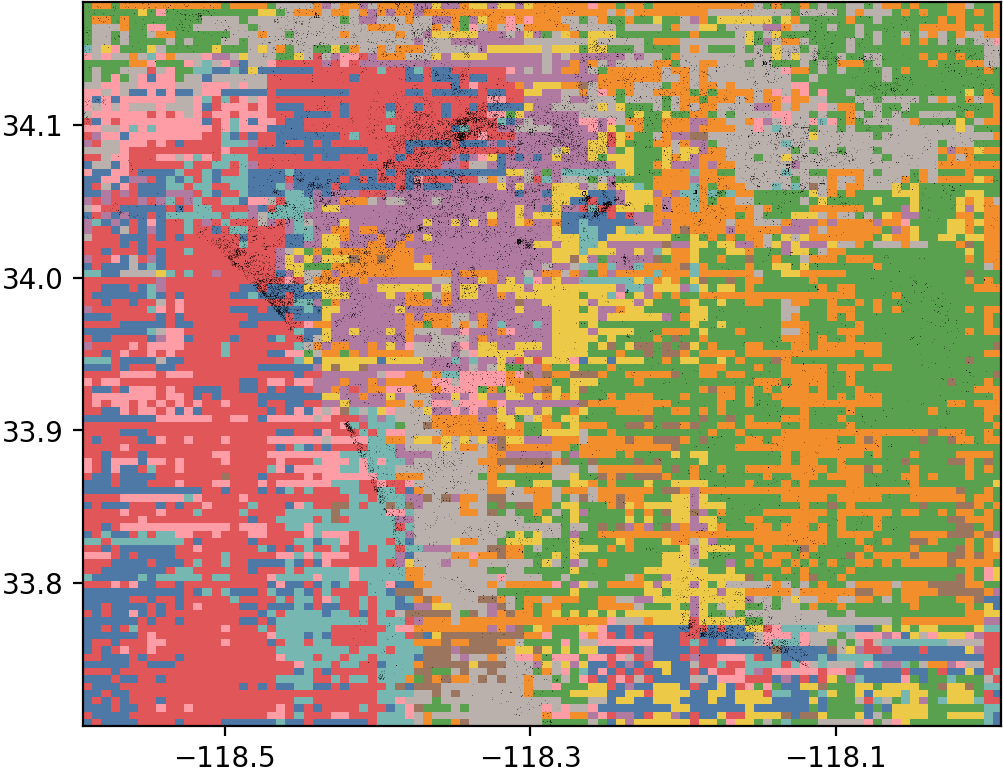}
&
\clusterpanel{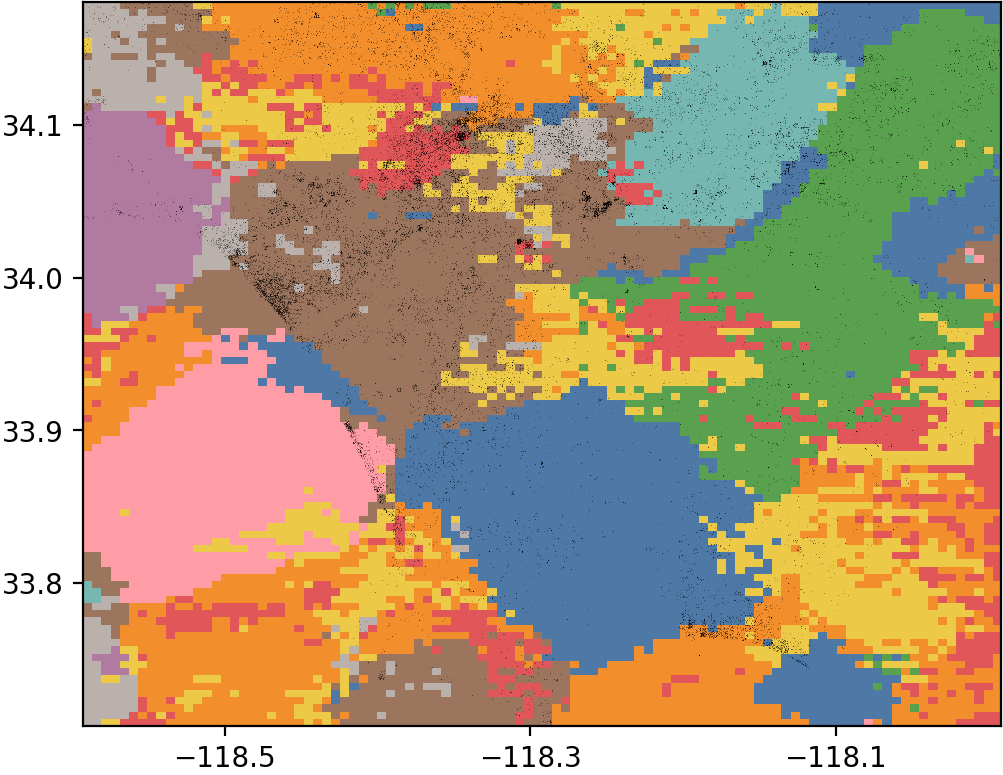}
&
\clusterpanel{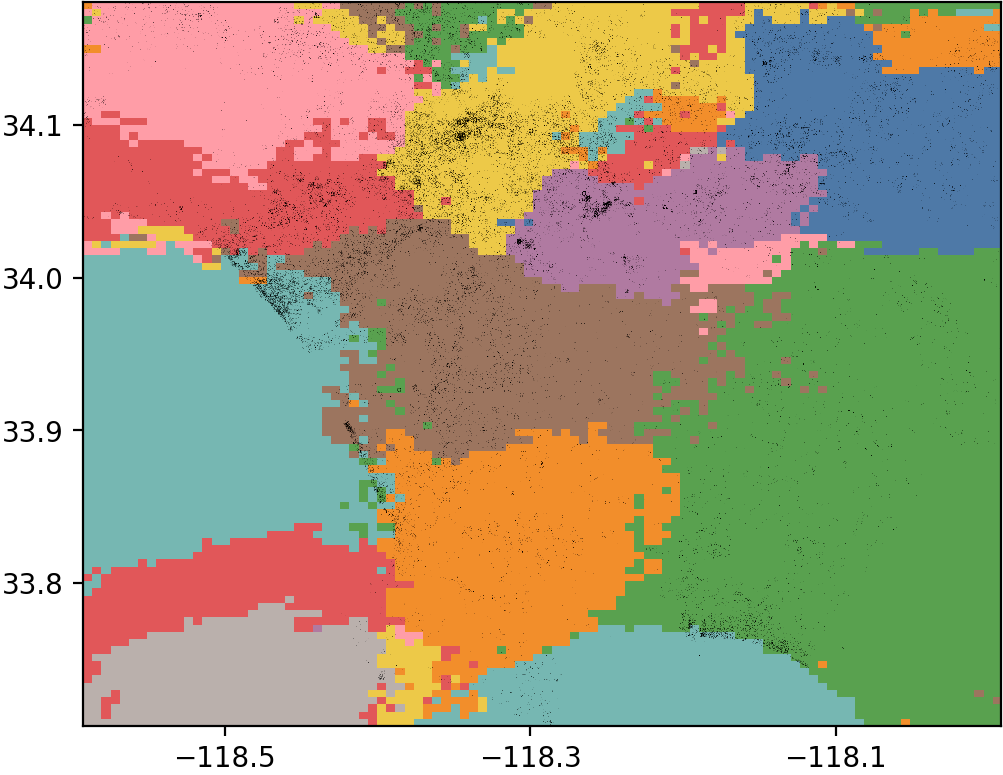}
\\[1pt]

\modellabel{Qwen}
&
\clusterpanel{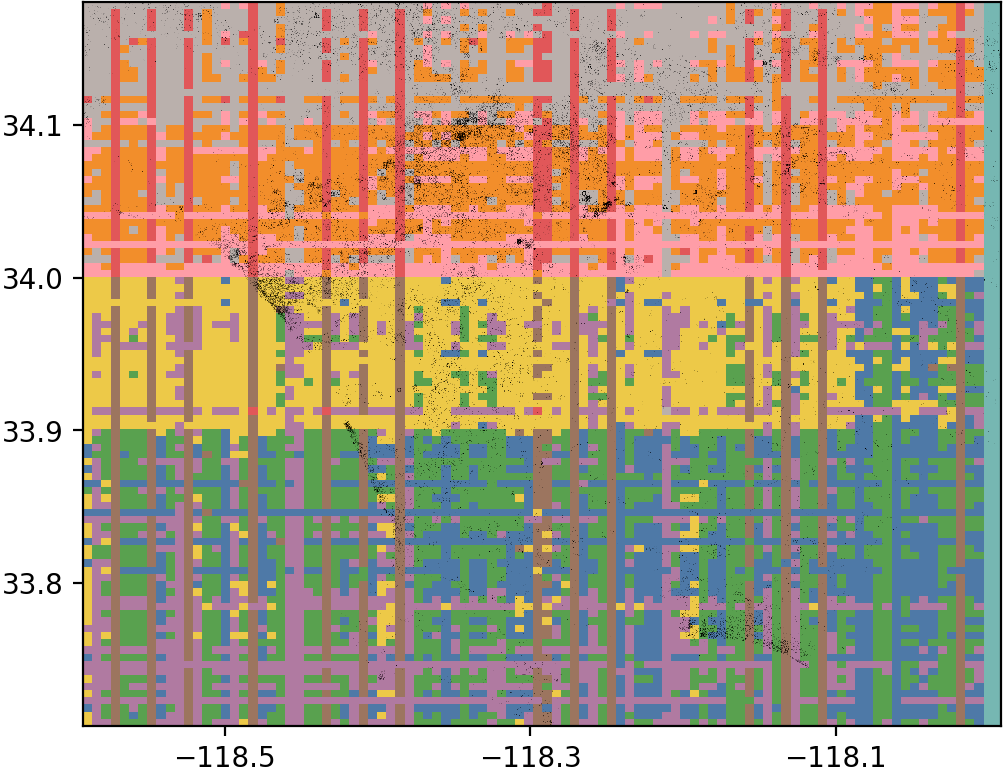}
&
\clusterpanel{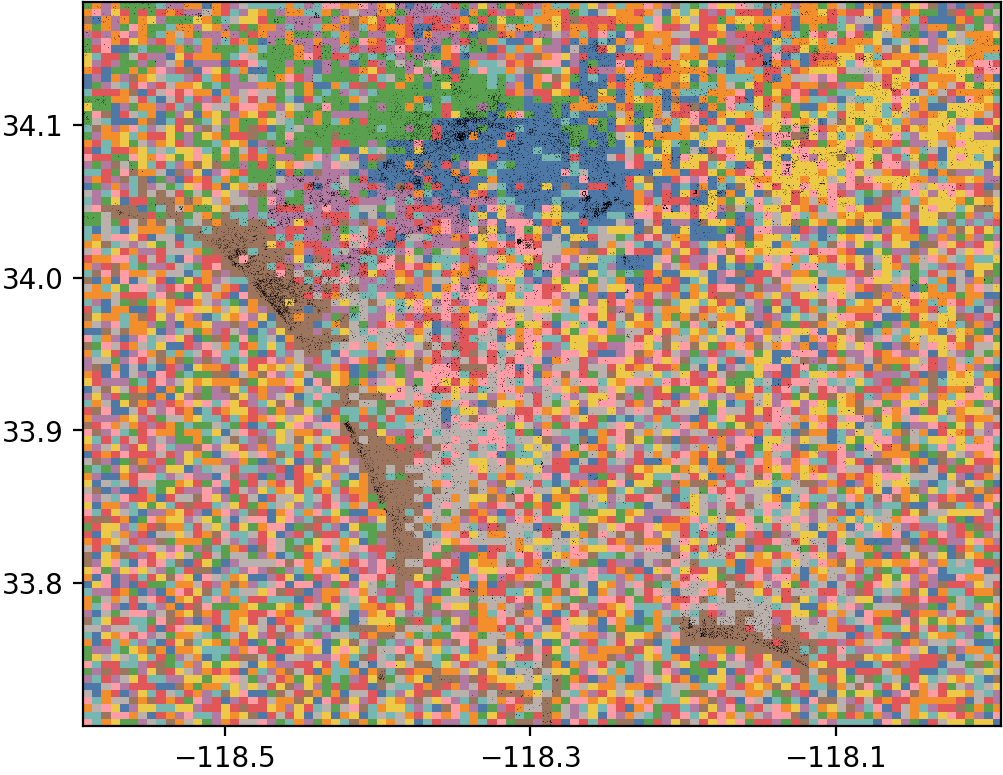}
&
\clusterpanel{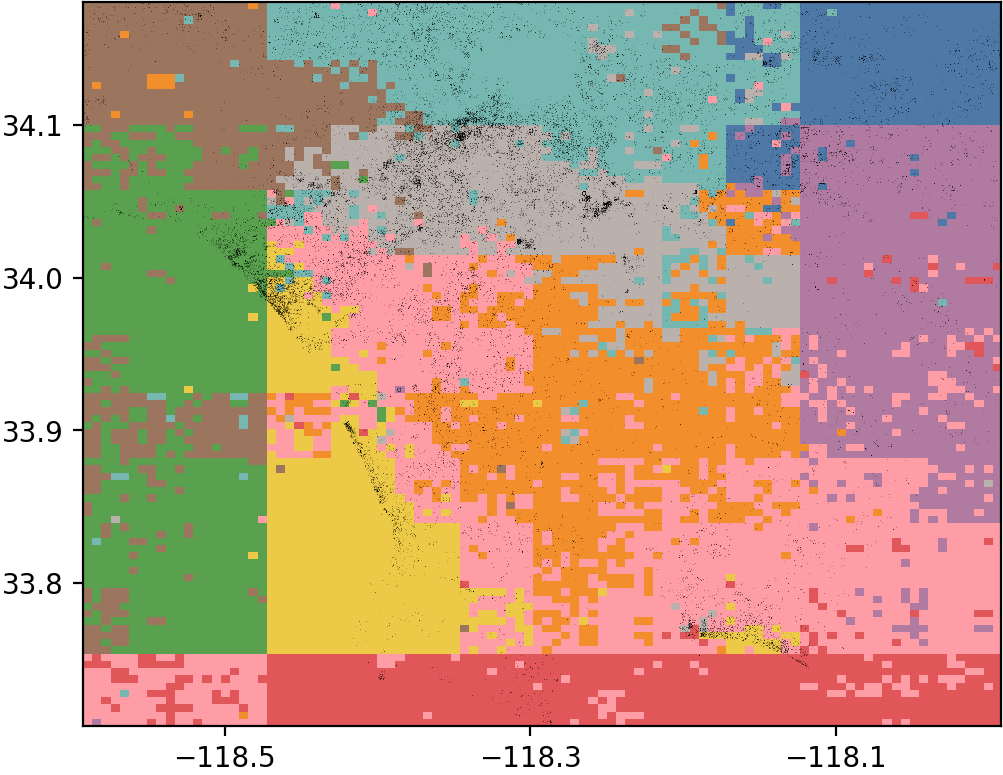}
&
\clusterpanel{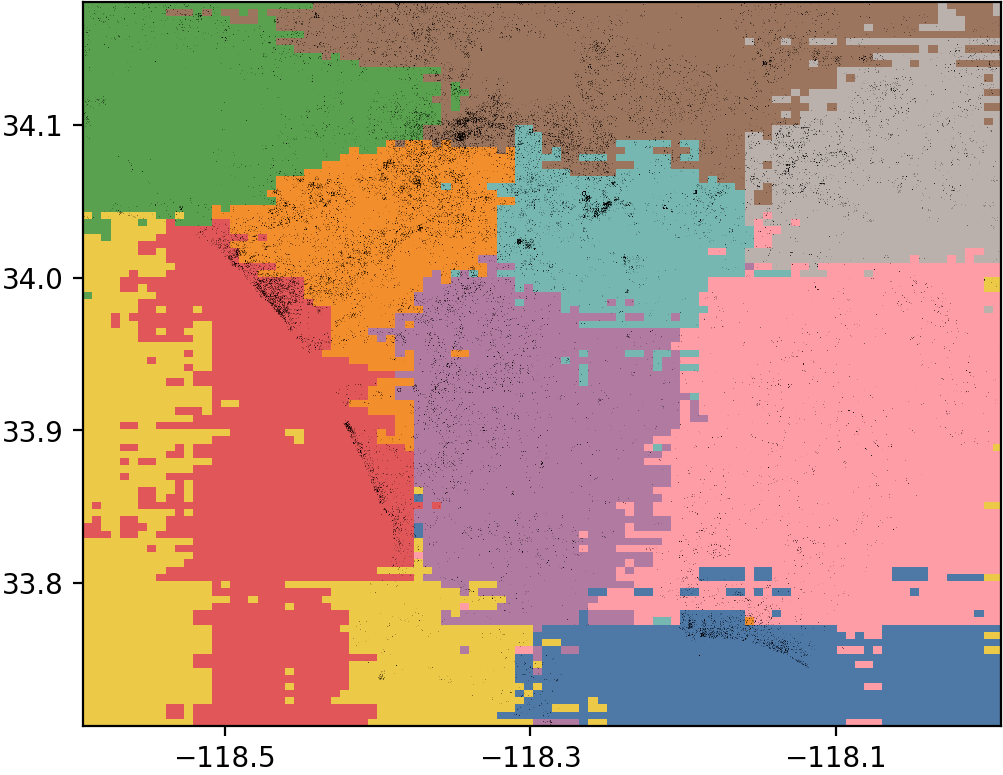}
&
\clusterpanel{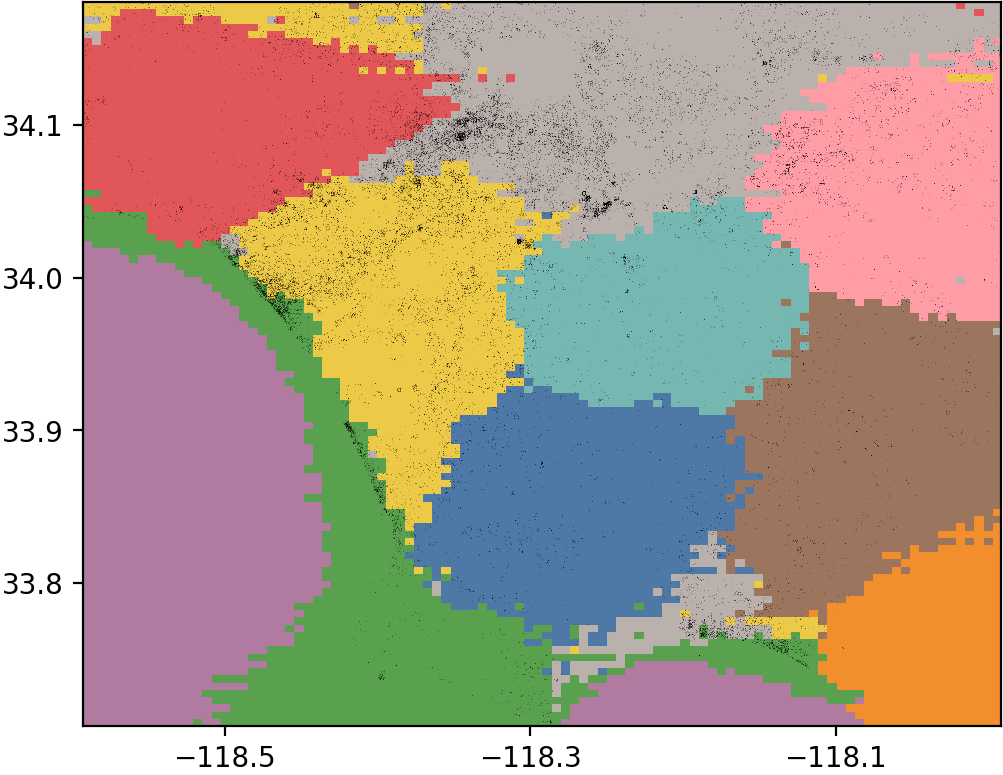}
&
\clusterpanel{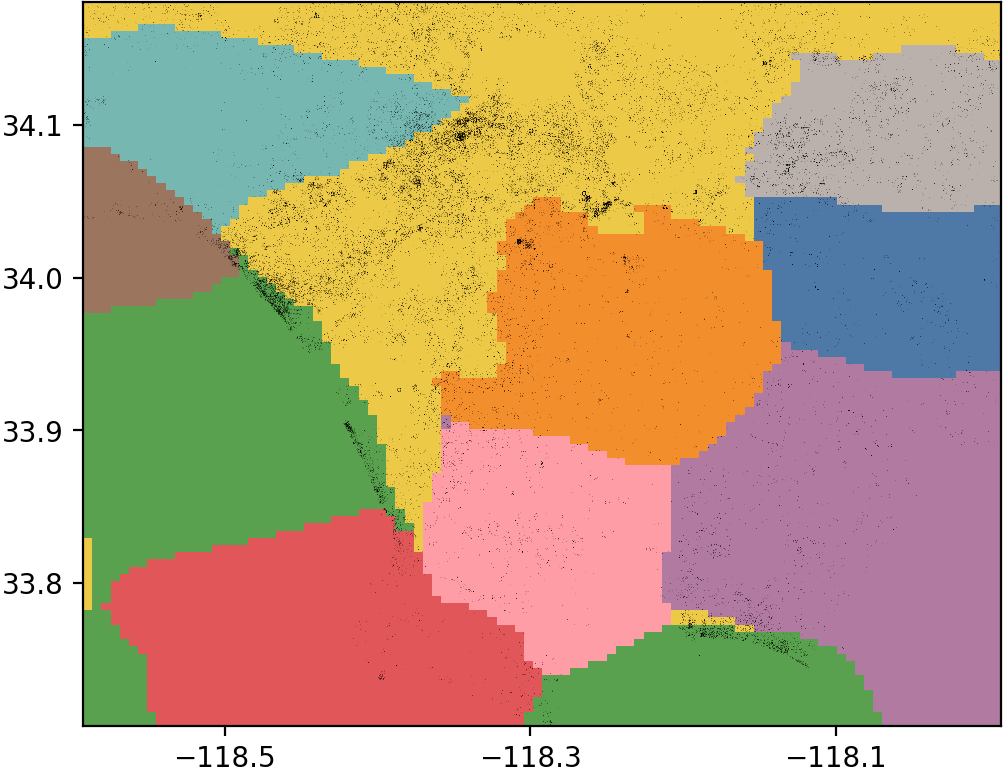}
\\[1pt]

\modellabel{Gemma}
&
\clusterpanel{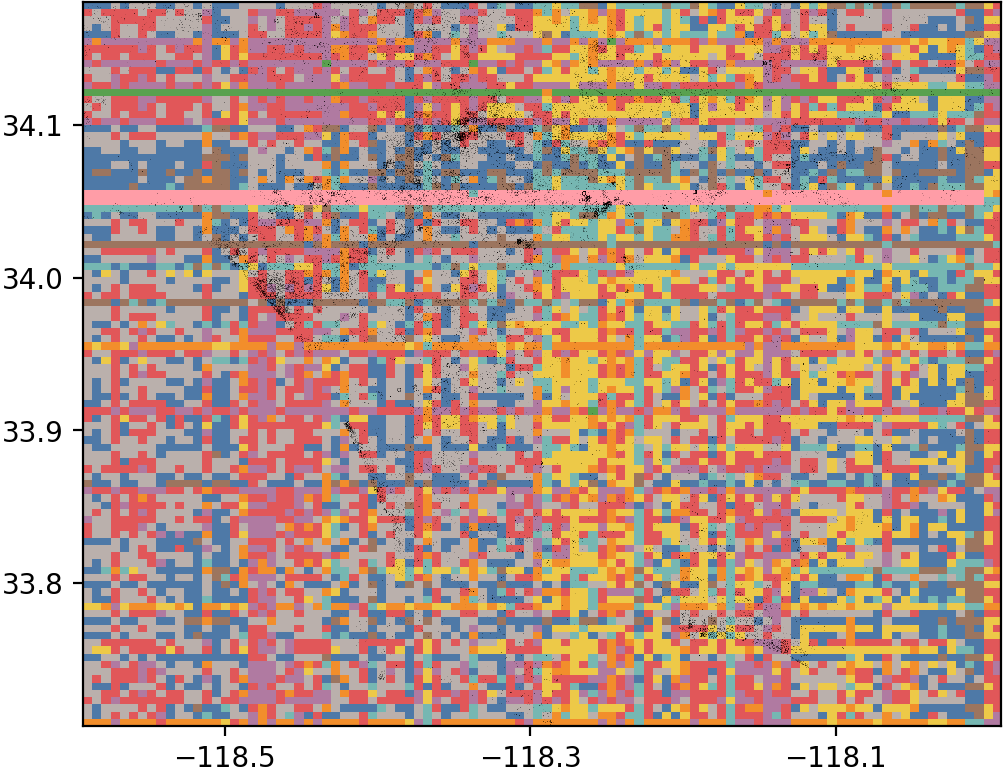}
&
\clusterpanel{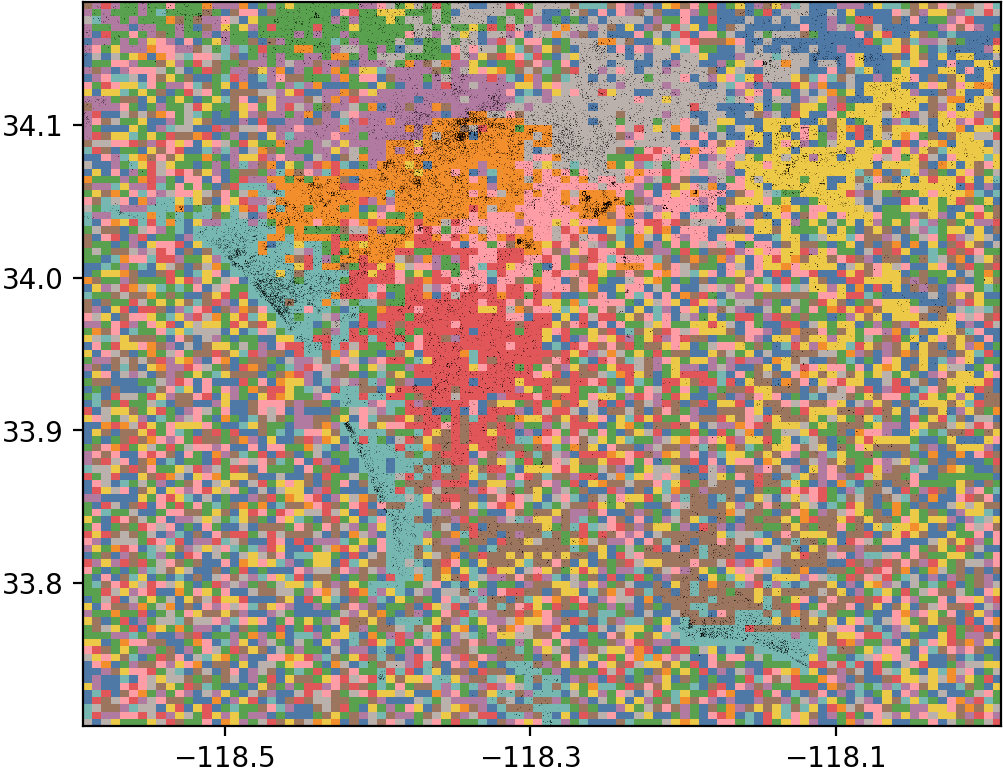}
&
\clusterpanel{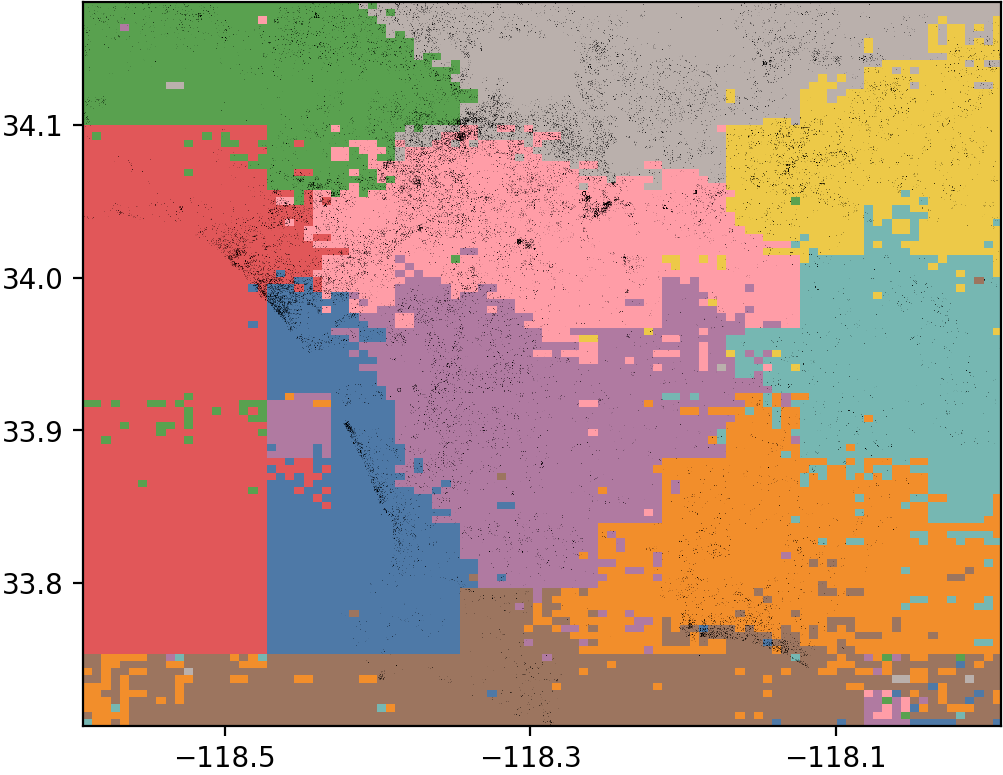}
&
\clusterpanel{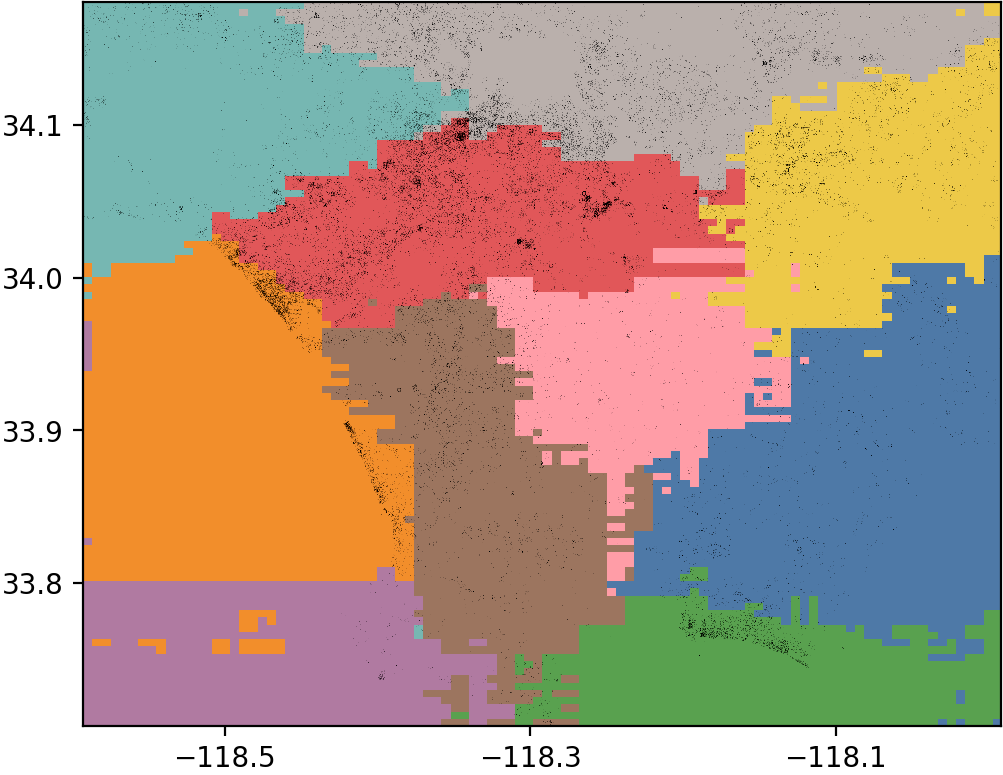}
&
\clusterpanel{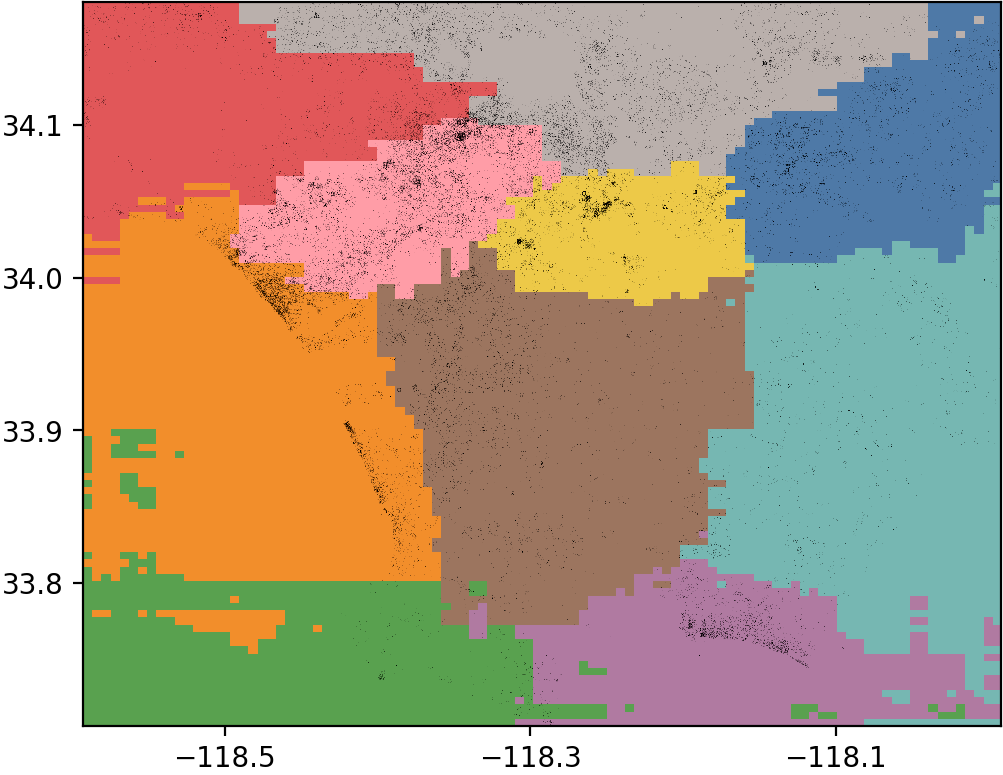}
&
\clusterpanel{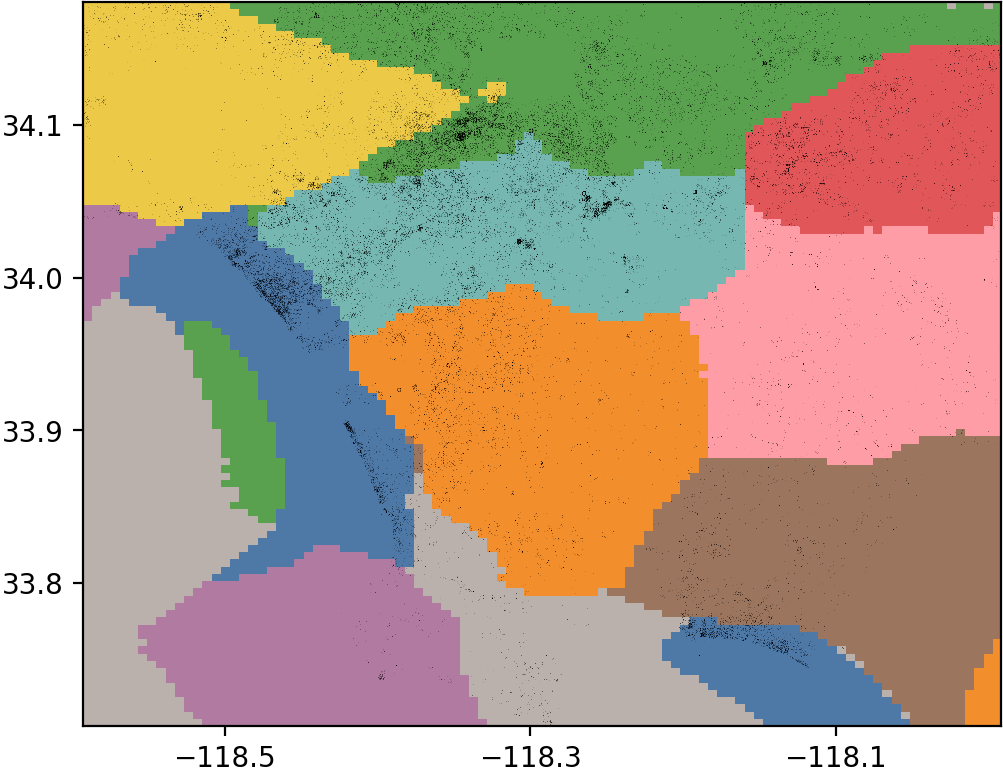}
\end{tabular}

\endgroup
\caption{\textbf{Captioning model's hidden representation cluster maps for the Los Angeles dataset.} Rows correspond to the GPT, Qwen, and Gemma model families, while columns correspond to \textbf{NAIVE}, \textbf{GRID}, \textbf{HASH}, \textbf{TEXT}, \textbf{HYBRID}, and \textbf{LATENT} spatial representations.}
\label{fig:la_cluster_maps}
\end{figure*}

To examine whether LLM hidden states encode geographic structure, we conduct layer-wise probing. For each location description, we extract its generation hidden state from five transformer layers and train an MLP to predict the corresponding latitude and longitude. Further details are provided in Appendix~\ref{appendix:probing}. \textbf{NAIVE} denotes models without task-specific tuning.

As shown in Figure~\ref{fig:interlayer_analysis}, all tuned models achieve lower probing errors than their \textbf{NAIVE} counterparts, indicating that explicit location conditioning produces more spatially structured hidden states. More importantly, \textbf{LATENT} is competitive with and often outperforms \textbf{TEXT} and \textbf{HYBRID} across model families and cities, particularly in intermediate and later layers. This is notable because the probing setup naturally favors \textbf{TEXT} and \textbf{HYBRID}: during captioning training, both \textbf{leak} the exact latitude and longitude alongside their corresponding descriptions, providing a direct shortcut for learning text--coordinate associations. If geographic knowledge were organized merely through such co-occurrence, these representations should produce the most decodable states. Instead, \textbf{LATENT} achieves comparable or better coordinate recoverability without exposing raw coordinates as text, suggesting that continuous encodings encourage LLMs to organize geographic information structurally by forming a continuous latent spatial field.

We further cluster the hidden states of the generation token across query locations and map the assignments back to space. Details and London results are provided in Appendix~\ref{appendix:cluster}. As shown in Figure~\ref{fig:la_cluster_maps}, \textbf{NAIVE} models exhibit little coherent spatial organization, with cluster assignments distributed almost randomly. Despite task-specific tuning, \textbf{GRID} remains spatially fragmented. The \textbf{HASH} and \textbf{TEXT} maps also exhibit abrupt transitions and numerous isolated regions. In contrast, \textbf{LATENT} produces more contiguous regions with clearer boundaries. This difference is especially evident for the text-only GPT-OSS model, where \textbf{TEXT} fails to organize hidden features coherently, whereas \textbf{LATENT} and \textbf{HYBRID} recover continuous regions reflecting geographic heterogeneity. 

Together with the stronger distributional alignment observed in grounding, the probing and clustering results support our hypothesis that continuous spatial encoding promotes a coherent latent spatial field for organizing geographic knowledge and enables spatial generalization.

\section{Conclusion}

In this work, we conceptualize geospatial knowledge in LLMs as a latent spatial field and formulate geospatial awareness through two capabilities: memory, which recovers observed location--knowledge associations, and reasoning, which generalizes such associations to unobserved regions. Based on this formulation, we introduce spatially explicit LLMs that incorporate continuous location encodings into LLM training. Through comprehensive experiments on two newly constructed benchmarks, we demonstrate that continuous spatial representations enable models to generalize more effectively across geographic space and generate useful information that benefits downstream tasks, suggesting that textual and continuous representations are complementary. Our probing and clustering analyses further show that LLMs develop spatially organized internal representations of geographic knowledge. These findings support our hypothesis that LLMs can organize geographic knowledge as a latent spatial field rather than relying solely on textual location--knowledge correlations. Together, our findings position continuous spatial representations as a promising foundation for building more geospatially aware and generalizable language models.

\subsection*{AI use statement}

In this work, we used generative AI tools for implement methods, clean and reformat dataset, design or provide feedback on research methodology or experiments.
We have not used generative AI tools for help develop theoretical models or conceptual frameworks, formulate mathematical claims, propose or refine hypotheses, assist with translation, support qualitative And thematic data analysis, interpret results.
And generate synthetic data sets, provide critical ingredients for proving mathematical claims, assist in the writing of proofs, are not applicable to this work.
Additionally, we used generative AI tools for create or edit software code, draft parts of a research paper, brainstorming, edit a research paper to improve readability, propose a title or keywords for a research paper. We have reviewed all AI-assisted work. We
checked LLM-generated code was verified and tested for correctness, LLM-polished paragraphs are carefully checked by all authors. We take responsibility for the final content of this work, including text, claims or artifacts produced with the aid of generative AI.

\subsection*{Reproducibility Statement}

All the datasets involved in the paper are publicly available. The Appendix provides complete descriptions of the data sources, dataset construction, preprocessing, benchmark splits, model training, and evaluation procedures. The two benchmarks introduced in this work and the source code required to reproduce our experiments are included in the supplementary material. The datasets and code will be released publicly on GitHub upon acceptance, and the trained model checkpoints will be released through Hugging Face.


\bibliography{iclr2027_conference}
\bibliographystyle{iclr2027_conference}

\appendix
\section{Appendix}

\subsection{Dataset Construction}
\label{appendix:dataset}

\paragraph{Data source.}
In this research, we exploited a new source of geospatial-textual coupled dataset, the Airbnb listings dataset~\citep{insideairbnb}. Each listing is associated with geographic coordinates and a human-written description provided by its host. These descriptions not only introduce the property, but also contain detailed geographic context, including nearby landmarks, neighborhood characteristics, transportation facilities, walking distances, and travel times. They therefore provide naturally occurring associations between geographic locations and textual descriptions of their surroundings. We independently process listings from Los Angeles and London to construct the two datasets used in our experiments. The following example illustrates the geographic information contained in a typical raw listing description.

\begin{datasetbox}{Example raw Airbnb description}
It's all about location with this stunning 1 Bedroom at a 5 min walk to the Sand! Be 3 blocks away from the nearest Beach. Be steps away from the popular Main Street Santa Monica and all the best restaurants \& bars the neighborhood has to offer. The area is walkable with a close-by convenience \& liquor store, produce market, bike rental shops, Urth Caffé, and ocean-view parks, to name a few. Venice is about 1 mile away, a straight shot down on Main St. Walk, bike, surf, skate and live like a local!
\end{datasetbox}

\paragraph{Geographic-description extraction.}
Raw listing descriptions also contain substantial information unrelated to location, such as room layouts, furniture, and property amenities. We therefore use \texttt{Llama-3.3-70B-Instruct}\footnote{\url{https://huggingface.co/meta-llama/Llama-3.3-70B-Instruct}} to extract only explicit geographic relations. The model is instructed to preserve stated place names, transportation modes; exclude property-internal information; and avoid introducing facts not present in the original description.

\begin{datasetbox}{Prompt for geographic-description extraction}
\textbf{System prompt:}

You are a helpful assistant that extracts location descriptions from Airbnb property overviews.

Focus solely on identifying phrases that describe the property's location, such as proximity to landmarks, transportation, or neighborhood features.

Ignore any information about the property's amenities, host details, or other non-location content.

Your output should be a JSON object with a single key \texttt{"locations"}, which contains a list of concise location phrases. For example:

\begin{quote}
\ttfamily\footnotesize
\{"locations": ["near the beach",\\
\phantom{\{"locations": [}"10-minute walk to the metro",\\
\phantom{\{"locations": [}"close to the London Eye",\\
\phantom{\{"locations": [}"in Kennington"]\}
\end{quote}

Each phrase should be self-contained and directly related to the property's location.

Do not include any explanations, reasoning, or additional text in your response---only the JSON object.

If no location information is present in the overview, return:

\begin{quote}
\ttfamily\footnotesize
\{"locations": []\}
\end{quote}

\medskip
\textbf{User prompt:}

\begin{quote}
\ttfamily\footnotesize
Airbnb Overview: `\textless DESCRIPTION\textgreater'

Extract location phrases and return the JSON object as specified.
\end{quote}
\end{datasetbox}

The structured output is parsed and converted into independent, self-contained descriptions. Empty or malformed outputs are removed, and repeated descriptions associated with the same listing are deduplicated. The example above produces descriptions such as:

\begin{datasetbox}{Extracted self-contained geographic descriptions}
\begin{itemize}
\setlength{\itemsep}{1pt}
\setlength{\parskip}{0pt}
\setlength{\parsep}{0pt}
\item Five-minute walk from the beach.
\item Three blocks from the nearest beach.
\item Steps away from Main Street in Santa Monica.
\item Restaurants and bars are within walking distance.
\item A convenience store is nearby.
\item Venice is approximately one mile away via Main Street.
\end{itemize}
\end{datasetbox}

For a listing at location $\mathbf{x}_{i}$, each extracted description $y_{i,j}$ forms an independent geo-text pair $(\mathbf{x}_{i},y_{i,j})$. This procedure transforms the original long-form listings into concise location--description associations suitable for geospatial captioning and grounding. Importantly, the extraction model is used only to filter and restructure information explicitly contained in each listing; it is not asked to generate additional geographic knowledge.

\paragraph{Spatial train--test split.}
After preprocessing, the Los Angeles and London datasets contain $226{,}250$ and $220{,}183$ independent location--description pairs, respectively. To evaluate whether an LLM can infer knowledge in spatially unobserved regions, we use a spatial block split rather than randomly dividing individual samples. Specifically, we partition each geographic domain into an $N \times N$ grid. We randomly select $20\%$ of the grid cells as held-out test regions and use the remaining $80\%$ for training. Each location--description pair is assigned to the training or test set according to the cell containing its coordinate. Consequently, this procedure yields spatially separated training and test sets, with the training data exhibiting substantial heterogeneity and sparsity in its geographic distribution.

\begin{figure*}[t]
    \centering
    \begin{subfigure}[b]{0.49\columnwidth}
        \centering
        \includegraphics[width=\linewidth]{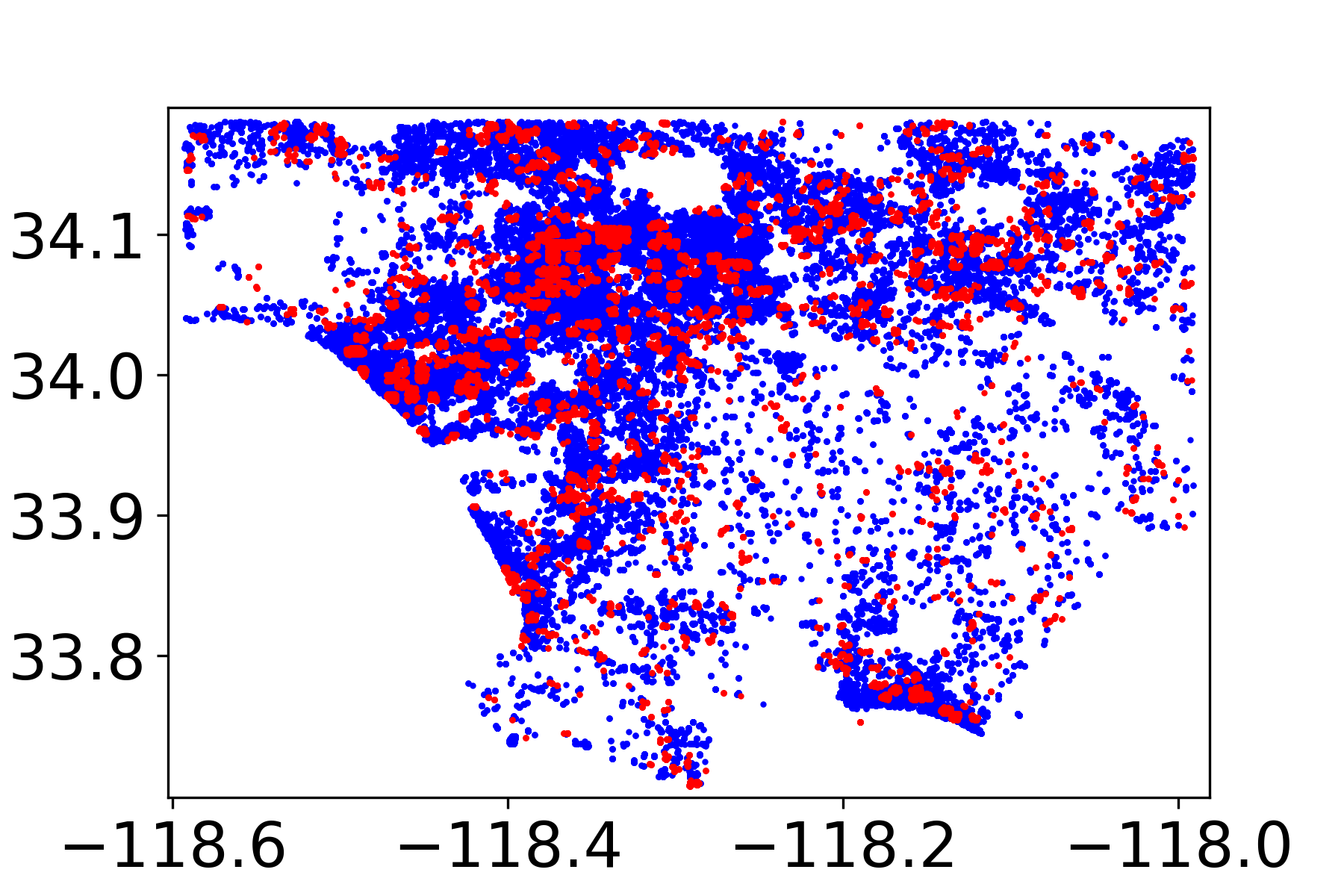}
        \caption{Los Angeles}
        \label{fig:la_dataset}
    \end{subfigure}
    \hfill
    \begin{subfigure}[b]{0.49\columnwidth}
        \centering
        \includegraphics[width=\linewidth]{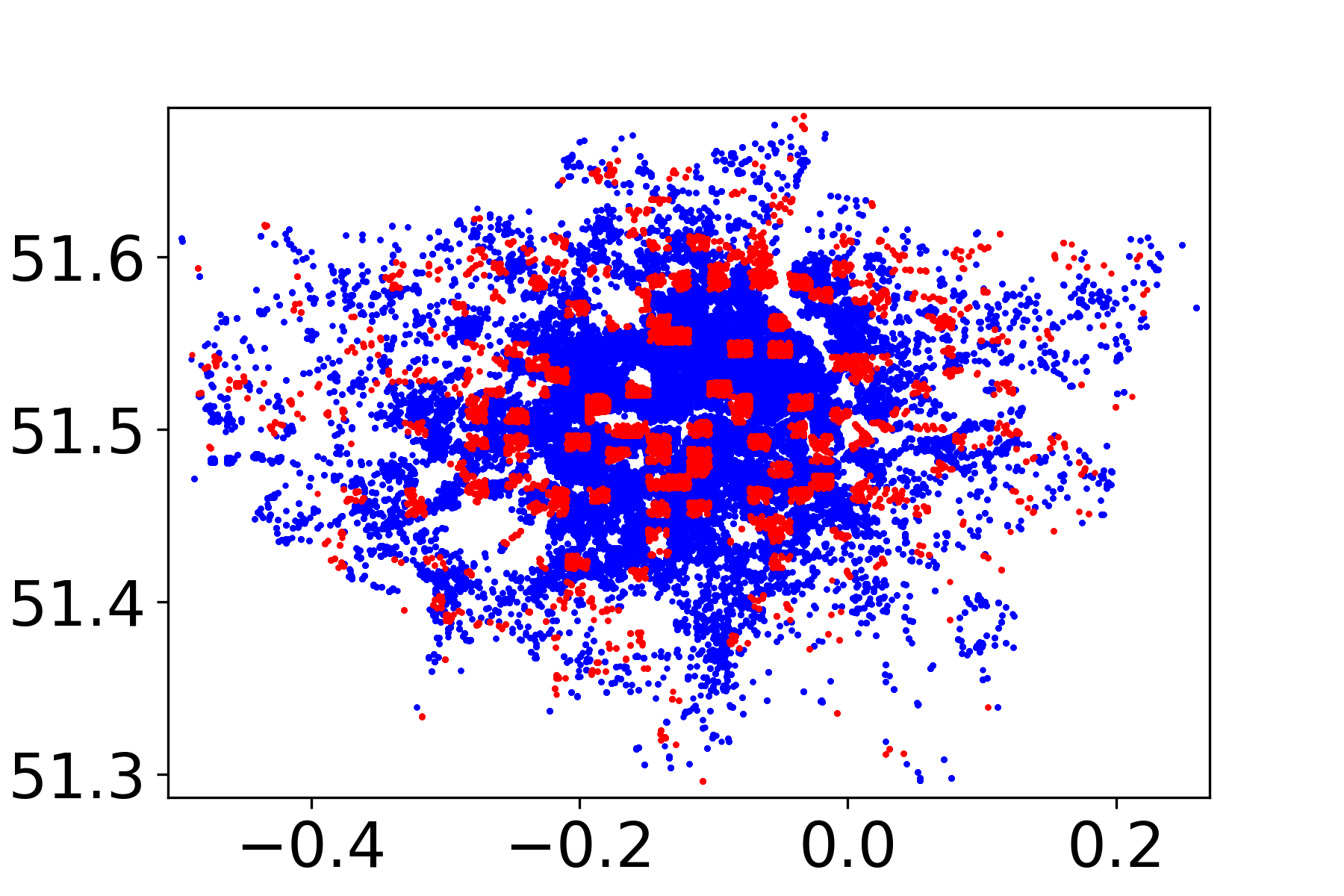}
        \caption{London}
        \label{fig:london_dataset}
    \end{subfigure}
    \caption{\textbf{Spatial distribution of the training and test samples.} Training samples colored in blue, testing samples colored in red.}
    \label{fig:dataset_distribution}
\end{figure*}

To evaluate reasoning across spatial gaps of different scales, we set $N_{\mathrm{LA}}=100$ and $N_{\mathrm{London}}=50$, corresponding to cell sizes of approximately $555.3 \times 526.6$ meters and $1{,}119.3 \times 1{,}260.1$ meters, respectively. The coarser London partition creates larger data-sparse regions, requiring models to generalize across wider spatial gaps. The resulting Los Angeles split contains $179{,}801$ training samples and $46{,}449$ test samples, while the London split contains $177{,}511$ training samples and $42{,}672$ test samples. Figure~\ref{fig:dataset_distribution} visualizes these distributions.

\subsection{Geospatial Captioning and Grounding Benchmarks}
\label{appendix:benchmark_construction}

Based on the spatially partitioned datasets introduced above, we construct a geospatial captioning benchmark and a geospatial grounding benchmark. Following the notation in Section~\ref{sec:formulation}, we set $\mathcal{V}=\mathcal{Y}$ and write the processed dataset as
\begin{equation}
    \mathcal{D}
    =
    \left\{
        (\mathbf{x}_{i}, y_{i})
    \right\}_{i=1}^{N}.
\end{equation}
where $\mathbf{x}_{i}\in\Omega$ is the listing location and $y_{i}\in\mathcal{Y}$ is an extracted location description. The benchmarks use the same underlying pairs and spatial split but evaluate different directions of geospatial understanding.

\subsubsection{Geospatial Captioning}
\label{sec:captioning_benchmark}

Geospatial captioning maps a location $\mathbf{x}_{i}$ to a natural-language description $y{i}$. A captioning model estimates $p_{\psi}(y\mid\mathbf{x})$ and generates a description for the queried location. The model is trained using $\mathcal{D}_{\mathrm{train}}$ and evaluated at locations in the testing regions $\mathcal{H}$. Because no location--description pairs from $\mathcal{H}$ are available during training, performance in these regions measures whether the model can infer geographic knowledge beyond its observed spatial support.

\subsubsection{Point-Level Geospatial Grounding}
\label{sec:point_grounding_benchmark}

Point-level grounding reverses the captioning direction. Given an individual description $y_{i}$, the grounding model estimates $q_{\phi}(\mathbf{x}\mid y_{i})$ over the geographic domain and retrieves the associated listing location $\mathbf{x}_{i}$. Each query therefore has one point-level target. This benchmark evaluates fine-grained text-to-location correspondence by measuring whether $\mathbf{x}_{i}$ falls within the highest-scoring $1\%$ or $5\%$ of spatial candidates.

\subsubsection{Region-Level Geospatial Grounding}
\label{sec:region_grounding_benchmark}

Many location descriptions refer to an area rather than a unique point. Moreover, free-text descriptions are highly redundant: phrases such as `near the beach'' and `close to the beach'' may express the same geographic relation. Semantic similarity alone, however, does not imply geographic specificity. For example, ``near restaurants'' may be semantically coherent but distributed throughout an entire city. We therefore construct region-level grounding labels in two stages: semantic clustering groups paraphrastic descriptions, while spatial filtering retains only clusters whose occurrences form a localized region.

\paragraph{Semantic clustering of location descriptions.}
\label{sec:semantic_grounding_clusters}

Let $\mathcal{U}\subseteq\mathcal{Y}$ be the set of unique descriptions after exact duplicates have been collapsed, and let $n(u)$ denote the number of samples containing $u\in\mathcal{U}$. We encode each unique description using a pretrained sentence encoder $E_{\mathrm{text}}$, specifically Sentence-BERT with \texttt{all-MiniLM-L6-v2}\footnote{\url{https://huggingface.co/sentence-transformers/all-MiniLM-L6-v2}}, and normalize its representation as
\begin{equation}
\mathbf{z}(u)
=
\frac{
E_{\mathrm{text}}(u)
}{
\left|
E_{\mathrm{text}}(u)
\right|_{2}
}.
\end{equation}
Cosine similarity $s$ and distance $d$ are then defined as
\begin{equation}
s(u,u')
=
\mathbf{z}(u)^{\top}\mathbf{z}(u'),
\qquad
d(u,u')
=
1-s(u,u').
\end{equation}

Descriptions in $\mathcal{U}$ are grouped using average-linkage agglomerative clustering with cosine-similarity threshold $\tau=0.95$, or equivalently cosine-distance threshold $1-\tau$. The number of clusters is determined by this threshold rather than specified in advance.

\paragraph{Assignment of infrequent descriptions.}
\label{sec:rare_grounding_assignment}

Let ${C_{1},\ldots,C_{J}}$ denote the clusters constructed from $\mathcal{U}$. For each cluster $C_{j}$, we compute the normalized centroid
\begin{equation}
\boldsymbol{\mu}_{j}
=
\frac{
\displaystyle
\sum_{u\in C_{j}}
\mathbf{z}(u)
}{
\displaystyle
\left|
\sum_{u\in C_{j}}
\mathbf{z}(u)
\right|_{2}
}.
\end{equation}
For each $u\in\mathcal{U}$, its closest centroid is
\begin{equation}
j^{\star}(u)
=
\underset{1\leq j\leq J}{\arg\max}
\;
\mathbf{z}(u)^{\top}
\boldsymbol{\mu}_{j}.
\end{equation}
The description is assigned to this cluster only when it satisfies the same similarity threshold:
\begin{equation}
    \operatorname{Assign}(u)
    =
    \begin{cases}
        C_{j^{\star}(u)},
        &
        \text{if }
        \mathbf{z}(u)^{\top}
        \boldsymbol{\mu}_{j^{\star}(u)}
        \geq
        \tau,
        \\[2mm]
        \{u\},
        &
        \text{otherwise}.
    \end{cases}
\end{equation}
An unmatched description therefore forms a singleton cluster. Let $\mathcal{C}$ denote the resulting collection of semantic clusters. Each cluster $C\in\mathcal{C}$ is assigned a canonical description equal to its most frequent member:
\begin{equation}
y_{C}
=
\underset{u\in C}{\arg\max}
\;
n(u).
\end{equation}

\paragraph{Spatial compactness filtering.}
\label{sec:spatial_grounding_filter}

As discussed above, descriptions such as ``near restaurants'' may occur throughout an entire city and therefore lack a distinguishable geographic region for grounding. To remove such spatially diffuse queries, we apply a spatial compactness filter that measures how tightly the locations associated with each grounding label are clustered.

For each cluster $C\in\mathcal{C}$, define the indices of its associated records as
\begin{equation}
    \mathcal{I}_{C}
    =
    \left\{
        i \in \{1,\ldots,N\} : y_i \in C
    \right\},
    \qquad
    n_C
    =
    \lvert \mathcal{I}_C \rvert.
\end{equation}

We first retain only clusters satisfying
\begin{equation}
    n_{\min}
    \leq
    n_C
    \leq
    n_{\max}.
\end{equation}
where $n_{\min}=40$ and $n_{\max}=500$. The lower bound removes clusters with insufficient observations to define a reliable region, while the upper bound removes broad, city-level descriptions.

For each remaining cluster, we compute the haversine distance $d_{\mathrm{hav}}$ from every location to its nearest distinct neighbor within the same cluster. For $i\in\mathcal{I}_{C}$, this distance is
\begin{equation}
d_{C}^{\mathrm{NN}}(i)
=
\underset{
\substack{
j\in\mathcal{I}_{C}\
j\neq i
}
}{\min}
\;
d_{\mathrm{hav}}
\left(
\mathbf{x}_{i},
\mathbf{x}_{j}
\right),
\end{equation}
where the haversine distance is computed using Earth radius $R=6{,}371 \mathrm{km}$. We summarize the within-cluster distances as
\begin{equation}
\overline{d}_{C}
=
\frac{1}{n_{C}}
\sum_{i\in\mathcal{I}_{C}}
d_{C}^{\mathrm{NN}}(i),
\qquad
\widetilde{d}_{C}
=
\underset{
i\in\mathcal{I}_{C}
}{\operatorname{median}}
\;
d_{C}^{\mathrm{NN}}(i).
\end{equation}
A cluster is retained only when
\begin{equation}
\overline{d}_{C}
\leq
d_{\mathrm{mean}}
\qquad
\text{and}
\qquad
\widetilde{d}_{C}
\leq
d_{\mathrm{med}},
\end{equation}
where $d_{\mathrm{mean}}=500 \mathrm{m}$ and $d_{\mathrm{med}}=300 \mathrm{m}$.

\paragraph{Final region-level labels.}
\label{sec:final_region_labels}

Each surviving cluster defines one region-level query. Its label consists of the canonical description $y_{C}$ and the locations of all records assigned to the cluster:
\begin{equation}
    \left(
        y_C,
        \left\{
            \mathbf{x}_i : i \in \mathcal{I}_C
        \right\}
    \right).
\end{equation}

\begin{figure*}[h]
\centering
\begingroup
\setlength{\tabcolsep}{1pt}
\renewcommand{\arraystretch}{0}

\newcommand{\groundpanel}[1]{%
    \includegraphics[
        width=0.327\textwidth
    ]{#1}%
}

\newcommand{\groundlabel}[1]{%
    \parbox[t]{0.327\textwidth}{%
        \centering
        \fontsize{8pt}{9pt}\selectfont
        \textbf{#1}%
    }%
}

\begin{tabular}{@{}ccc@{}}
\groundpanel{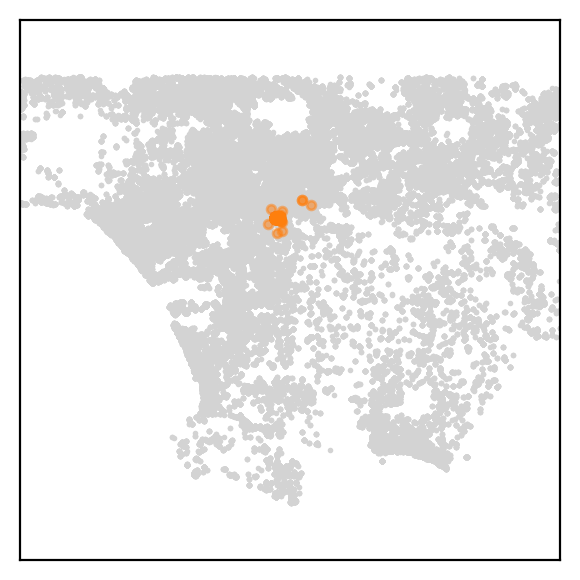}
&
\groundpanel{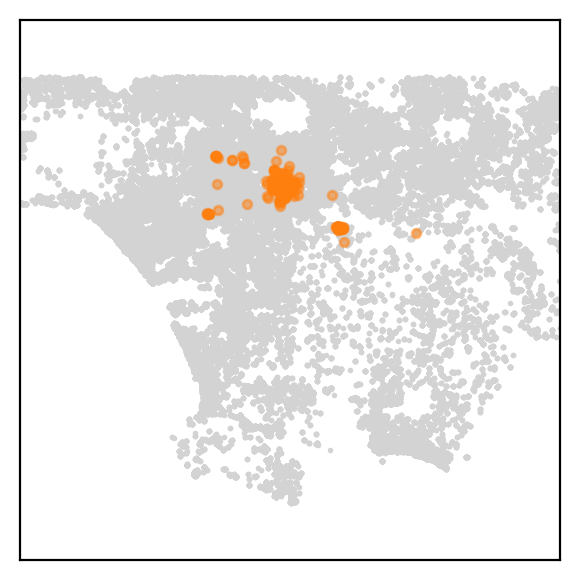}
&
\groundpanel{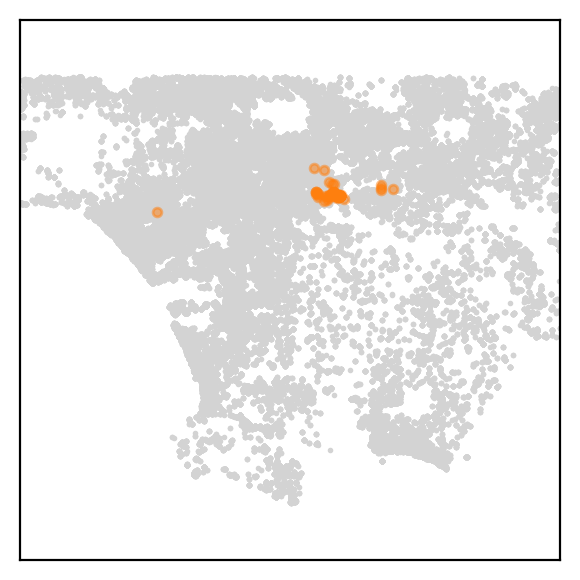}
\\[-1pt]
\groundlabel{5 mins to USC}
&
\groundlabel{In Koreatown}
&
\groundlabel{In Little Tokyo}
\\[2pt]

\groundpanel{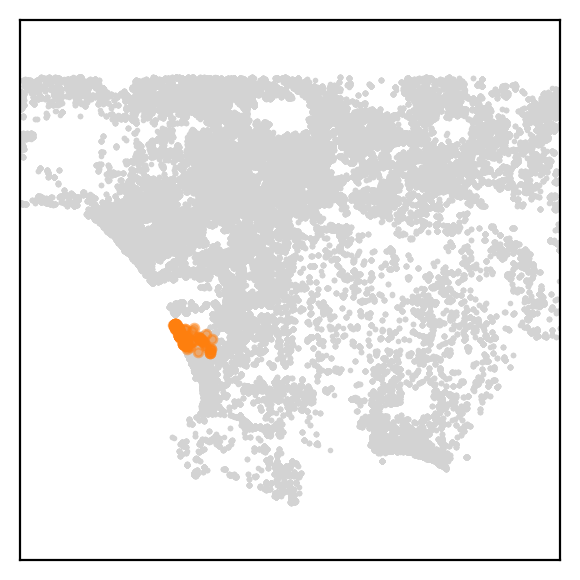}
&
\groundpanel{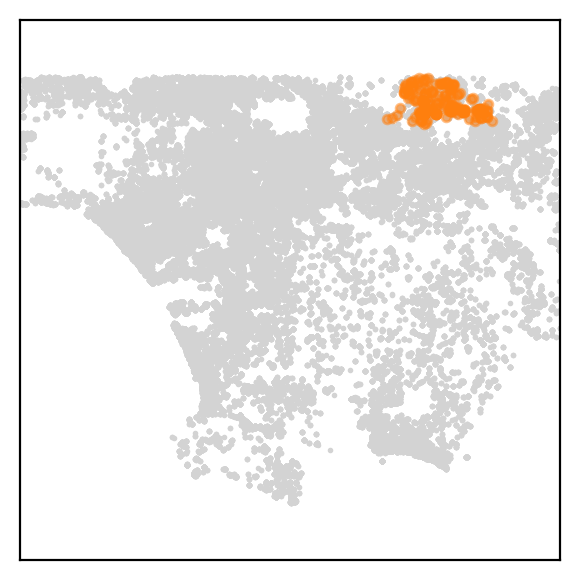}
&
\groundpanel{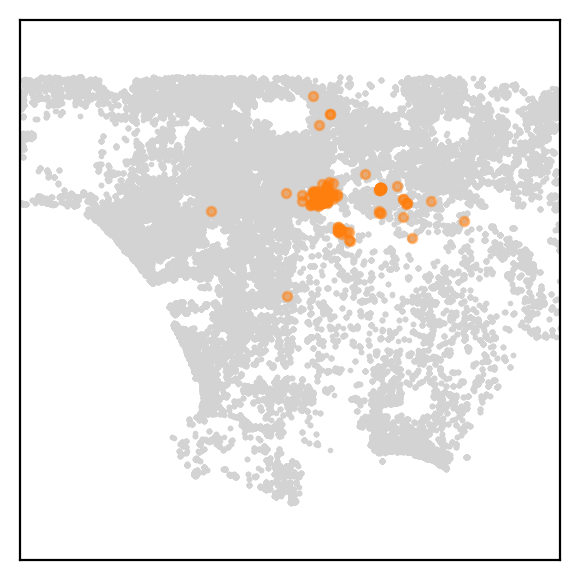}
\\[-1pt]
\groundlabel{In Manhattan Beach}
&
\groundlabel{In Pasadena}
&
\groundlabel{In the heart of Downtown Los Angeles}
\\[2pt]

\groundpanel{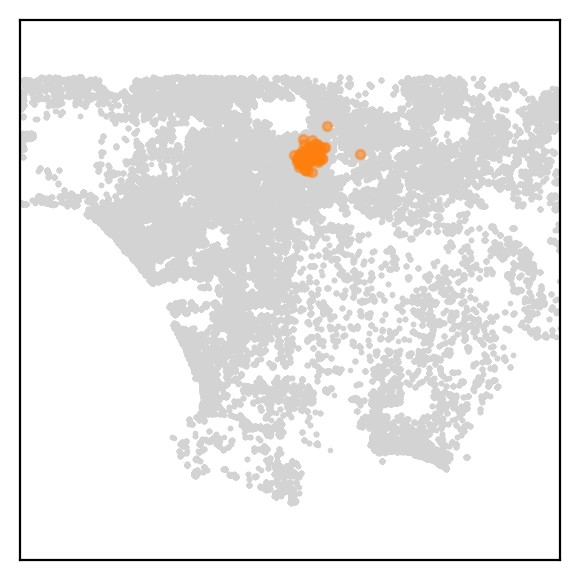}
&
\groundpanel{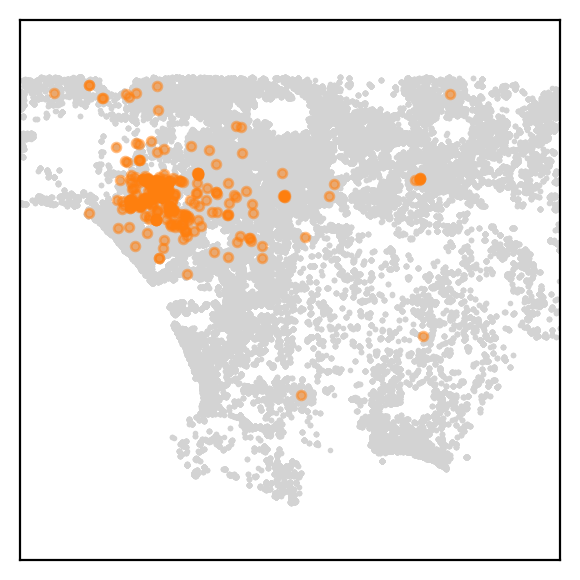}
&
\groundpanel{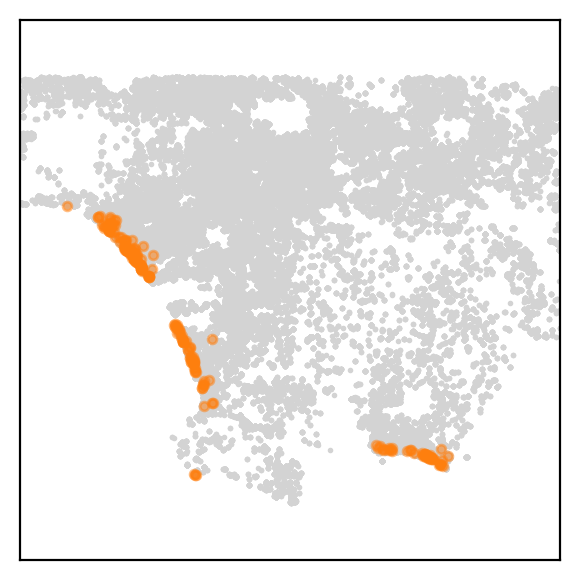}
\\[-1pt]
\groundlabel{Near Silver Lake Reservoir}
&
\groundlabel{Near UCLA}
&
\groundlabel{Steps from the beach}
\end{tabular}

\endgroup
\vspace{-1mm}
\caption{\textbf{Examples from the Los Angeles geospatial grounding benchmark.} Orange points denote target locations and gray points denote the underlying data distribution; labels give the ground-truth descriptions.}
\label{fig:la_grounding_demos}
\vspace{-2mm}
\end{figure*}

\begin{figure*}[t]
\centering
\begingroup
\setlength{\tabcolsep}{1pt}
\renewcommand{\arraystretch}{0}

\newcommand{\groundpanel}[1]{%
    \includegraphics[
        width=0.327\textwidth
    ]{#1}%
}

\newcommand{\groundlabel}[1]{%
    \parbox[t]{0.327\textwidth}{%
        \centering
        \fontsize{8pt}{9pt}\selectfont
        \textbf{#1}%
    }%
}

\begin{tabular}{@{}ccc@{}}
\groundpanel{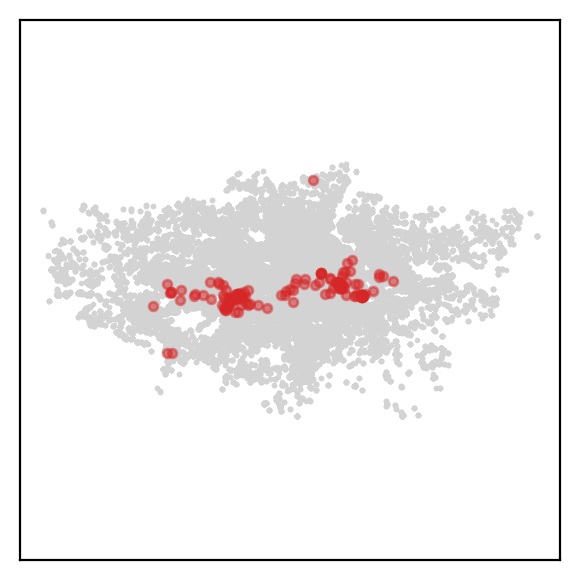}
&
\groundpanel{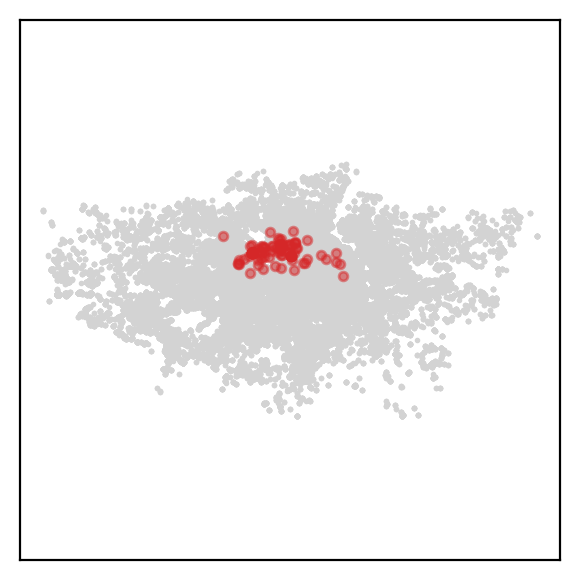}
&
\groundpanel{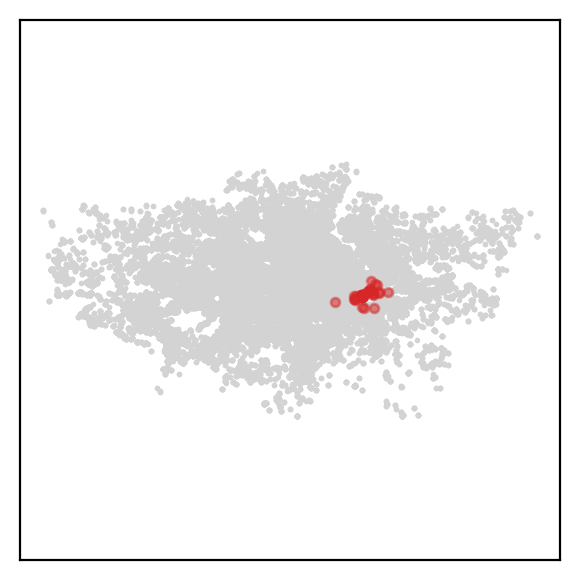}
\\[-1pt]
\groundlabel{Along the river}
&
\groundlabel{Close to Camden}
&
\groundlabel{In Greenwich Park}
\\[2pt]

\groundpanel{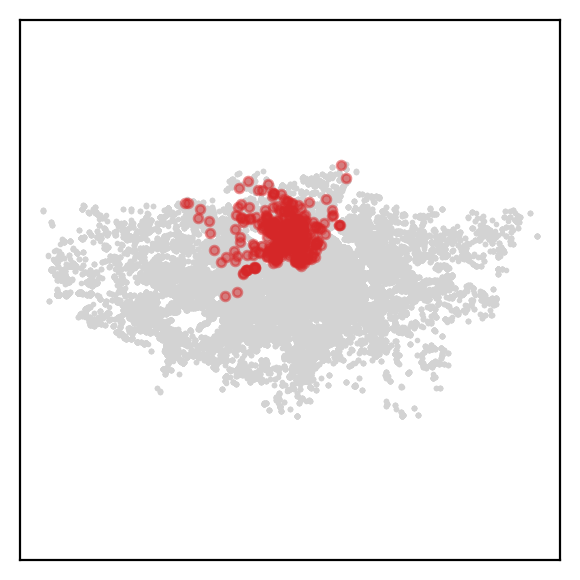}
&
\groundpanel{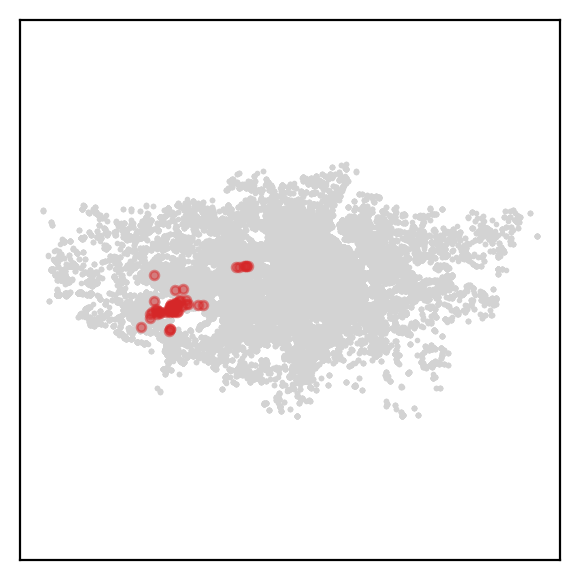}
&
\groundpanel{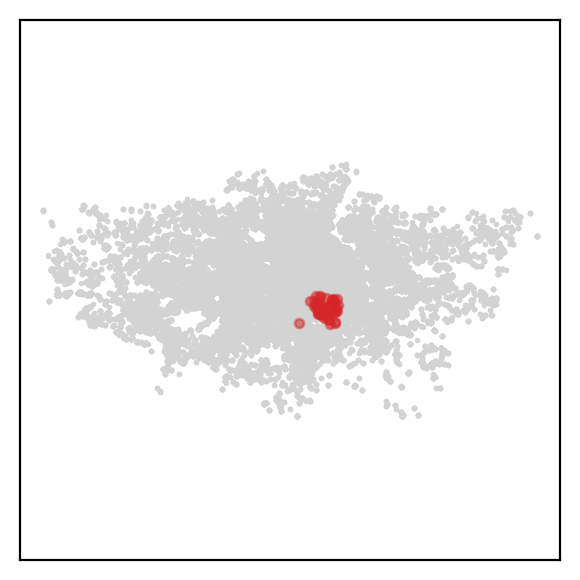}
\\[-1pt]
\groundlabel{In North London}
&
\groundlabel{In Richmond}
&
\groundlabel{Near Peckham Rye}
\\[2pt]

\groundpanel{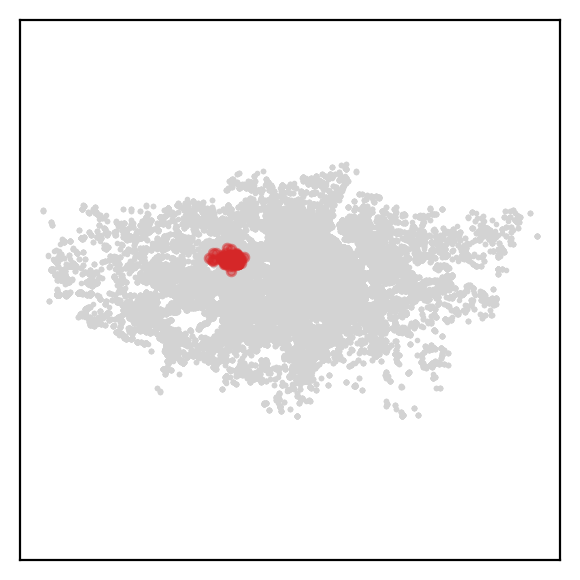}
&
\groundpanel{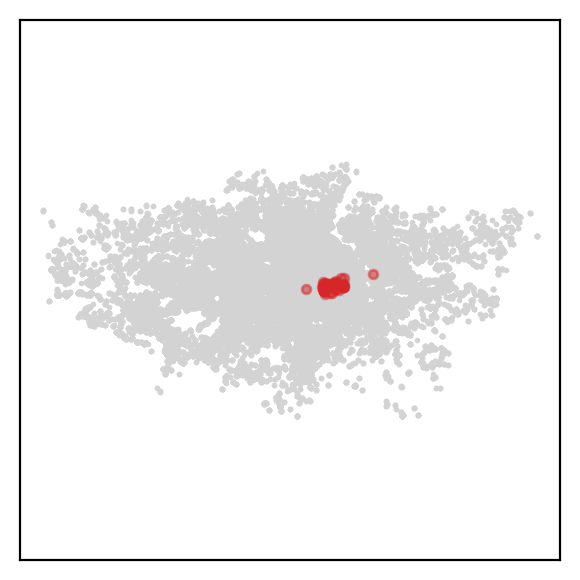}
&
\groundpanel{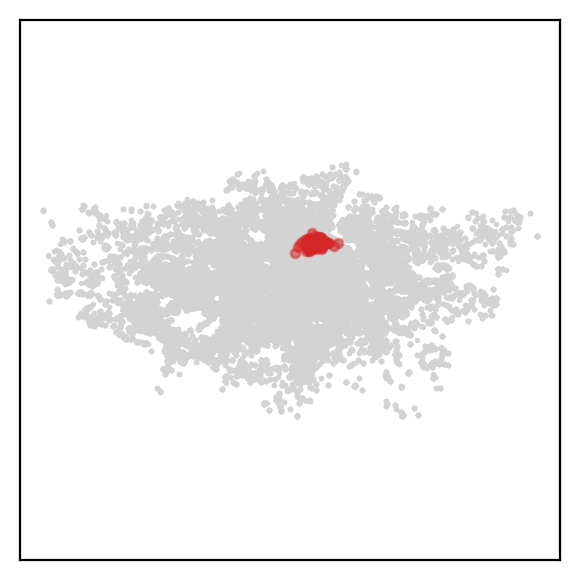}
\\[-1pt]
\groundlabel{Near Queen's Park}
&
\groundlabel{Near Southwark Park}
&
\groundlabel{Near Stoke Newington Church Street}
\end{tabular}

\endgroup
\vspace{-1mm}
\caption{\textbf{Examples from the London geospatial grounding benchmark.} Red points denote target locations and gray points denote the underlying data distribution; labels give the ground-truth descriptions.}
\label{fig:london_grounding_demos}
\vspace{-2mm}
\end{figure*}

After this process, $334$ and $625$ region-level grounding queries remain in the Los Angeles and London datasets, respectively. On average, each query is grounded by $52.61$ unique locations in Los Angeles and $41.77$ in London. A few sample queries and their corresponding samples are presented in Figure~\ref{fig:la_grounding_demos} and Figure~\ref{fig:london_grounding_demos}.

\begin{algorithm}[t]
\caption{Construction of semantically and spatially grounded region labels}
\label{alg:grounding-labels}
\begin{algorithmic}[1]

\REQUIRE Dataset
$\mathcal{D}=\{(\mathbf{x}_i,y_i)\}_{i=1}^{N}$;
encoder $E_{\mathrm{text}}$;
head size $K$;
similarity threshold $\tau$;
support bounds $[n_{\min},n_{\max}]$;
spatial thresholds $d_{\mathrm{mean}}$ and $d_{\mathrm{med}}$

\STATE Form the unique-description set $\mathcal{U}$ with frequencies
$n(u)$; compute normalized embeddings $\mathbf{z}(u)$

\STATE Split $\mathcal{U}$ into the frequent set
$\mathcal{U}_{\mathrm{freq}}$ containing the top-$K$ descriptions and
$\mathcal{U}_{\mathrm{rare}}
=
\mathcal{U}\setminus\mathcal{U}_{\mathrm{freq}}$

\STATE Cluster $\mathcal{U}_{\mathrm{freq}}$ using average-linkage
agglomerative clustering with cosine-distance threshold $1-\tau$,
producing $\{C_1,\ldots,C_J\}$

\STATE Compute the normalized centroid $\boldsymbol{\mu}_j$ of each
cluster $C_j$

\FOR{$u\in\mathcal{U}_{\mathrm{rare}}$}
    \STATE Find
    $j^{\star}(u)
    =
    \arg\max_j
    \mathbf{z}(u)^{\top}\boldsymbol{\mu}_j$

    \STATE Assign $u$ to $C_{j^{\star}(u)}$ if
    $\mathbf{z}(u)^{\top}
    \boldsymbol{\mu}_{j^{\star}(u)}
    \geq\tau$; otherwise, create a singleton cluster
\ENDFOR

\STATE Let $\mathcal{C}$ denote the resulting semantic clusters and set
$y_C=\arg\max_{u\in C}n(u)$ for each $C\in\mathcal{C}$

\STATE Define
$\mathcal{I}_C
=
\{i\in\{1,\ldots,N\}:y_i\in C\}$
and
$n_C
=
\lvert\mathcal{I}_C\rvert$

\STATE Discard clusters for which
$n_C\notin[n_{\min},n_{\max}]$

\STATE For each remaining cluster, compute the within-cluster haversine
nearest-neighbor distances $d_C^{\mathrm{NN}}(i)$

\STATE Retain $C$ if
$\overline{d}_C\leq d_{\mathrm{mean}}$
and
$\widetilde{d}_C\leq d_{\mathrm{med}}$

\RETURN Retained labels
$\left\{
    \left(
        y_C,
        \{\mathbf{x}_i:i\in\mathcal{I}_C\}
    \right)
\right\}$

\end{algorithmic}
\end{algorithm}

Together, these benchmarks evaluate complementary aspects of geospatial understanding. Captioning tests inference from location to geographic knowledge, point-level grounding evaluates precise text-to-location retrieval, and region-level grounding evaluates the recovery of localized spatial distributions. Their shared spatial split also allows captions generated in $\mathcal{H}$ to augment a separately trained grounding model, providing an extrinsic evaluation of whether the inferred descriptions contain useful geographic knowledge.

\textbf{Two benchmarks are included in the supplementary material and will be publicly released.}

\subsection{Evaluation Metrics}
\label{appendix:metric}

\subsubsection{Metrics for Geospatial Captioning}
\label{appendix:metric_captioning}

Geospatial captioning is inherently open-ended: the same location can be correctly described through different landmarks, neighborhoods, transportation options, or spatial relations. The available human-written descriptions should therefore be treated as valid references rather than as unique ground-truth answers. We consequently evaluate captioning through many-to-many alignment between the generated and reference descriptions associated with each location. For each evaluated location $\mathbf{x}_i$, let
\begin{equation}
    \widehat{\mathcal{Y}}_i
    =
    \left\{
        \widehat{y}_{i,r}
    \right\}_{r=1}^{N_g},
    \qquad
    \mathcal{Y}_i
    =
    \left\{
        y_{i,j}
    \right\}_{j=1}^{M_i},
\end{equation}
denote its generated and reference description sets, respectively. We generate $N_g=4$ descriptions for every location, while $M_i$ is the number of available references for $\mathbf{x}_i$. To augment the grounding model, each generated description is paired with its corresponding location $\mathbf{x}_i$. Using the same augmentation locations across methods isolates performance differences arising from the generated descriptions rather than from location selection.

Let $s(\widehat{y},y)$ denote a pairwise similarity function between a generated description and a reference. We instantiate $s$ separately using BERTScore~\citep{bertscore} (\texttt{sentence-transformers/all-MiniLM-L6-v2}\footnote{\url{https://huggingface.co/sentence-transformers/all-MiniLM-L6-v2}}) F1 and ROUGE-L~\citep{rouge} F1, two commonly used metrics in open-ended text generation. These metrics capture complementary aspects of caption quality. BERTScore is tolerant of semantic paraphrases, whereas ROUGE-L measures lexical and sequence-level agreement. For example, `near the beach' and 'close to the coast' may express similar geographic meaning despite limited token overlap, while exact landmark names, neighborhood names, and distance expressions remain important lexical evidence. For each metric, we construct the similarity matrix
\begin{equation}
    S_{i,r,j}
    =
    s
    \left(
        \widehat{y}_{i,r},
        y_{i,j}
    \right).
\end{equation}

The set-level precision measures whether each generated description is supported by at least one reference. Each generated description is matched to its most similar reference, and the resulting similarities are averaged:
\begin{equation}
    P_i
    =
    \frac{1}{N_g}
    \sum_{r=1}^{N_g}
    \max_{1\leq j\leq M_i}
    S_{i,r,j}.
\end{equation}
Conversely, set-level recall measures how well the generated descriptions cover the available references. Each reference is matched to its most similar generated description:
\begin{equation}
    R_i
    =
    \frac{1}{M_i}
    \sum_{j=1}^{M_i}
    \max_{1\leq r\leq N_g}
    S_{i,r,j}.
\end{equation}
Their harmonic mean gives the alignment F1 score for location $\mathbf{x}_i$:
\begin{equation}
    F_{1,i}
    =
    \frac{
        2P_iR_i
    }{
        P_i+R_i+\varepsilon
    },
\end{equation}
where $\varepsilon=10^{-12}$ ensures numerical stability.

Let $\mathcal{I}_{\mathrm{eval}}$ denote the indices of the evaluated locations. We first compute the three scores independently for each location and then average them across the evaluation set:
\begin{equation}
    \overline{P}
    =
    \frac{1}{
        \left|
            \mathcal{I}_{\mathrm{eval}}
        \right|
    }
    \sum_{i\in\mathcal{I}_{\mathrm{eval}}}
    P_i,
    \qquad
    \overline{R}
    =
    \frac{1}{
        \left|
            \mathcal{I}_{\mathrm{eval}}
        \right|
    }
    \sum_{i\in\mathcal{I}_{\mathrm{eval}}}
    R_i,
\end{equation}
and
\begin{equation}
    \overline{F}_1
    =
    \frac{1}{
        \left|
            \mathcal{I}_{\mathrm{eval}}
        \right|
    }
    \sum_{i\in\mathcal{I}_{\mathrm{eval}}}
    F_{1,i}.
\end{equation}
This aggregation is applied independently to BERTScore F1 and ROUGE-L F1, producing semantic and lexical set-alignment scores, respectively. The many-to-many evaluation rewards generated descriptions that are supported by the references while also measuring how well the generated set covers their diversity. Nevertheless, it remains a reference-based intrinsic evaluation and cannot by itself establish whether the generated descriptions contain spatially useful knowledge, motivating the extrinsic grounding evaluation described next.

\subsubsection{Metrics for Geospatial Grounding}
\label{appendix:metric_grounding}

\paragraph{Predefined Grid Over Space}
\label{appendix:grounding_grid}

For each city, longitude and latitude are independently rescaled to $[-1,1]$ using the bounding box of the coordinates. We then divide the whole region into a uniform grid $\mathcal{G}$ with resolution $M\times M$. We use $M=100$, giving $|\mathcal{G}|=10{,}000$ grid cells.

For each grid cell $\mathbf{x}\in\mathcal{G}$ and description $y\in\mathcal{Y}$, the grounding model produces a location representation $E_{\phi}^{\mathrm{loc}}(\mathbf{x})$ and a text representation $E_{\phi}^{\mathrm{text}}(y)$. Their cosine similarity is
\begin{equation}
    s_{\phi}(\mathbf{x},y)
    =
    \frac{
        \left\langle
            E_{\phi}^{\mathrm{text}}(y),
            E_{\phi}^{\mathrm{loc}}(\mathbf{x})
        \right\rangle
    }{
        \left\|E_{\phi}^{\mathrm{text}}(y)\right\|_2
        \left\|E_{\phi}^{\mathrm{loc}}(\mathbf{x})\right\|_2
    }.
\end{equation}

For a coverage ratio $\alpha\in(0,1]$, let $k_{\alpha}=\lceil\alpha|\mathcal{G}|\rceil$, and let $s_{\phi}^{[k_{\alpha}]}(y)$ denote the $k_{\alpha}$-th largest value in $\{s_{\phi}(\mathbf{x},y):\mathbf{x}\in\mathcal{G}\}$. The predicted high-similarity region is
\begin{equation}
    \mathcal{R}_{\alpha}(y)
    =
    \left\{
        \mathbf{x}\in\mathcal{G}
        :
        s_{\phi}(\mathbf{x},y)
        \geq
        s_{\phi}^{[k_{\alpha}]}(y)
    \right\}.
\end{equation}
Ties at the threshold are broken deterministically so that $|\mathcal{R}_{\alpha}(y)|=k_{\alpha}$. We denote by $\pi_{\mathcal{G}}(\mathbf{x})$ the grid cell nearest to a continuous coordinate $\mathbf{x}$. All reported results are means over five independent training runs, with standard deviations reported after $\pm$.

\paragraph{Point-Level Grounding}
\label{appendix:point_metrics}

Point-level grounding evaluates whether the location paired with a description falls inside its predicted high-similarity region. Let $\mathcal{I}_{\mathrm{test}}$ denote the indices of the in-bounds test pairs $(\mathbf{x}_i,y_i)\in\mathcal{D}_{\mathrm{h}}$. The hit rate at coverage $\alpha$ is
\begin{equation}
    \operatorname{Hit Rate}@\alpha
    =
    \frac{1}{|\mathcal{I}_{\mathrm{test}}|}
    \sum_{i\in\mathcal{I}_{\mathrm{test}}}
    \mathbf{1}
    \left[
        \pi_{\mathcal{G}}(\mathbf{x}_i)
        \in
        \mathcal{R}_{\alpha}(y_i)
    \right].
\end{equation}
We report $\operatorname{Hit Rate}@1\%$ and $\operatorname{Hit Rate}@5\%$, corresponding to $\alpha=0.01$ and $\alpha=0.05$. Higher values indicate better point-level grounding.

\paragraph{Region-Level Grounding}
\label{appendix:region_metrics}

For a region-level grounding label $C\in\mathcal{C}$, let $y_C$ be its canonical description and let $\mathcal{I}_C$ index its $n_C=|\mathcal{I}_C|$ support locations. Region-level grounding is evaluated using both coverage rate and alignment between the complete predicted and empirical spatial distributions.

\paragraph{Coverage rate}

The coverage score measures the fraction of support locations captured by the predicted high-similarity region, averaged equally across grounding labels:
\begin{equation}
    \operatorname{Coverage}@\alpha
    =
    \frac{1}{|\mathcal{C}|}
    \sum_{C\in\mathcal{C}}
    \frac{1}{n_C}
    \sum_{i\in\mathcal{I}_C}
    \mathbf{1}
    \left[
        \pi_{\mathcal{G}}(\mathbf{x}_i)
        \in
        \mathcal{R}_{\alpha}(y_C)
    \right].
\end{equation}
We use $\alpha=0.01$ by default. Higher values indicate that the predicted region covers a larger proportion of the locations associated with each description.

\paragraph{Empirical Target Distribution}

We construct a discrete target distribution $p_C$ over $\mathcal{G}$ using isotropic Gaussian kernel density estimation. With bandwidth $\sigma=0.05$ in normalized coordinates,
\begin{equation}
    \widetilde{p}_C(\mathbf{x})
    =
    \sum_{i\in\mathcal{I}_C}
    \exp
    \left(
        -
        \frac{
            \left\|\mathbf{x}-\mathbf{x}_i\right\|_2^2
        }{
            2\sigma^2
        }
    \right),
    \qquad
    \mathbf{x}\in\mathcal{G},
\end{equation}
and
\begin{equation}
    p_C(\mathbf{x})
    =
    \frac{
        \widetilde{p}_C(\mathbf{x})
    }{
        \displaystyle
        \sum_{\mathbf{x}'\in\mathcal{G}}
        \widetilde{p}_C(\mathbf{x}')
    }.
\end{equation}

\paragraph{Predicted Distribution}

To obtain the discrete form of the grounding distribution $q_{\phi}(\mathbf{x}\mid y_C)$ defined in the main text, we floor the cosine-similarity field at a small constant $\varepsilon>0$ and normalize it over $\mathcal{G}$:
\begin{equation}
    \widetilde{q}_{\phi,C}(\mathbf{x})
    =
    \max
    \left\{
        s_{\phi}(\mathbf{x},y_C),
        \varepsilon
    \right\},
\end{equation}
\begin{equation}
    q_{\phi,C}(\mathbf{x})
    =
    \frac{
        \widetilde{q}_{\phi,C}(\mathbf{x})
    }{
        \displaystyle
        \sum_{\mathbf{x}'\in\mathcal{G}}
        \widetilde{q}_{\phi,C}(\mathbf{x}')
    }.
\end{equation}
This normalization preserves relative similarity across the complete geographic domain rather than only within $\mathcal{R}_{\alpha}(y_C)$.

\paragraph{Jensen--Shannon Divergence}

Given the target distribution $p_C$ and predicted distribution $q_{\phi,C}$, we use Jensen--Shannon divergence (JSD) to measure their cell-wise distributional agreement. JSD is symmetric, bounded, and well defined for distributions with partially non-overlapping support, making it suitable for comparing spatial probability fields. Let
\begin{equation}
    m_C(\mathbf{x})
    =
    \frac{1}{2}
    \left(
        p_C(\mathbf{x})
        +
        q_{\phi,C}(\mathbf{x})
    \right).
\end{equation}
For discrete distributions $a$ and $b$ over $\mathcal{G}$, the Kullback--Leibler divergence is
\begin{equation}
    \operatorname{KL}(a\|b)
    =
    \sum_{\mathbf{x}\in\mathcal{G}}
    a(\mathbf{x})
    \log
    \frac{a(\mathbf{x})}{b(\mathbf{x})}.
\end{equation}
The Jensen--Shannon divergence for cluster $C$ is
\begin{equation}
    \operatorname{JSD}_C
    =
    \frac{1}{2}
    \operatorname{KL}(p_C\|m_C)
    +
    \frac{1}{2}
    \operatorname{KL}(q_{\phi,C}\|m_C).
\end{equation}
Using the natural logarithm, JSD is bounded by
\begin{equation}
    0
    \leq
    \operatorname{JSD}_C
    \leq
    \log 2.
\end{equation}
A value of zero indicates identical distributions, while larger values indicate weaker overlap between their probability mass. We report the mean across region-level labels:
\begin{equation}
    \overline{\operatorname{JSD}}
    =
    \frac{1}{|\mathcal{C}|}
    \sum_{C\in\mathcal{C}}
    \operatorname{JSD}_C,
    \qquad
    \widetilde{\operatorname{JSD}}
    =
    \underset{C\in\mathcal{C}}{\operatorname{median}}
    \operatorname{JSD}_C.
\end{equation}
JSD captures differences in the shape and concentration of the two distributions but does not consider the distance between mismatched regions. For example, two predictions displaced by different geographic distances may receive similar JSD values if their overlap with the target is comparable.

\paragraph{Rasterized Wasserstein Distance}

We additionally report Wasserstein distance as a complement to JSD. While JSD primarily measures probability overlap, Wasserstein distance measures how far probability mass must be shifted to align the predicted and target distributions. To match our evaluation implementation, we enumerate the cells of $\mathcal{G}$ in row-major raster order as $\mathbf{x}^{(1)},\ldots,\mathbf{x}^{(|\mathcal{G}|)}$. For any distribution $a$ over $\mathcal{G}$, its cumulative mass along this ordering is
\begin{equation}
    F_a(k)
    =
    \sum_{r=1}^{k}
    a
    \left(
        \mathbf{x}^{(r)}
    \right).
\end{equation}
The rasterized Wasserstein-1 distance for cluster $C$ is
\begin{equation}
    W_{1,C}^{\mathrm{raster}}
    =
    \sum_{k=1}^{|\mathcal{G}|}
    \left|
        F_{p_C}(k)
        -
        F_{q_{\phi,C}}(k)
    \right|.
\end{equation}
We report its mean across region-level labels:
\begin{equation}
    \overline{W}_{1}^{\mathrm{raster}}
    =
    \frac{1}{|\mathcal{C}|}
    \sum_{C\in\mathcal{C}}
    W_{1,C}^{\mathrm{raster}},
    \qquad
    \widetilde{W}_{1}^{\mathrm{raster}}
    =
    \underset{C\in\mathcal{C}}{\operatorname{median}}
    W_{1,C}^{\mathrm{raster}}.
\end{equation}
Lower values indicate that the predicted probability mass is more closely aligned with the target field along the raster ordering.

Together, the two metrics provide complementary information: JSD evaluates whether probability mass is assigned to the same grid cells with a similar distributional shape, whereas rasterized Wasserstein distance measures the extent of distribution between the complete fields. A model must perform well on both metrics to achieve strong overlap without substantially shifting the predicted spatial distribution.

\paragraph{Metric Complementarity}

$\operatorname{Hit Rate}@\alpha$ and $\operatorname{Coverage}@\alpha$ evaluate whether annotated locations occur within a binary high-similarity region. In contrast, $\operatorname{JSD}$ and $W_1^{\mathrm{raster}}$ compare how probability mass is distributed across the complete geographic domain. A model may therefore achieve high coverage while still obtaining poor distributional alignment.

\subsection{Geospatial Captioning Model}
\label{appendix:captioning}

We fine-tune each LLM to approximate the conditional distribution $p_{\psi}(y\mid\mathbf{x})$, where $\mathbf{x}$ is a geographic location and $y$ is a natural-language description of that location. To ensure a controlled comparison, all model--representation combinations use identical instruction content while retaining each model's native chat template and special tokens.

\begin{tcolorbox}[
    colback=gray!3,
    colframe=black!35,
    boxrule=0.5pt,
    arc=1mm,
    left=1.5mm,
    right=1.5mm,
    top=1mm,
    bottom=1mm
]
\small
\textbf{System:} You are a helpful assistant for geo-localization. Always use only the information contained in the provided representation to generate the description.

\vspace{1mm}

\textbf{User:} You are at location \texttt{<GEO>}. Generate a detailed description of the location.
\end{tcolorbox}

\begin{wraptable}{r}{0.55\columnwidth}
\vspace{-1.0em}
\centering
\caption{Training configurations.}
\label{tab:captioning_config}
\vspace{-0.5em}
\begingroup
\fontsize{8.5pt}{9.5pt}\selectfont
\setlength{\tabcolsep}{3pt}
\renewcommand{\arraystretch}{1.05}
\begin{tabularx}{\linewidth}{@{}Xr@{}}
\toprule
Configuration & Value \\
\midrule
Base model & Frozen \\
Token embeddings & Frozen \\
Batch size & $16$ \\
Gradient accumulation & $12$ \\
LoRA learning rate & $1\times10^{-5}$ \\
Geo-module learning rate & $1\times10^{-4}$ \\
Weight decay & $0$ \\
Learning-rate scheduler & Cosine \\
Warmup ratio & $0.01$ \\
Encoder-only epochs & $1$ \\
Maximum gradient norm & $1.0$ \\
Samples per epoch & $10,000$ \\
Training epochs & $100$ \\
\bottomrule
\end{tabularx}
\endgroup
\vspace{-1.0em}
\end{wraptable}

The \texttt{<GEO>} placeholder is replaced by the corresponding location representation. For \textbf{GRID}~\citep{geollama}, we use the same spatial partition as the train--test split, with $N_{\mathrm{LA}}=100$ and $N_{\mathrm{London}}=50$, and replace \texttt{<GEO>} with the identifier of the cell containing the location, such as \texttt{grid\_ID}. For \textbf{HASH}~\citep{reageo}, Geohash, as one of the most typical geospatial indexing method, converts geographic coordinates into a short Base32 string by using a hierarchical grid structure that repeatedly bisects latitude and longitude ranges into smaller sub-intervals and assigns binary bits based on the coordinate's location, producing strings such as \texttt{gcpugf00y}. For \textbf{TEXT}~\citep{yuhan2025}, the coordinate is provided directly as a textual string, we use six decimals for the representation, such as \texttt{(34.062403, -118.372775)}. For \textbf{LATENT}, the placeholder is replaced in the input embedding sequence by the continuous spatial representation $\mathbf{z}_{\mathbf{x}}$. For \textbf{HYBRID}, the textual coordinate and continuous spatial representation are provided together as $\texttt{(34.062403, -118.372775)}\,\mathbf{z}_{\mathbf{x}}$. The latent representation is inserted through the model's input embedding interface rather than converted into a textual token.

Given a target description tokenized as $y=(y_1,\ldots,y_T)$ and the representation-dependent prompt $c(\mathbf{x})$, we optimize the standard supervised fine-tuning objective
\begin{equation}
    \mathcal{L}_{\mathrm{SFT}}
    =
    -
    \sum_{t=1}^{T}
    \log
    p_{\psi}
    \left(
        y_t
        \mid
        c(\mathbf{x}),
        y_{<t}
    \right).
\end{equation}

The pretrained LLM parameters remain frozen, and the model is adapted using LoRA~\citep{lora}. The original token embeddings are also frozen. We fine-tune each backbone while freezing its pretrained weights and applying LoRA to the same attention projections, with rank $r=64$, scaling $\alpha=128$, and dropout $0.05$. For the dense models, adapters are also applied to the SwiGLU feed-forward projections implemented as standard linear layers. This configuration yields $262.28$M trainable LLM parameters for Gemma-4-12B and $116.39$M for Qwen3.5-9B. Qwen's linear-attention blocks remain unchanged. GPT-OSS uses a mixture-of-experts architecture. To match Qwen's trainable-parameter count, adapters are instantiated for all $32$ experts, we assign the expert adapters rank $r_{\mathrm{MoE}}=8$ and scaling $\alpha_{\mathrm{MoE}}=16$, preserving the update ratio $\alpha_{\mathrm{MoE}}/r_{\mathrm{MoE}}=\alpha/r=2$. Each expert's fused gate--up and down projections receives an adapter, while the router remains frozen and expert dropout is set to $0$. The resulting GPT-OSS adapter contains $120.32$M trainable parameters, comparable to Qwen.

For \textbf{LATENT} and \textbf{HYBRID}, the geospatial encoder and projection modules are trained from scratch and therefore use a learning rate ten times larger than that of the LoRA parameters. We first perform one encoder-only epoch to initialize the geospatial modules before jointly optimizing them with the LoRA adapters.

At inference time, we generate four descriptions for each location using temperature $0.8$, top-$k$ sampling with $k=30$, and nucleus sampling with $p=0.95$. These four descriptions form the generated set $\widehat{\mathcal{Y}}_i$ used by the captioning evaluation described in Appendix~\ref{appendix:metric_captioning}.

\subsection{Distance-Stratified Evaluation}
\label{appendix:distance_stratified}

The aggregate captioning scores defined in Appendix~\ref{appendix:metric_captioning} may obscure how geospatial reasoning changes with a test location's distance from the observed training data. We therefore stratify locations in unobserved regions by their distance to $\mathcal{X}_{\mathrm{obs}}$ and compare spatial generalization ability of different methods within each distance range, separately for each city and LLM backbone.

\paragraph{Distance to Observed Locations.}

For this analysis, we restrict $\mathcal{I}_{\mathrm{eval}}$ to unique test locations evaluated by every method, ensuring a paired comparison across representations. For each $\mathbf{x}_{i}\in\mathcal{H}$, we calculate its distance to the observed training data in meters, to the nearest location in $\mathcal{X}_{\mathrm{obs}}$:
\begin{equation}
    d_{i}
    =
    \min_{\mathbf{x}'\in\mathcal{X}_{\mathrm{obs}}}
    \operatorname{hav}
    \left(
        \mathbf{x}_{i},
        \mathbf{x}'
    \right).
\end{equation}
Here, $\operatorname{hav}(\cdot,\cdot)$ denotes the haversine distance. Larger $d_{i}$ indicates that the model must bridge a wider spatial gap from the observed training locations.

\paragraph{Per-Location Scores.}

For each representation method $m$, let $F_{1,i}^{m}$ denote the per-location many-to-many alignment score computed from $\widehat{\mathcal{Y}}_{i}^{m}$ and $\mathcal{Y}_{i}$ using the procedure defined in Appendix~\ref{appendix:metric_captioning}. We perform the distance-stratified analysis independently for BERTScore F1. 

\paragraph{Distance Binning and Paired Contrasts.}

We divide $\mathcal{I}_{\mathrm{eval}}$ into $K=5$ equal-count bins $\{B_{b}\}_{b=1}^{K}$ according to the quintiles of $\{d_{i}:i\in\mathcal{I}_{\mathrm{eval}}\}$. Equal-count binning avoids sparsely populated long-distance intervals because the distribution of $d_{i}$ is right-skewed. For representation method $m$, the mean score within bin $B_{b}$ is
\begin{equation}
    \overline{F}_{1,b}^{m}
    =
    \frac{1}{|B_{b}|}
    \sum_{i\in B_{b}}
    F_{1,i}^{m}.
\end{equation}

$\overline{F}_{1,b}^{m}$ is then plotted against the median value of $d_{i}$ within each bin.

\paragraph{Uncertainty.}

We construct $95\%$ percentile bootstrap confidence intervals by resampling locations with replacement within each bin using $N_{\mathrm{boot}}=1000$ replicates. The interval bounds are the $2.5$th and $97.5$th percentiles of the resulting bootstrap means. These intervals capture variation from sampling test locations in unobserved regions, but not randomness from decoding or from the fixed \texttt{k-NN} sampling procedure.

\begin{figure*}[t]
\centering
\begingroup
\setlength{\tabcolsep}{1pt}
\renewcommand{\arraystretch}{0}

\newcommand{\distancepanel}[1]{%
    \raisebox{-0.5\height}{%
        \includegraphics[
            width=0.31\textwidth
        ]{#1}%
    }%
}

\newcommand{\distancerow}[1]{%
    \raisebox{-0.5\height}{%
        \rotatebox[origin=c]{90}{%
            \scriptsize\bfseries #1%
        }%
    }%
}

\begin{tabular}{@{}cccc@{}}
&
\scriptsize\textbf{GPT-OSS}
&
\scriptsize\textbf{Qwen3.5}
&
\scriptsize\textbf{Gemma4}
\\[1pt]

\distancerow{Los Angeles}
&
\distancepanel{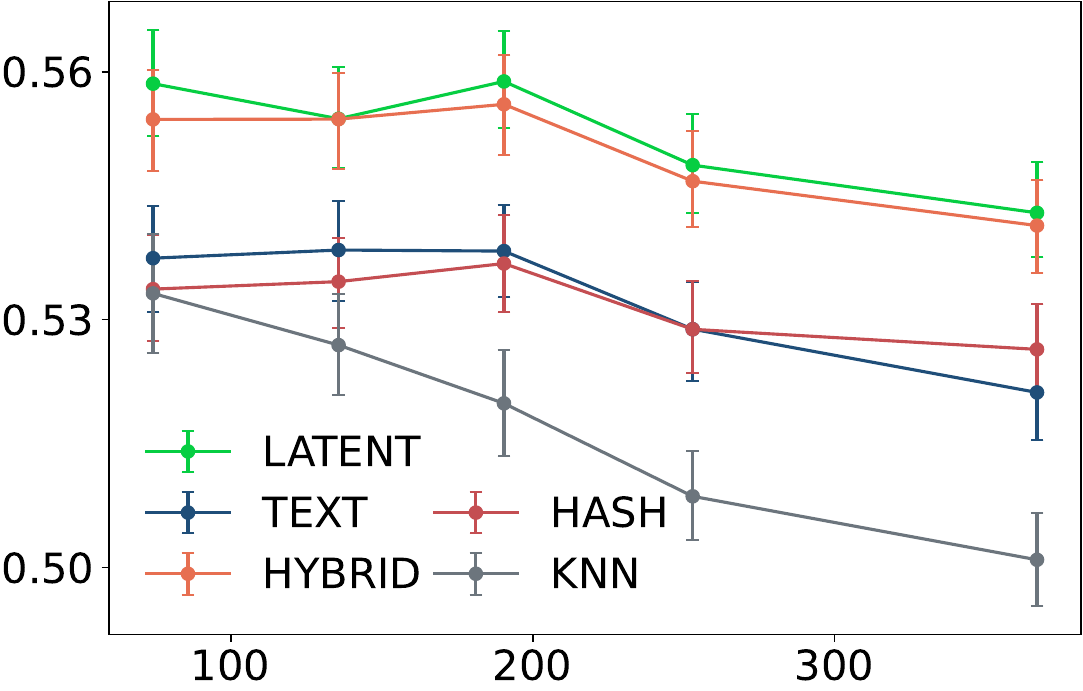}
&
\distancepanel{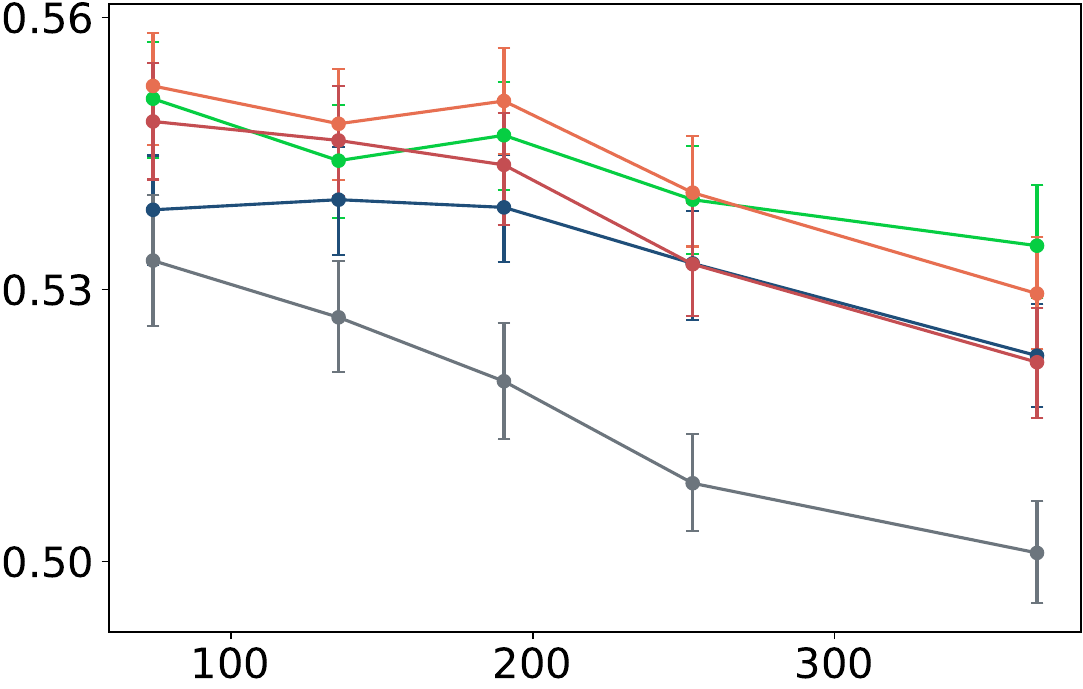}
&
\distancepanel{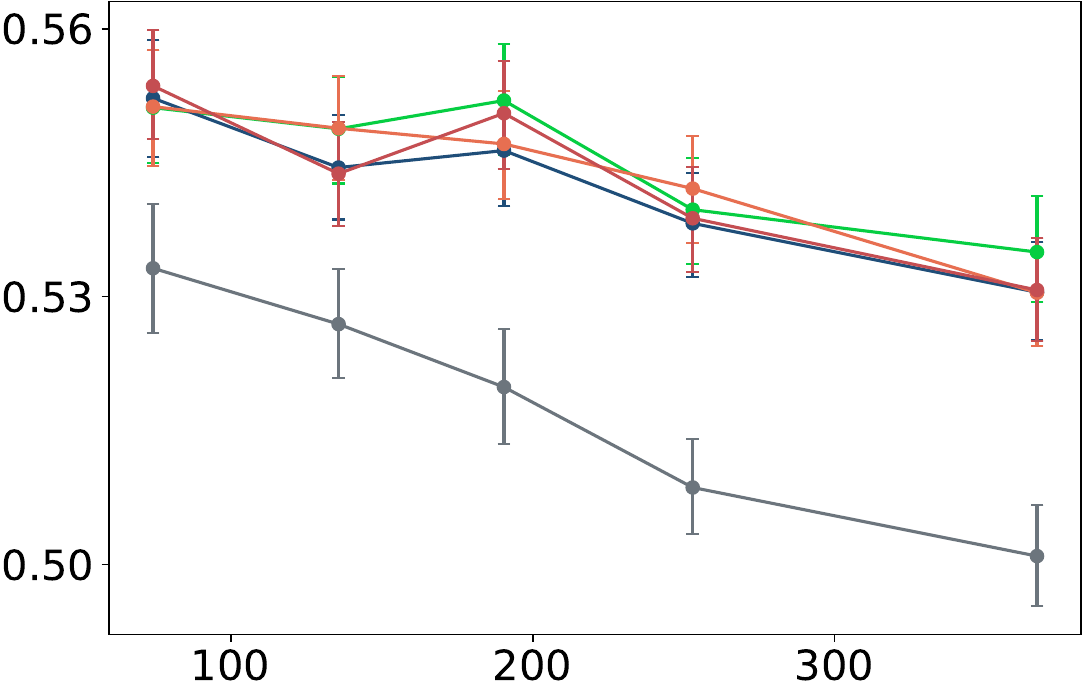}
\\[1pt]

\distancerow{London}
&
\distancepanel{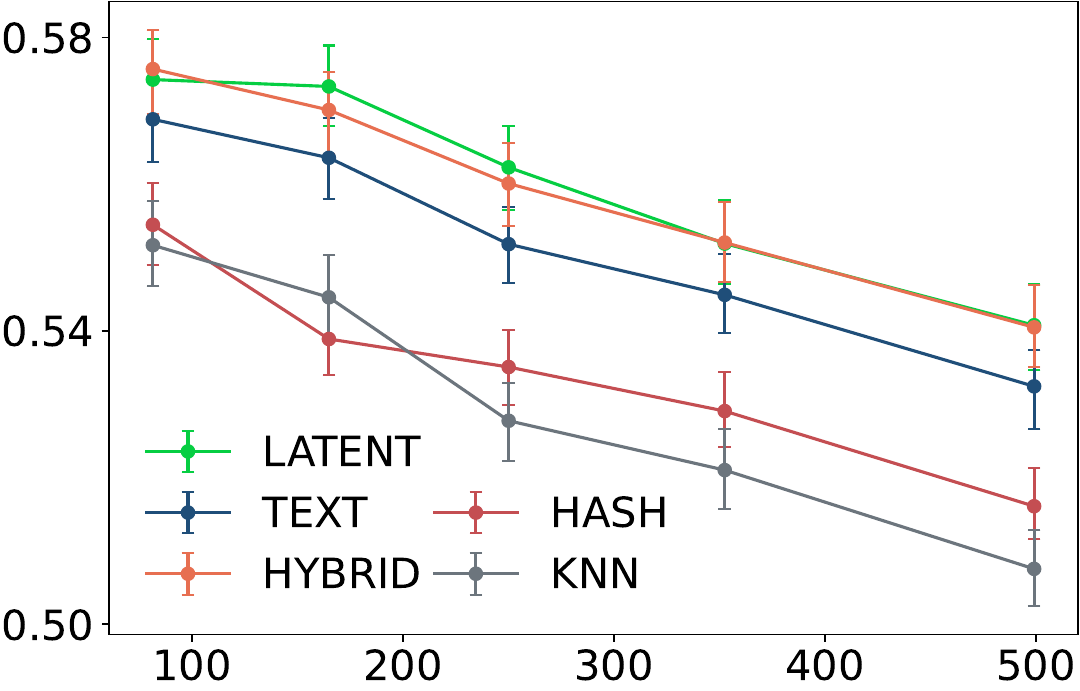}
&
\distancepanel{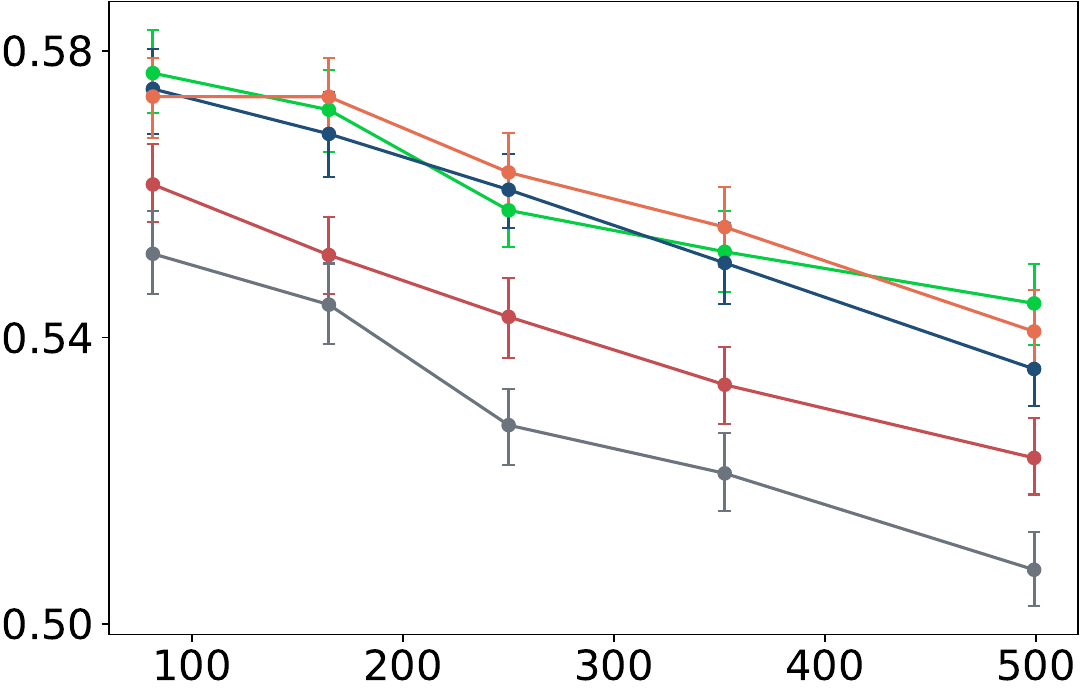}
&
\distancepanel{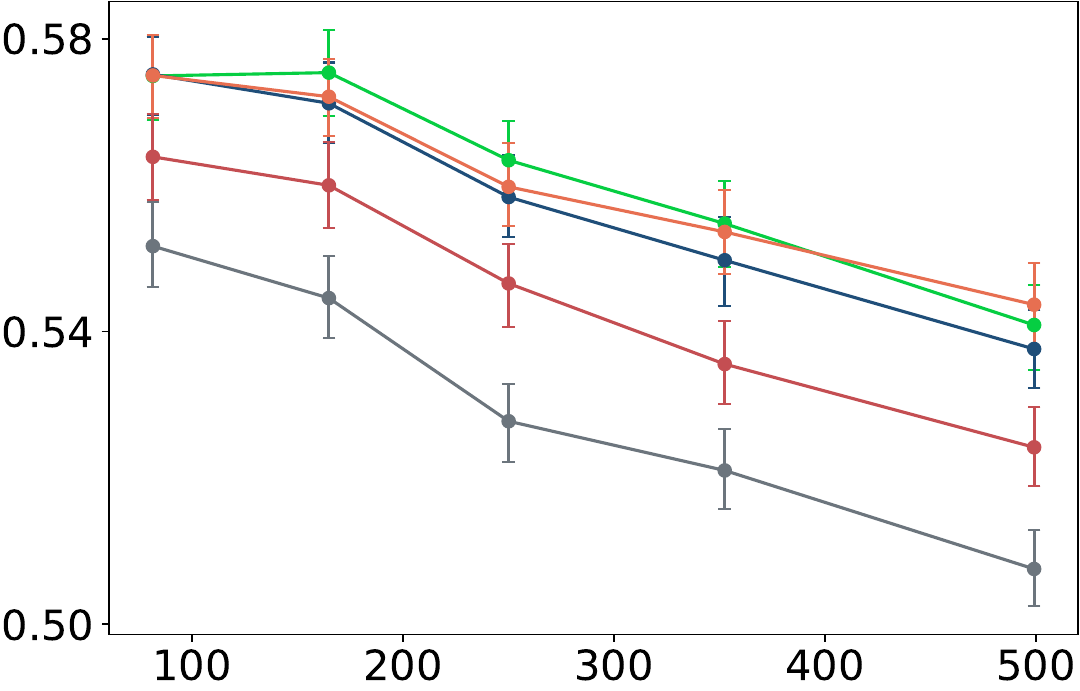}
\end{tabular}

\endgroup
\caption{\textbf{Distance-stratified geospatial captioning performance.} Rows correspond to Los Angeles and London, and columns correspond to the GPT-OSS, Qwen3.5, and Gemma4 backbones. The horizontal axis reports the median nearest-observation distance $d_i$, in meters, within each equal-count distance bin $B_b$. The vertical axis reports $\overline{F}_{1,b}^{m}$, the mean per-location BERTScore F1 achieved by representation method $m$ within bin $B_b$. Curves denote representation methods, and error bars indicate $95\%$ bootstrap confidence intervals.}
\label{fig:distance_stratified}
\end{figure*}

\paragraph{Results}

To interpret Figure~\ref{fig:distance_stratified}, we focus on the rate at which performance decreases as the distance from observed training locations increases. A flatter curve, corresponding to a smaller negative slope, indicates stronger spatial generalization because the model maintains higher captioning accuracy across wider spatial gaps. For visual clarity, we omit \textbf{GRID} because of its consistently poor performance. Across nearly all model--city settings, the plotted LLM-based methods degrade more gradually than \texttt{k-NN}, which shows a significant performance decrease after 200 meters, indicating that spatially explicit tuning enables reasoning beyond nearest-neighbor retrieval. In the farthest-distance bin, \textbf{LATENT} achieves the highest performance in five of the six settings, while \textbf{HYBRID}, which also incorporates the continuous latent encoding, performs best in the remaining setting. The difference between \textbf{LATENT} and the remaining methods becomes larger at the farthest bins. These results suggest that continuous spatial representations improve generalization across varying distances and that their advantage becomes more pronounced as the spatial gap increases.

\subsection{GeoGrounding Model}
\label{appendix:grounding}

GeoGrounding aligns natural-language descriptions with geographic locations through a CLIP-style contrastive objective~\citep{clip}. Given a location--description pair $(\mathbf{x}_i,y_i)$, we use \texttt{Qwen/Qwen3-Embedding-4B}\footnote{\url{https://huggingface.co/Qwen/Qwen3-Embedding-4B}} as the text encoder and take the hidden state of the final token as the pooled description representation. The text encoder remains frozen during training, while a trainable MLP projects its output into the shared embedding space. The location branch encodes $\mathbf{x}_i$ using the Geo2Vec position encoding $PE$~\citep{geo2vec}, followed by a separate trainable MLP. The normalized representations are
\begin{equation}
    E_{\phi}^{\mathrm{text}}(y_i)
    =
    \operatorname{Norm}
    \left(
        P_{\phi}^{\mathrm{text}}
        \left(
            H_{\mathrm{Qwen}}(y_i)
        \right)
    \right),
\end{equation}
\begin{equation}
    E_{\phi}^{\mathrm{loc}}(\mathbf{x}_i)
    =
    \operatorname{Norm}
    \left(
        P_{\phi}^{\mathrm{loc}}
        \left(
            PE(\mathbf{x}_i)
        \right)
    \right),
\end{equation}
where $P_{\phi}^{\mathrm{text}}$ and $P_{\phi}^{\mathrm{loc}}$ are the trainable projection MLPs and $\operatorname{Norm}(\cdot)$ denotes $\ell_2$ normalization.

For a batch of $B$ paired descriptions and locations, their temperature-scaled similarity is
\begin{equation}
    s_{ij}
    =
    \frac{
        E_{\phi}^{\mathrm{text}}(y_i)^{\top}
        E_{\phi}^{\mathrm{loc}}(\mathbf{x}_j)
    }{
        \tau
    },
\end{equation}
where $\tau = 1.0$ is the contrastive temperature. We optimize the symmetric text-to-location and location-to-text losses
\begin{equation}
    \mathcal{L}_{\mathrm{text\rightarrow loc}}
    =
    -
    \frac{1}{B}
    \sum_{i=1}^{B}
    \log
    \frac{
        \exp(s_{ii})
    }{
        \sum_{j=1}^{B}
        \exp(s_{ij})
    },
\end{equation}
\begin{equation}
    \mathcal{L}_{\mathrm{loc\rightarrow text}}
    =
    -
    \frac{1}{B}
    \sum_{i=1}^{B}
    \log
    \frac{
        \exp(s_{ii})
    }{
        \sum_{j=1}^{B}
        \exp(s_{ji})
    },
\end{equation}
and
\begin{equation}
    \mathcal{L}_{\mathrm{ground}}
    =
    \frac{1}{2}
    \left(
        \mathcal{L}_{\mathrm{text\rightarrow loc}}
        +
        \mathcal{L}_{\mathrm{loc\rightarrow text}}
    \right).
\end{equation}

At inference time, we precompute $E_{\phi}^{\mathrm{loc}}(\mathbf{x})$ for every location $\mathbf{x}$ in the predefined $M\times M$ grid $\mathcal{G}$. Given a textual query $y$, we calculate its representation once and compare it with every grid-cell representation:
\begin{equation}
    s_{\phi}(\mathbf{x},y)
    =
    E_{\phi}^{\mathrm{text}}(y)^{\top}
    E_{\phi}^{\mathrm{loc}}(\mathbf{x}),
    \qquad
    \mathbf{x}\in\mathcal{G}.
\end{equation}
The resulting similarity field provides a dense grounding map over the geographic domain, from which the highest-scoring locations or regions associated with the query can be retrieved.

\subsection{Layer-Wise Probing}
\label{appendix:probing}

\begin{figure}[h]
    \centering

    \begin{subfigure}[b]{0.33\columnwidth}
        \centering
        \includegraphics[width=\linewidth]{figs/probes/la_probe_km_openai_mlp.pdf}
        \caption{GPT-OSS}
        \label{fig:GPT-OSS_probe}
    \end{subfigure}%
    \hspace{0em}%
    \begin{subfigure}[b]{0.33\columnwidth}
        \centering
        \includegraphics[width=\linewidth]{figs/probes/la_probe_km_qwen_mlp.pdf}
        \caption{Qwen3.5}
        \label{fig:qwen_probe}
    \end{subfigure}%
    \hspace{0em}%
    \begin{subfigure}[b]{0.33\columnwidth}
        \centering
        \includegraphics[width=\linewidth]{figs/probes/la_probe_km_gemma_mlp.pdf}
        \caption{Gemma4}
        \label{fig:gemma_probe}
    \end{subfigure}

    \vspace{0.2em}

    \begin{subfigure}[b]{0.33\columnwidth}
        \centering
        \includegraphics[width=\linewidth]{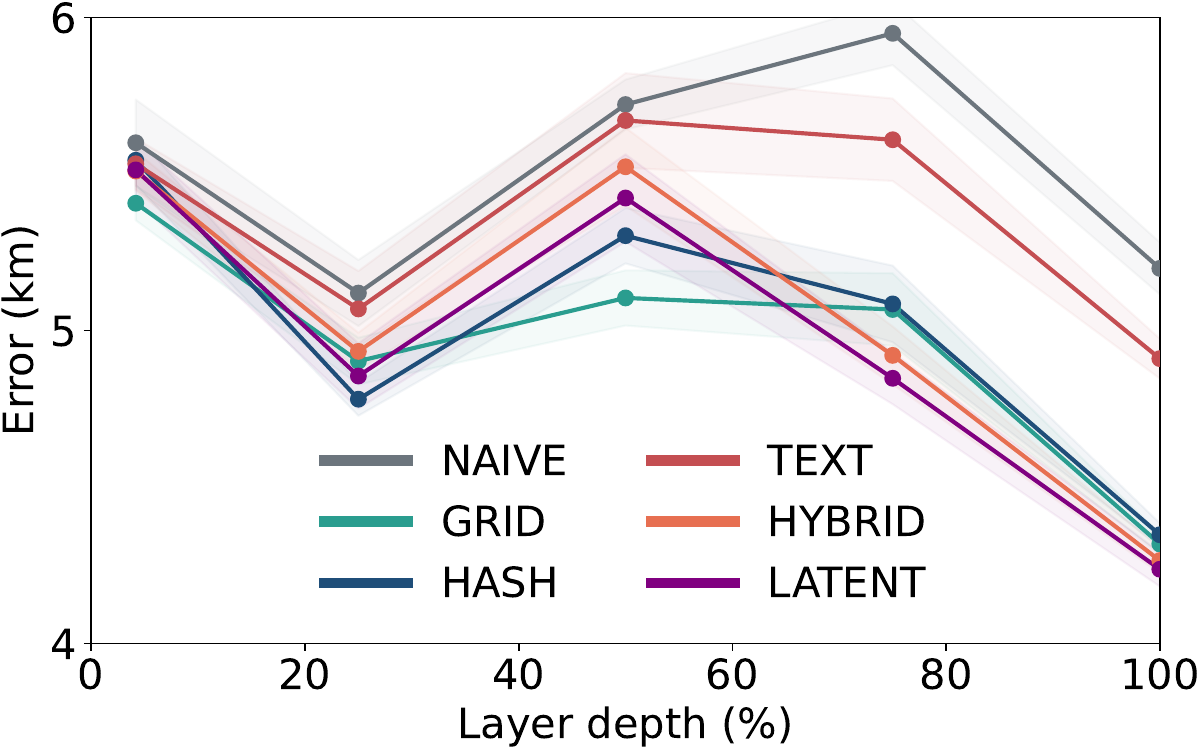}
        \caption{GPT-OSS}
        \label{fig:GPT-OSS_probe_london}
    \end{subfigure}%
    \hspace{0em}%
    \begin{subfigure}[b]{0.33\columnwidth}
        \centering
        \includegraphics[width=\linewidth]{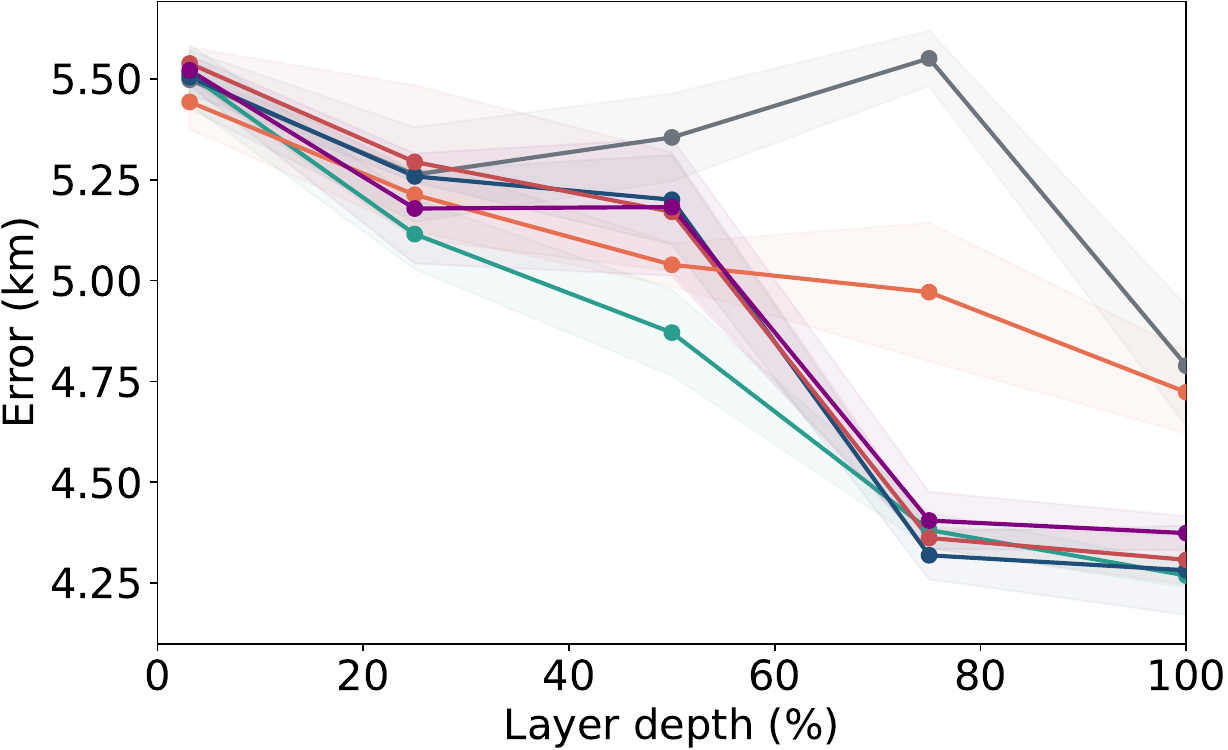}
        \caption{Qwen3.5}
        \label{fig:qwen_probe_london}
    \end{subfigure}%
    \hspace{0em}%
    \begin{subfigure}[b]{0.33\columnwidth}
        \centering
        \includegraphics[width=\linewidth]{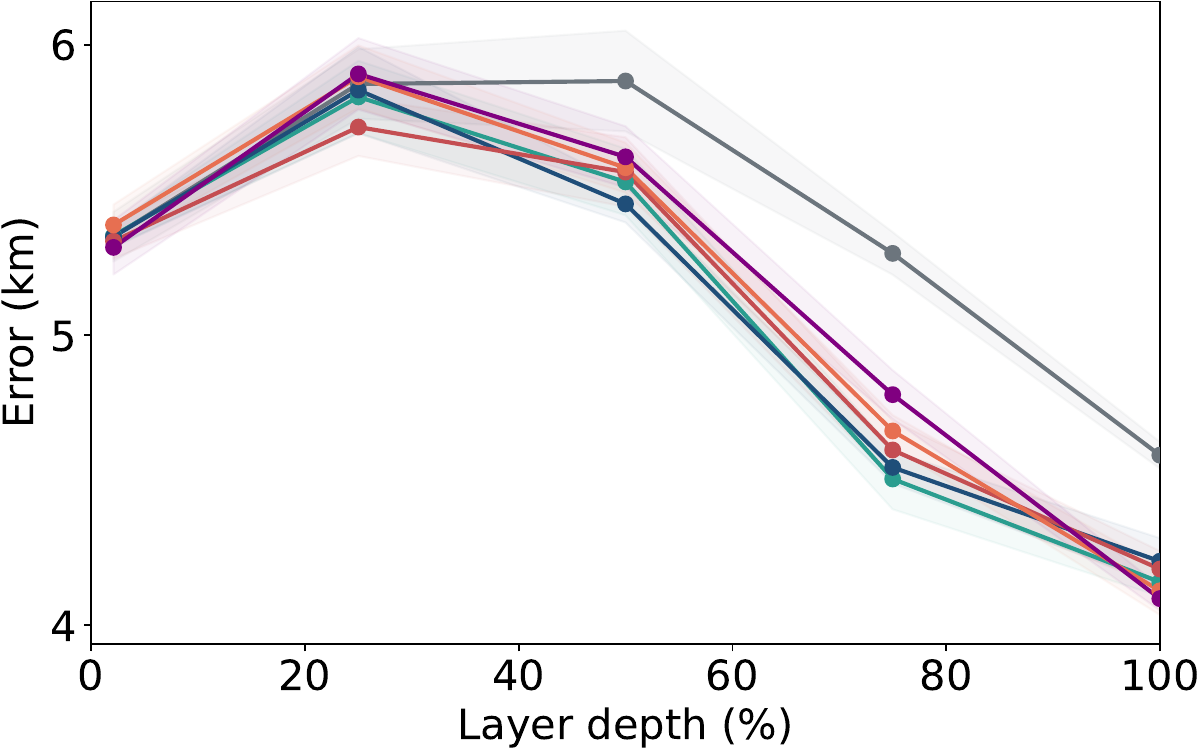}
        \caption{Gemma4}
        \label{fig:gemma_probe_london}
    \end{subfigure}

    \caption{
    \textbf{Layer-wise geospatial probing performance across model families.}
    The top row reports results on Los Angeles, and the bottom row reports results on London. Lower probing error indicates that geographic location is more explicitly encoded in the hidden representations. The curves report mean performance, while the shaded regions indicate standard deviations.
    }
    \label{fig:interlayer_analysis_appendix}
\end{figure}

Prior work uses Ridge linear probes to examine whether pretrained LLMs encode space and time in their hidden representations~\citep{spacetime}. Given place names and their corresponding coordinates, hidden states are extracted using last-token pooling and used to predict geographic coordinates. The underlying intuition is that if geographic structure is encoded in the model, location information should be recoverable from its hidden representations using a lightweight probe.

We extend this approach to examine how precisely our tuned LLMs encode geographic space and how this information evolves across layers. For each description $y_i$, we prompt the model to predict its location and manually prefill the beginning of the assistant response as follows:

\begin{tcolorbox}[
    colback=gray!5,
    colframe=gray!40,
    boxrule=0.5pt,
    arc=1mm,
    left=4pt,
    right=4pt,
    top=4pt,
    bottom=4pt
]
\small
\textbf{System:} You are a helpful assistant for geo-localization. Always use only the information contained in the provided representation to generate the description.

\textbf{User:} Answer the question using your geospatial knowledge and memory regarding the given description. What are the latitude and longitude coordinates of: \texttt{<DESCRIPTION>}?

\textbf{Assistant:} \texttt{<DESCRIPTION>} is at \texttt{<TOKEN\_EXTRACTED>}.
\end{tcolorbox}

Here, \texttt{<DESCRIPTION>} is replaced by $y_i$, and \texttt{<TOKEN\_EXTRACTED>} marks the first autoregressive position after the prefixed response, before any coordinate is generated. We extract the hidden state at this position from five selected transformer layers. This design is intended to elicit the geographic information associated with each description, allowing every model to express its strongest recoverable spatial awareness under a consistent prompting protocol.

Let $\mathbf{h}_{i}^{(\ell)}$ denote the extracted representation at layer $\ell$. We train a separate multilayer perceptron $r_{\ell}$ for each layer to predict the corresponding coordinate $\mathbf{x}_i$:
\begin{equation}
    \widehat{\mathbf{x}}_{i}^{(\ell)}
    =
    r_{\ell}
    \left(
        \mathbf{h}_{i}^{(\ell)}
    \right).
\end{equation}
Unlike a linear probe, the MLP can recover nonlinearly encoded location information, allowing us to assess the amount of geographic information available in each representation rather than only its linear separability.

For each representation method and LLM, we randomly divide $\mathcal{D}_{\mathrm{train}}$ into probing-training and probing-testing subsets. Each probe is trained for $10$ epochs using the mean squared error objective
\begin{equation}
    \mathcal{L}_{\mathrm{probe}}^{(\ell)}
    =
    \frac{1}{|\mathcal{I}_{\mathrm{probe}}|}
    \sum_{i\in\mathcal{I}_{\mathrm{probe}}}
    \left\|
        r_{\ell}
        \left(
            \mathbf{h}_{i}^{(\ell)}
        \right)
        -
        \mathbf{x}_{i}
    \right\|_{2}^{2},
\end{equation}
with a learning rate of $1\times10^{-3}$. We repeat the training and evaluation pipeline five times using different random splits and report the mean regression error with its standard deviation.

If an LLM organizes geographic knowledge as a latent spatial field, its hidden representations should retain information about where each description is geographically situated. Lower testing regression error therefore indicates that location information is more accurately recoverable from the corresponding layer. This analysis provides evidence about the internal organization of geographic knowledge.

In Figure~\ref{fig:interlayer_analysis_appendix}, across most settings, probing error also decreases substantially toward the later layers and reaches its minimum near the final layer. This trend suggests that geographic information is progressively organized across the model depth, with later representations expressing the most precise spatial awareness.

\begin{figure*}[!t]
\centering
\begingroup
\setlength{\tabcolsep}{1pt}
\renewcommand{\arraystretch}{0}

\newcommand{\rowlabel}[1]{%
    \begin{minipage}[c][0.1\textheight][c]{0.087\textwidth}
        \centering
        \small\bfseries #1
    \end{minipage}%
}

\newcommand{\clustermap}[1]{%
    \begin{minipage}[c][0.11\textheight][c]{0.30\textwidth}
        \centering
        \includegraphics[
            width=\linewidth,
            height=0.1\textheight,
            keepaspectratio
        ]{#1}
    \end{minipage}%
}

\begin{tabular}{@{}cccc@{}}
&
\textbf{GPT}
&
\textbf{Qwen}
&
\textbf{Gemma}
\\[-1pt]

\rowlabel{NAIVE}
&
\clustermap{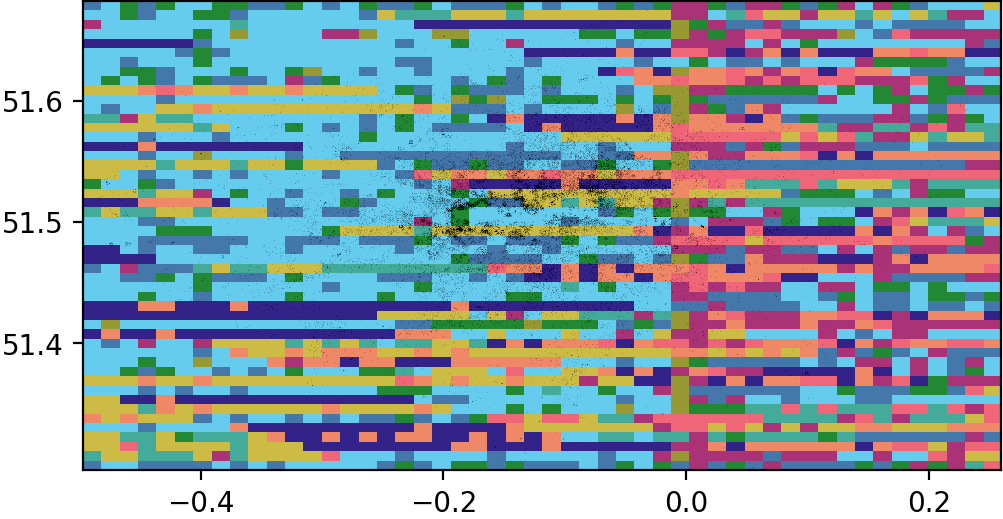}
&
\clustermap{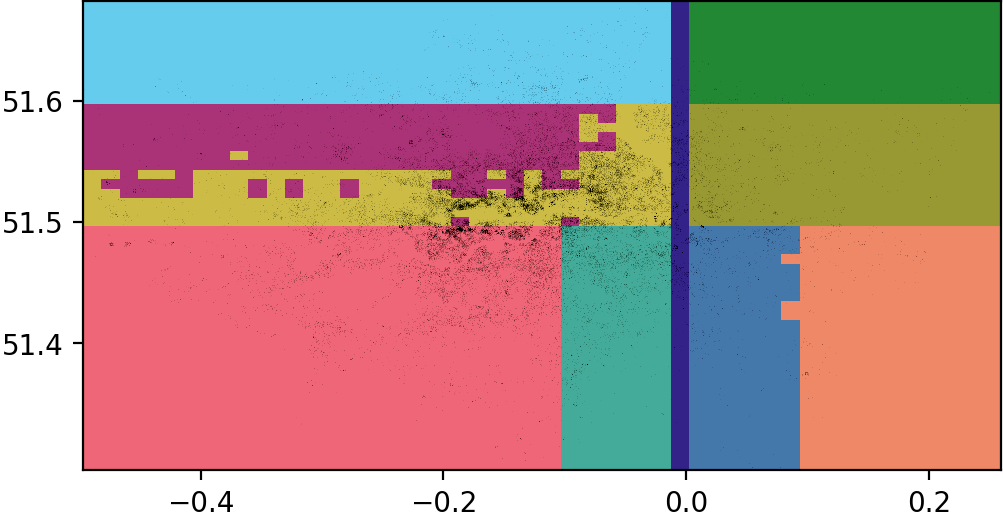}
&
\clustermap{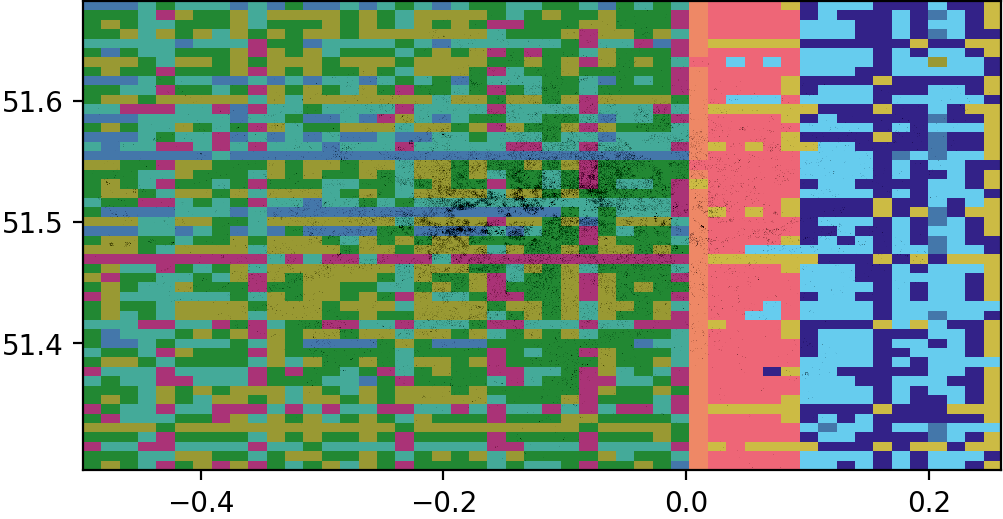}
\\[-4pt]

\rowlabel{GRID}
&
\clustermap{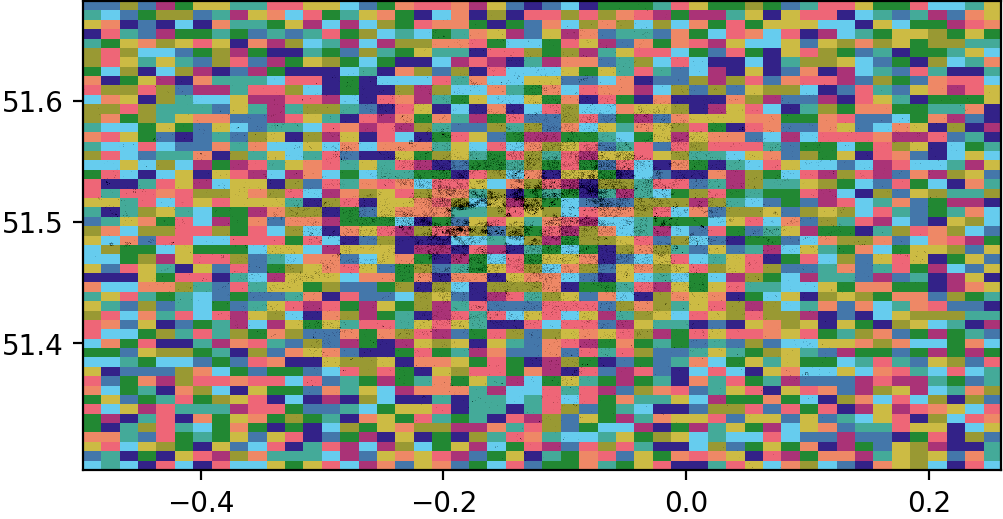}
&
\clustermap{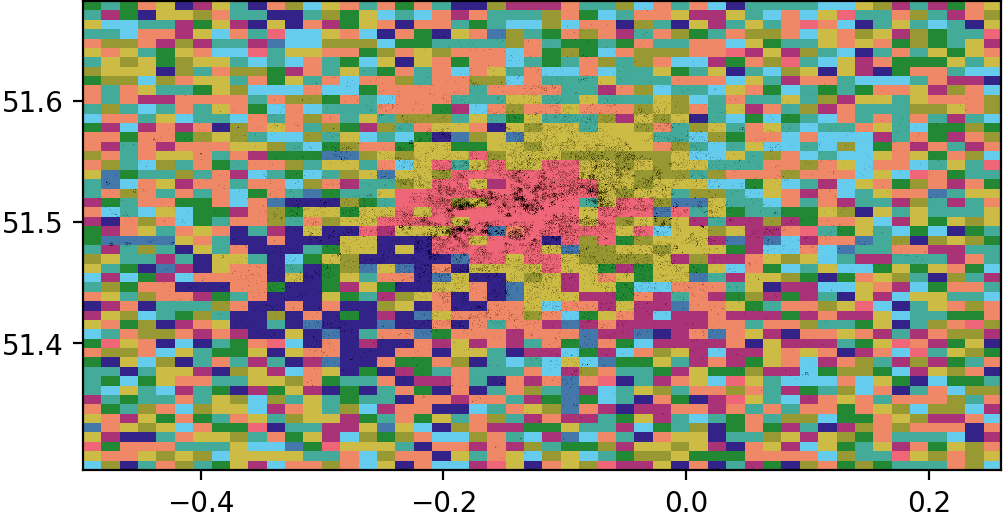}
&
\clustermap{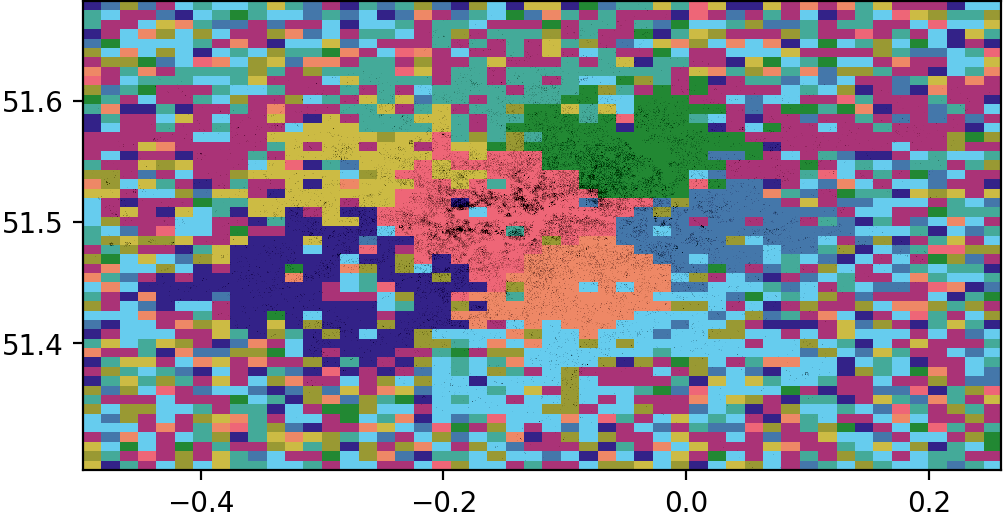}
\\[-4pt]

\rowlabel{HASH}
&
\clustermap{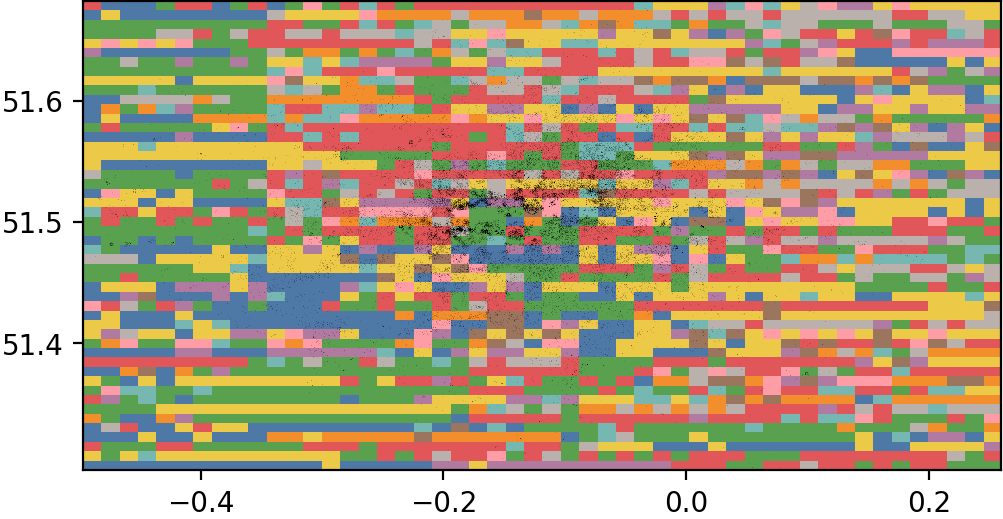}
&
\clustermap{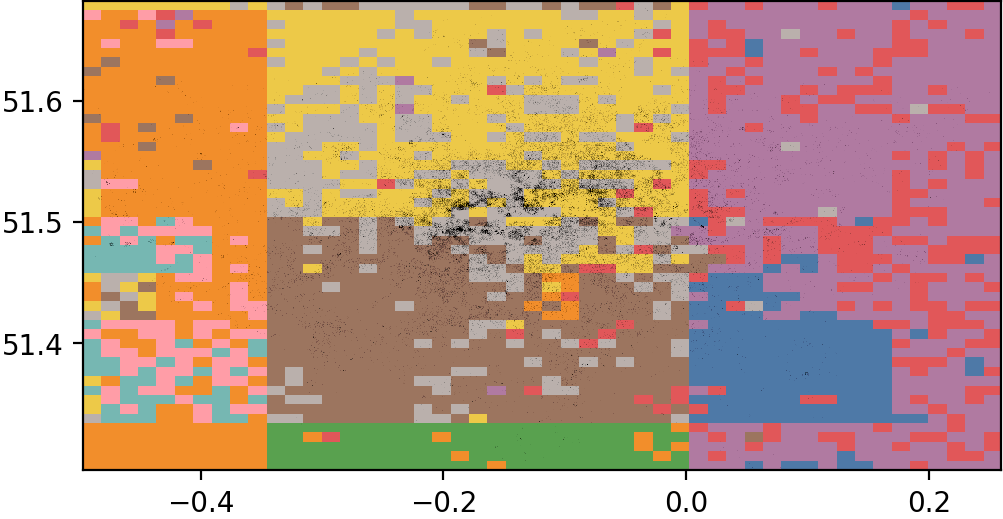}
&
\clustermap{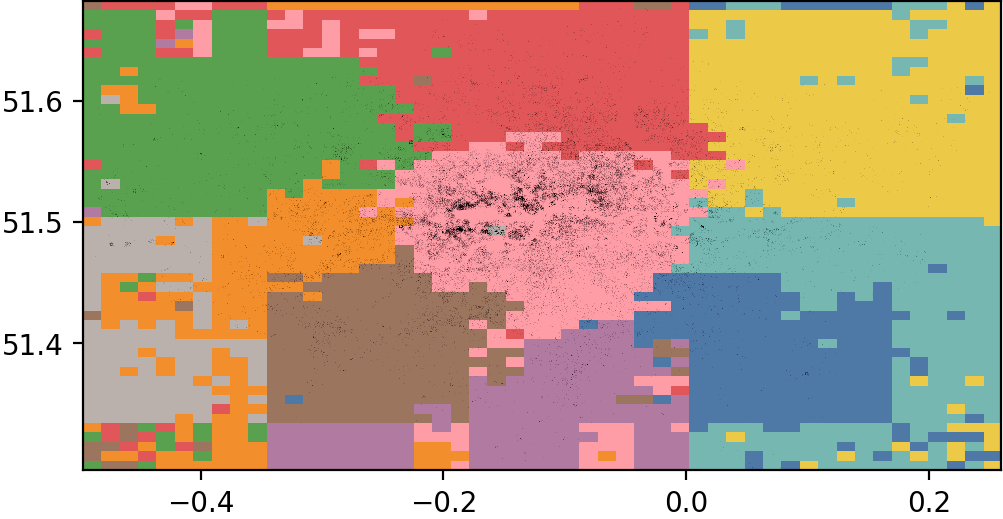}
\\[-4pt]

\rowlabel{TEXT}
&
\clustermap{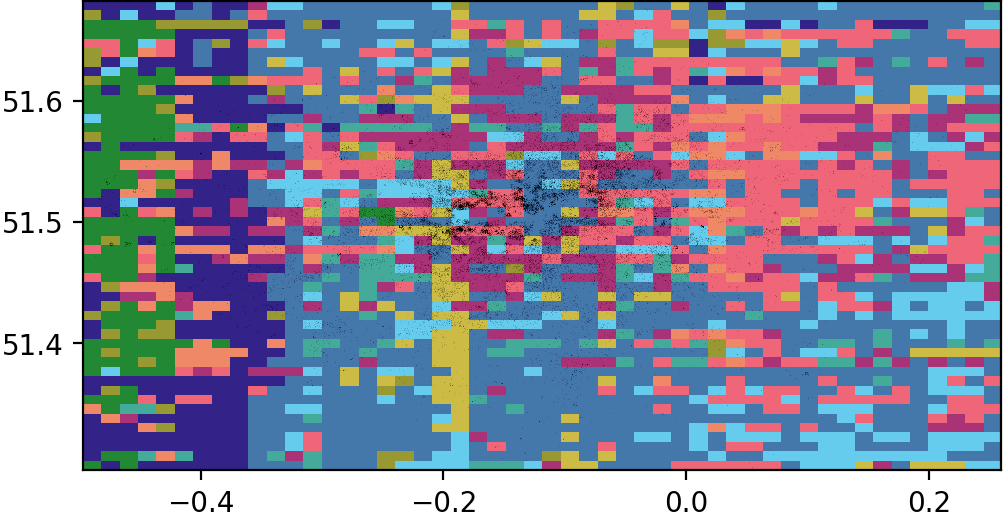}
&
\clustermap{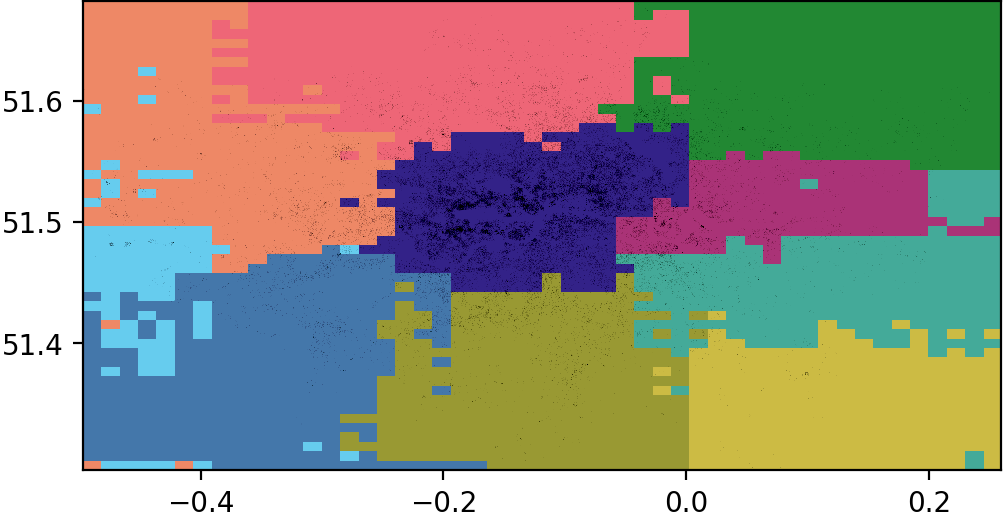}
&
\clustermap{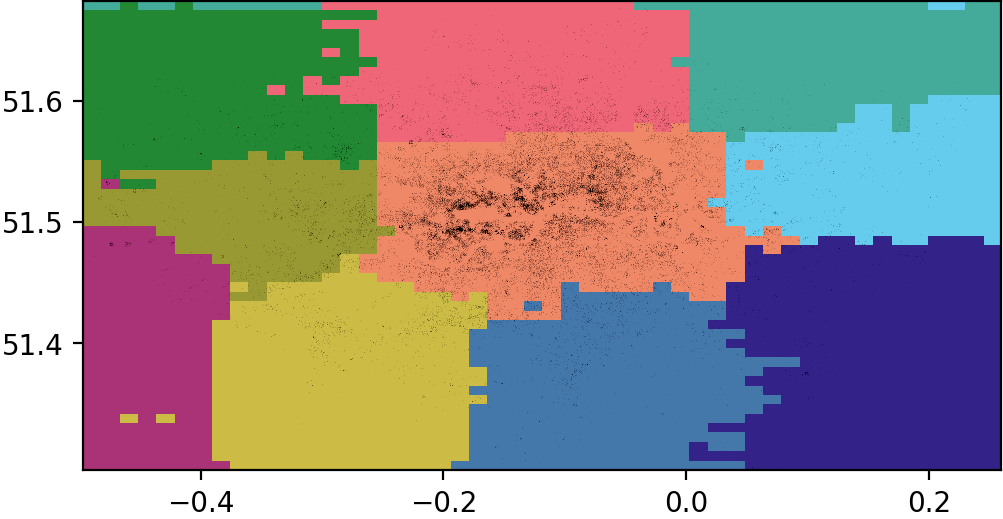}
\\[-4pt]

\rowlabel{HYBRID}
&
\clustermap{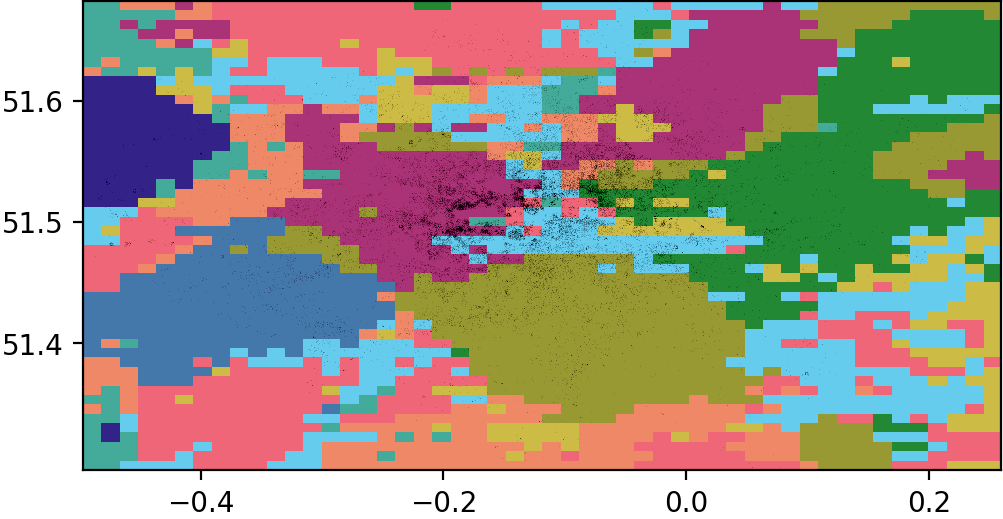}
&
\clustermap{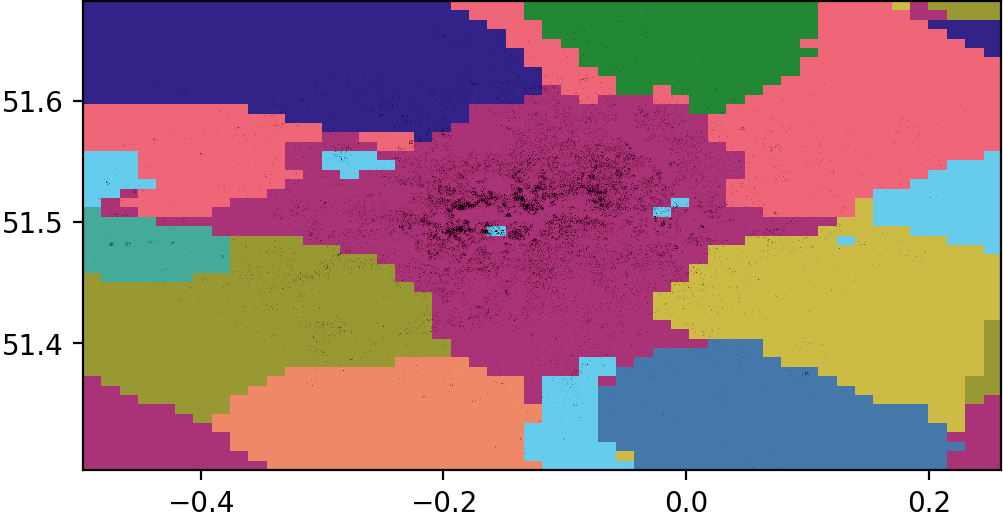}
&
\clustermap{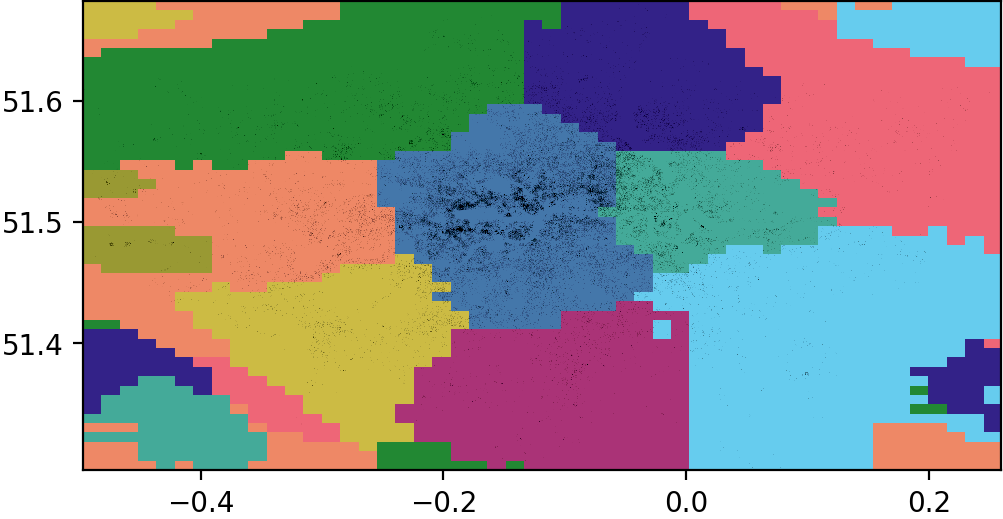}
\\[-4pt]

\rowlabel{LATENT}
&
\clustermap{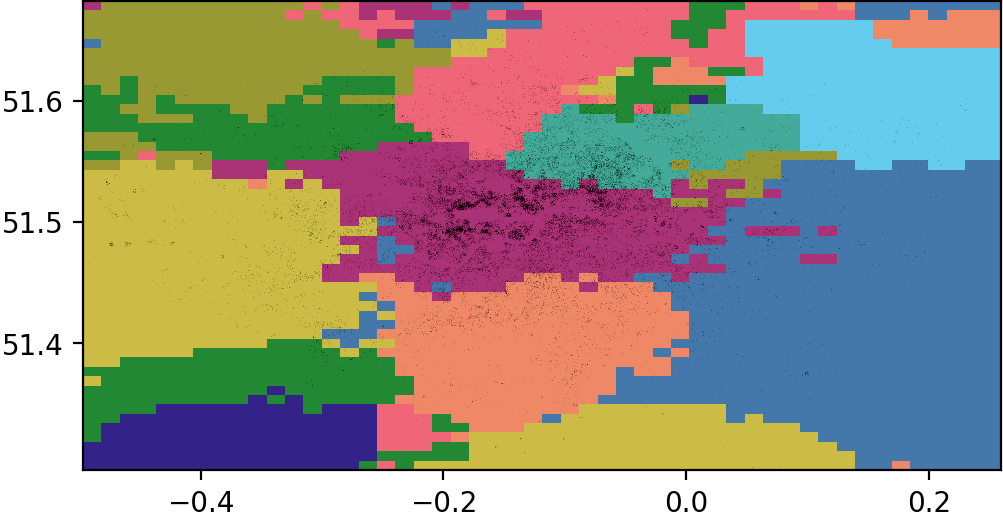}
&
\clustermap{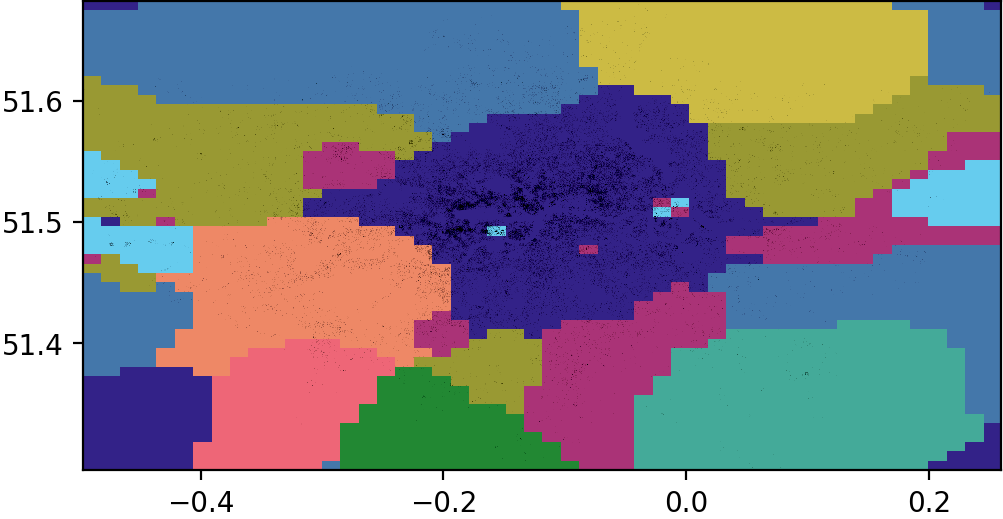}
&
\clustermap{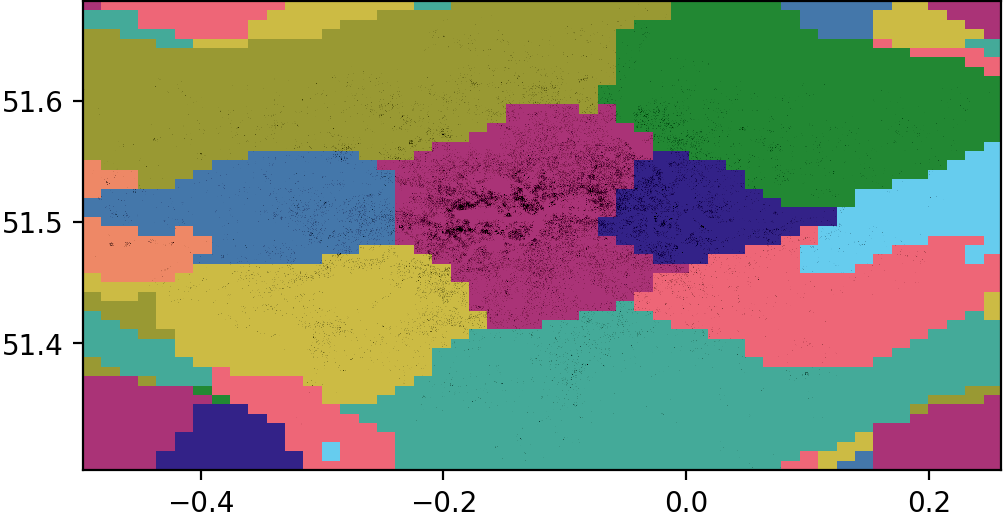}
\end{tabular}

\endgroup
\caption{\textbf{Cluster maps for the London dataset across model families and spatial representations.} Columns correspond to GPT, Qwen, and Gemma, while rows correspond to \textbf{NAIVE}, \textbf{GRID}, \textbf{HASH}, \textbf{TEXT}, \textbf{HYBRID}, and \textbf{LATENT}.}
\label{fig:london_cluster_maps_appendix}
\end{figure*}

\begin{figure*}[!t]
\centering
\begingroup
\setlength{\tabcolsep}{1pt}
\renewcommand{\arraystretch}{0}

\newcommand{\rowlabel}[1]{%
    \begin{minipage}[c][0.145\textheight][c]{0.087\textwidth}
        \centering
        \small\bfseries #1
    \end{minipage}%
}

\newcommand{\clustermap}[1]{%
    \begin{minipage}[c][0.145\textheight][c]{0.30\textwidth}
        \centering
        \includegraphics[
            width=\linewidth,
            height=0.14\textheight,
            keepaspectratio
        ]{#1}
    \end{minipage}%
}

\begin{tabular}{@{}cccc@{}}
&
\textbf{GPT}
&
\textbf{Qwen}
&
\textbf{Gemma}
\\[2pt]

\rowlabel{NAIVE}
&
\clustermap{figs/cluster/la/la_openai_cluster_map_NAIVE.png}
&
\clustermap{figs/cluster/la/la_qwen_cluster_map_NAIVE.png}
&
\clustermap{figs/cluster/la/la_gemma_cluster_map_NAIVE.png}
\\

\rowlabel{GRID}
&
\clustermap{figs/cluster/la/la_openai_cluster_map_GRID.png}
&
\clustermap{figs/cluster/la/la_qwen_cluster_map_GRID.png}
&
\clustermap{figs/cluster/la/la_gemma_cluster_map_GRID.png}
\\

\rowlabel{HASH}
&
\clustermap{figs/cluster/la/la_openai_cluster_map_HASH.png}
&
\clustermap{figs/cluster/la/la_qwen_cluster_map_HASH.png}
&
\clustermap{figs/cluster/la/la_gemma_cluster_map_HASH.png}
\\

\rowlabel{TEXT}
&
\clustermap{figs/cluster/la/la_openai_cluster_map_DIRECT.png}
&
\clustermap{figs/cluster/la/la_qwen_cluster_map_DIRECT.png}
&
\clustermap{figs/cluster/la/la_gemma_cluster_map_DIRECT.png}
\\

\rowlabel{HYBRID}
&
\clustermap{figs/cluster/la/la_openai_cluster_map_HYBRID.png}
&
\clustermap{figs/cluster/la/la_qwen_cluster_map_HYBRID.png}
&
\clustermap{figs/cluster/la/la_gemma_cluster_map_HYBRID.png}
\\

\rowlabel{LATENT}
&
\clustermap{figs/cluster/la/la_openai_cluster_map_LATENT.png}
&
\clustermap{figs/cluster/la/la_qwen_cluster_map_LATENT.png}
&
\clustermap{figs/cluster/la/la_gemma_cluster_map_LATENT.png}
\end{tabular}

\endgroup
\caption{\textbf{Cluster maps for the Los Angeles dataset across model families and spatial representations.} Columns correspond to GPT, Qwen, and Gemma, while rows correspond to \textbf{NAIVE}, \textbf{GRID}, \textbf{HASH}, \textbf{TEXT},\textbf{HYBRID}, and \textbf{LATENT}.}
\label{fig:la_cluster_maps_appendix}
\end{figure*}

\subsection{Organization of Space}
\label{appendix:cluster}

To further investigate how LLMs internally organize geographic space and whether the learned spatial field is continuous, we cluster hidden representations elicited at different query locations and map the resulting assignments back to geographic space. Unlike the layer-wise probing analysis, this experiment does not train a decoder using geographic coordinates. Instead, it directly examines whether the representational similarity learned by the LLM corresponds to coherent spatial regions.

\paragraph{Feature extraction.}

Similar to the grid we used in evaluation, we uniformly sample a query grid $\mathcal{G}=\{\mathbf{x}_j\}_{j=1}^{N_q \times N_q}$ over each city's geographic domain, $N_q$ follows the training-split settings. For every query location $\mathbf{x}_j$, its corresponding \texttt{<GEO>} representation is inserted into the following prompt:

\begin{tcolorbox}[
    colback=gray!3,
    colframe=black!35,
    boxrule=0.5pt,
    arc=1mm,
    left=1.5mm,
    right=1.5mm,
    top=1mm,
    bottom=1mm
]
\small
\textbf{System:} You are a helpful assistant for geo-localization. Always use only the information contained in the provided representation to generate the description.

\vspace{1mm}

\textbf{User:} You are at location \texttt{<GEO>}. Generate a detailed description of the location.

\vspace{1mm}

\textbf{Assistant:} \texttt{<TOKEN\_EXTRACTED>}
\end{tcolorbox}

Here, \texttt{<TOKEN\_EXTRACTED>} marks the first assistant-generation position and is not provided as literal input to the model. We extract the hidden state at this position, before any location description has been generated. Let $\mathbf{h}_{m}(\mathbf{x}_j)\in\mathbb{R}^{d}$ denote the resulting hidden representation for model configuration $m$ at location $\mathbf{x}_j$.

\paragraph{Representation clustering.}

Before clustering, each hidden representation is $\ell_2$-normalized. This normalization reduces the influence of representation magnitude and emphasizes differences in representational direction.

For each model configuration, we independently apply $K$-means clustering to the normalized hidden states. We use $K=10$ clusters and $n_{init}=10$ centroid initializations and a fixed random seed for all model configurations. The resulting assignments are reshaped to the original $N_q\times N_q$ query grid and visualized over geographic space.

\paragraph{Interpretation.}

The clustering algorithm receives only hidden representations and does not use geographic coordinates when forming the clusters. Therefore, spatially compact clusters and smooth transitions between neighboring regions must arise from geographic structure encoded in the LLM's representations themselves. A fragmented map indicates that nearby locations are mapped to dissimilar hidden states, whereas contiguous regions and coherent boundaries indicate a smoother latent spatial field. Because clustering is performed independently for each configuration, cluster indices and colors do not correspond across models. The Los Angeles results are presented in Figure~\ref{fig:la_cluster_maps_appendix}, and the corresponding London results are shown in Figure~\ref{fig:london_cluster_maps_appendix}.

Without spatially explicit tuning, \textbf{NAIVE} exhibits only weak horizontal and vertical bands, likely because its textual coordinate input varies systematically with latitude and longitude. These coarse patterns do not separate the cities into sufficiently coherent regions to support useful spatially aware geographic inference. \textbf{GRID} remains largely fragmented because its identifiers encode no relationships between cells; the model must reconstruct these relationships solely from the similarity of training observations, imposing an additional learning burden. Very few coherent regions appear primarily in areas with exceptionally high observation density, consistent with its weak spatial generalization. \textbf{HASH} provides a hierarchical spatial index whose shared prefixes expose structure at multiple scales, reducing dependence on a single grid resolution. It consequently produces clearer spatial patterns than \textbf{GRID} for Qwen and Gemma, although the maps remain fragmented, particularly near cluster boundaries. For the text-only GPT-OSS model, \textbf{HASH} still fails to produce a well-organized spatial structure.

The \textbf{TEXT} representation similarly fails to organize space coherently for GPT-OSS in either city. For Qwen and Gemma, its clusters broadly follow spatial proximity but contain fragmented components near their boundaries as well. Its spatial organization is largely constrained by absolute position. In contrast, \textbf{LATENT} produces smoother transitions, more contiguous regions, and clearer spatial boundaries. It can also connect spatially disjoint but semantically similar regions. For example, the coastal areas around Santa Monica and Long Beach are grouped despite their geographic separation. Although this connectivity does not imply identical generated descriptions, it suggests that the learned field captures shared semantic--geographic structure rather than spatial proximity alone. Finally, \textbf{HYBRID} combines the local specificity of \textbf{TEXT} with the continuity of \textbf{LATENT}, largely preserving proximity-based organization while producing smoother cluster boundaries. This effect is particularly clear for GPT-OSS, where \textbf{HYBRID} recovers coherent spatial structure absent from the corresponding \textbf{TEXT} map.

\subsection{Qualitative Analysis}
\label{appendix:case}

We present qualitative case studies for \textbf{Geospatial Captioning} and \textbf{Geospatial Grounding} to illustrate how different spatial representations affect model behavior.

\paragraph{Geospatial Captioning.}

We evaluate the spatial awareness of the Gemma4-based models through response prefilling. Given a query location, we manually specify the beginning of the response and ask each model to complete the distance to Los Angeles International Airport (LAX):

\begin{table*}[!t]
\centering
\caption{\textbf{Generated distances to LAX under different spatial representations.} All distances are reported in miles. Bold values indicate the closest prediction to the measured distance for each query.}
\label{tab:lax_distance_examples}
\small
\setlength{\tabcolsep}{9pt}
\renewcommand{\arraystretch}{1.08}
\begin{tabular}{lcccccc}
\toprule
\textbf{ID}
& \textbf{Measured Distance}
& \textbf{NAIVE}
& \textbf{TEXT}
& \textbf{GRID}
& \textbf{LATENT}
& \textbf{HYBRID} \\
\midrule
$0$ & $1.3$  & $14.1$          & $15.0$ & $12.0$          & $\mathbf{2.5}$  & $10.0$ \\
$1$ & $7.3$  & $23.1$          & $15.0$ & $10.5$          & $\mathbf{10.0}$ & $15.0$ \\
$2$ & $9.6$  & $15.1$          & $15.0$ & $\mathbf{12.5}$ & $20.0$          & $20.0$ \\
$3$ & $14.0$ & $18.1$          & $20.0$ & $\mathbf{12.5}$ & $18.0$          & $17.0$ \\
$4$ & $14.0$ & $15.1$          & $30.0$ & $\mathbf{14.0}$ & $25.0$          & $30.0$ \\
$5$ & $18.0$ & $\mathbf{18.1}$ & $25.0$ & $12.5$          & $25.0$          & $20.0$ \\
$6$ & $21.3$ & $15.1$          & $25.0$ & $12.5$          & $\mathbf{24.0}$ & $30.0$ \\
\bottomrule
\end{tabular}
\end{table*}

\begin{figure}[h]
\centering
\includegraphics[
    width=0.6\textwidth
]{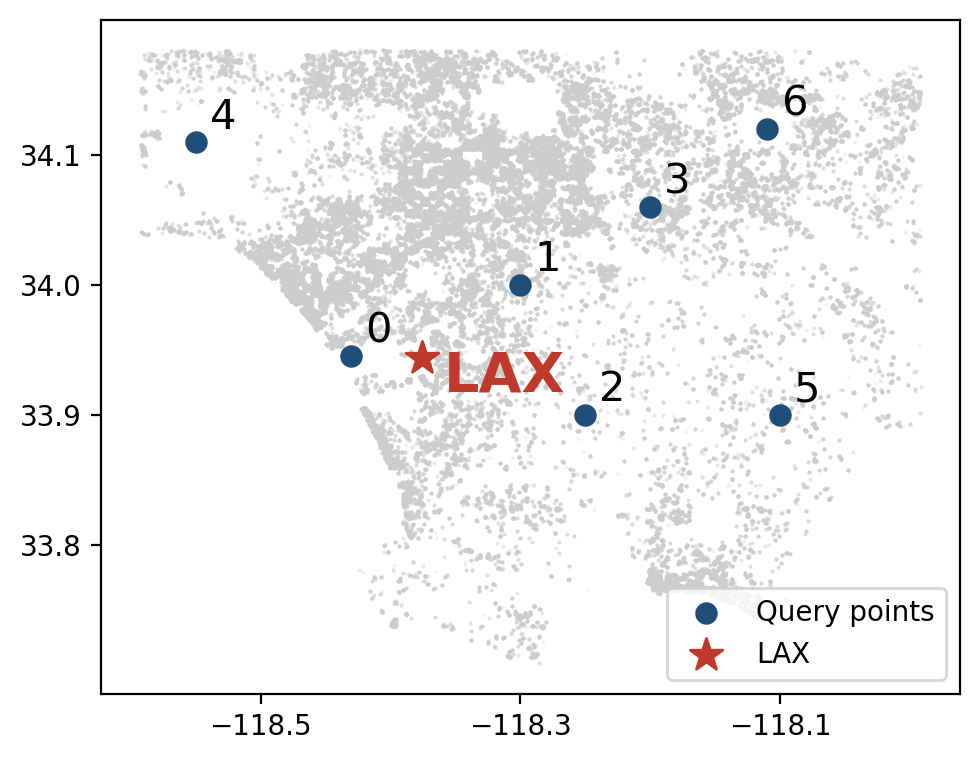}
\caption{\textbf{Query locations used for the LAX distance-estimation case study.} The seven locations are selected from spatially held-out regions and indexed from $0$ to $6$.}
\label{fig:lax_distance_examples}
\end{figure}

\begin{tcolorbox}[
    colback=gray!3,
    colframe=black!35,
    boxrule=0.5pt,
    arc=1mm,
    left=1.5mm,
    right=1.5mm,
    top=1mm,
    bottom=1mm
]
\small
\textbf{System:} You are a helpful assistant for geo-localization. Always use only the information contained in the provided representation to generate the description.

\vspace{1mm}

\textbf{User:} You are at location \texttt{<GEO>}. Generate a detailed description of the location.

\vspace{1mm}

\textbf{Assistant:} The distance to LAX airport, in mile, is
\end{tcolorbox}

We deliberately select seven query locations from data-sparse held-out regions containing no training observations. As shown in Figure~\ref{fig:lax_distance_examples}, these locations cover different directions and distances from LAX. Table~\ref{tab:lax_distance_examples} reports the numerical distance extracted from each model's completed response. Ground-truth values are calculated using the great-circle distance between each query location and LAX.

The examples reveal clear differences in how the representations respond to spatial variation. For the closest location, ID $0$, \textbf{LATENT} produces the closest estimate, while the other models substantially overestimate the distance. \textbf{LATENT} also produces the closest estimates for IDs $1$ and $6$, distinguishing locations near LAX from those substantially farther away. In contrast, \textbf{NAIVE} predictions remain within a relatively narrow range, while \textbf{GRID} produces nearly constant values between $10.5$ and $14.0$ miles, suggesting that it does not reliably capture continuous changes in distance. \textbf{TEXT} and \textbf{HYBRID} exhibit greater sensitivity to location, but their estimates remain less consistent.

Most predictions, particularly those from \textbf{TEXT}, \textbf{LATENT}, and \textbf{HYBRID}, overestimate the distance. One plausible explanation is a mismatch between the evaluation target and the geographic expressions in the training data. Our measured distance measures straight-line distance, whereas Airbnb descriptions often express approximate driving distances or travel times along road networks. The models may therefore inherit a systematic travel-distance bias from the training corpus. While these examples are qualitative, they illustrate that continuous spatial representations allow the generated values to respond more meaningfully to changes in geographic location. The smooth variation across space also indicates that they are learning a continuous spatial latent field to support their understanding of space.

\paragraph{Geospatial Grounding.}

\begin{figure*}[!t]
\centering
\begingroup
\setlength{\tabcolsep}{1.5pt}
\renewcommand{\arraystretch}{0}

\newcommand{\querylabel}[1]{%
    \begin{minipage}[c][0.115\textheight][c]{0.105\textwidth}
        \centering
        \small\bfseries #1
    \end{minipage}%
}

\newcommand{\groundingmap}[1]{%
    \begin{minipage}[c][0.115\textheight][c]{0.292\textwidth}
        \centering
        \includegraphics[
            width=\linewidth,
            height=0.112\textheight,
            keepaspectratio
        ]{#1}
    \end{minipage}%
}

\begin{tabular}{@{}cccc@{}}
&
\textbf{W/o Aug.}
&
\textbf{W/ Aug.}
&
\textbf{Difference}
\\[-1pt]

\querylabel{Near the\\ocean}
&
\groundingmap{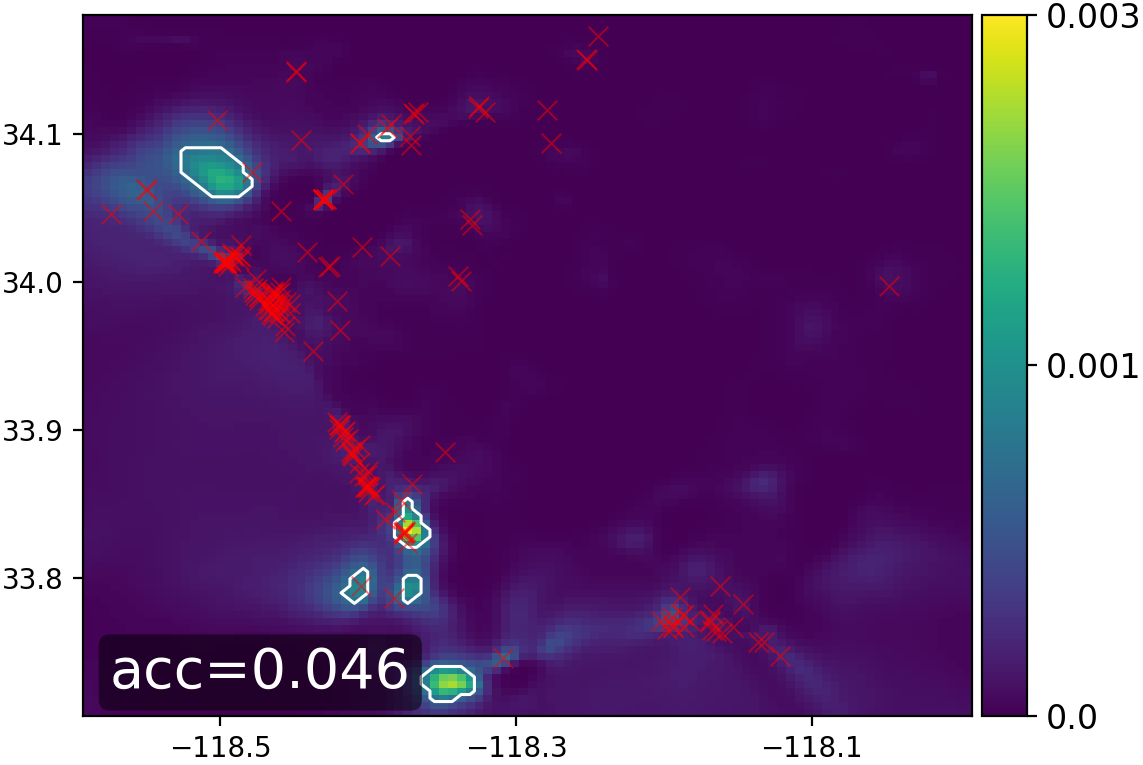}
&
\groundingmap{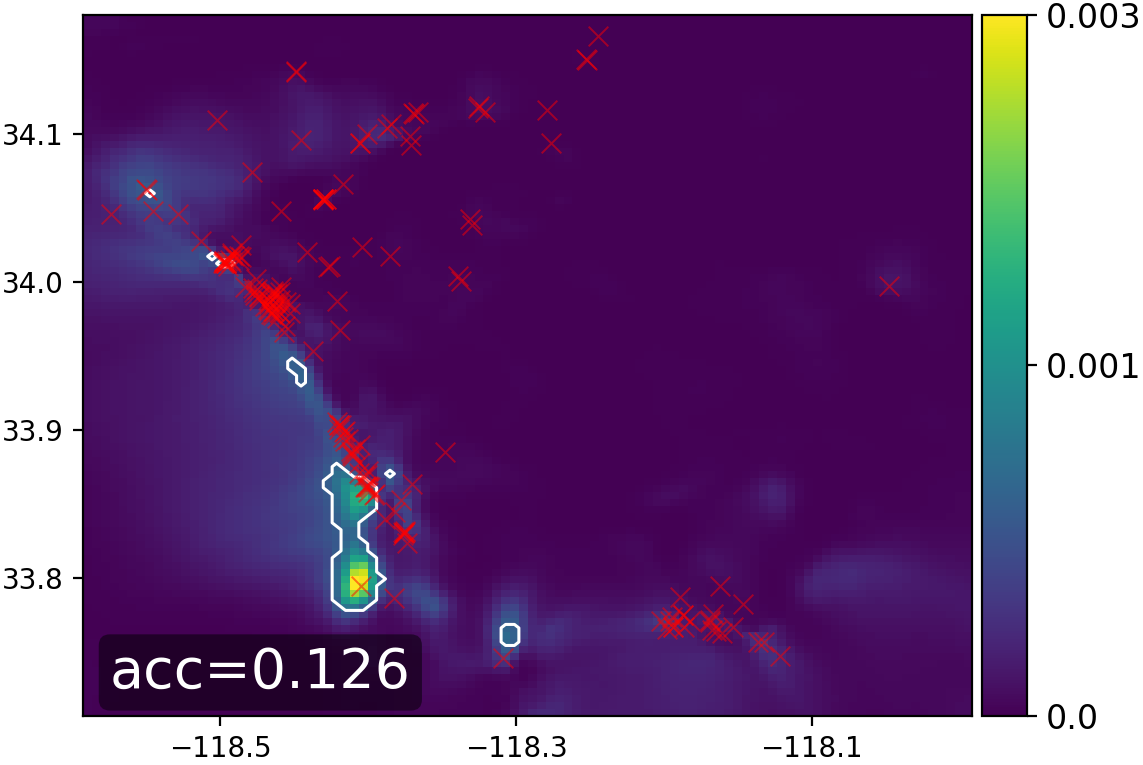}
&
\groundingmap{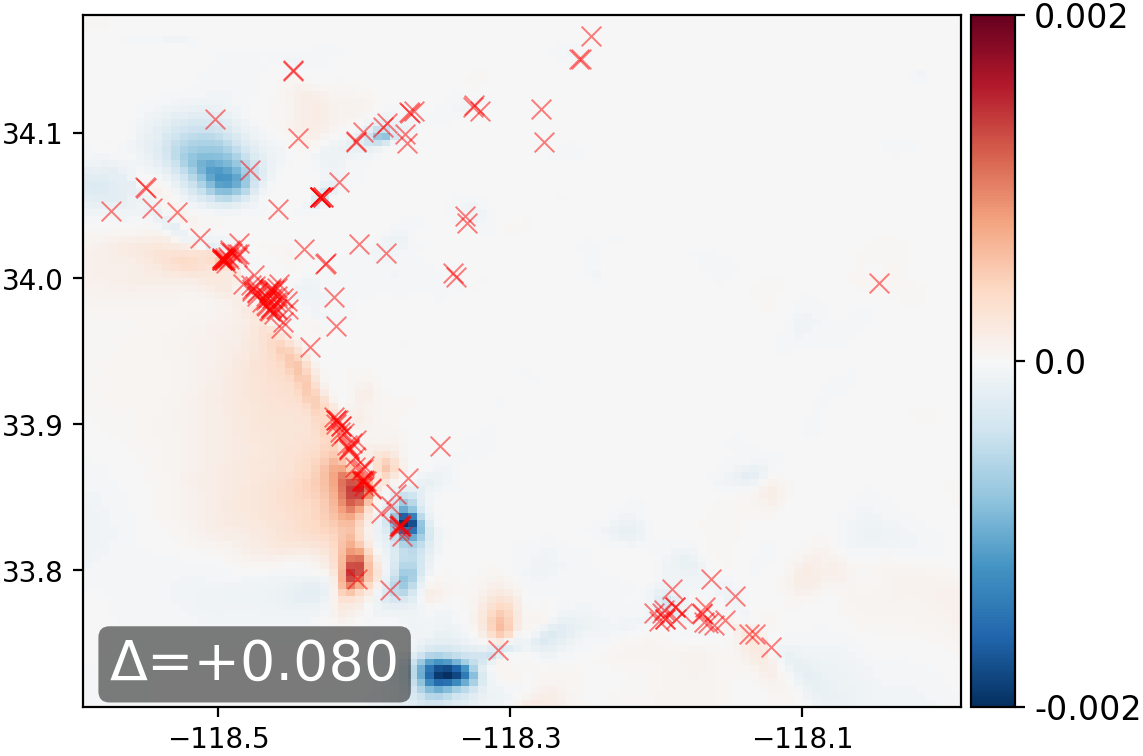}
\\[-1pt]

\querylabel{Close to\\Pasadena}
&
\groundingmap{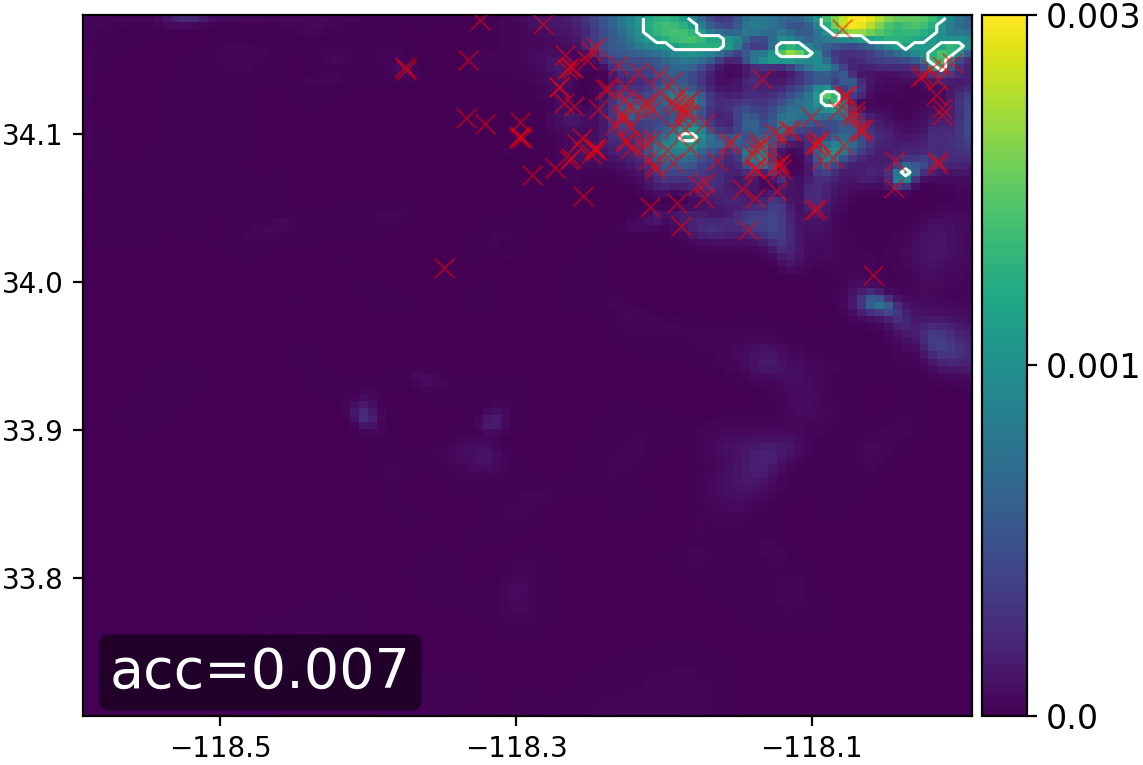}
&
\groundingmap{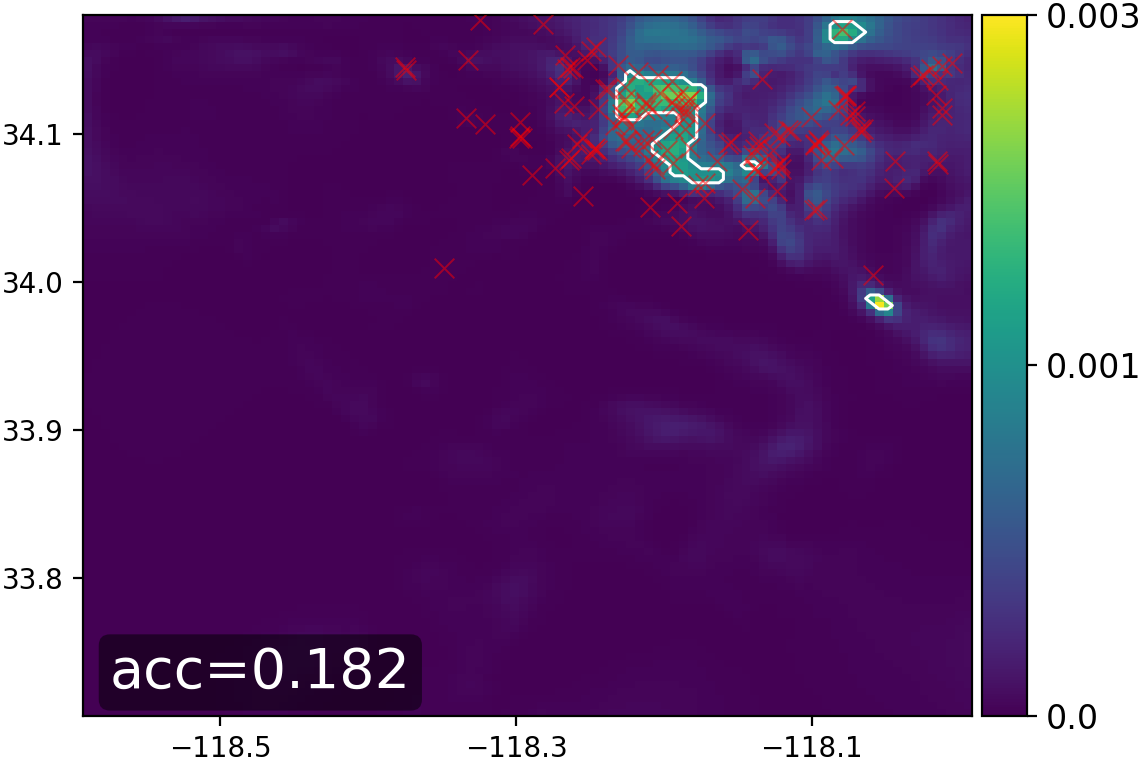}
&
\groundingmap{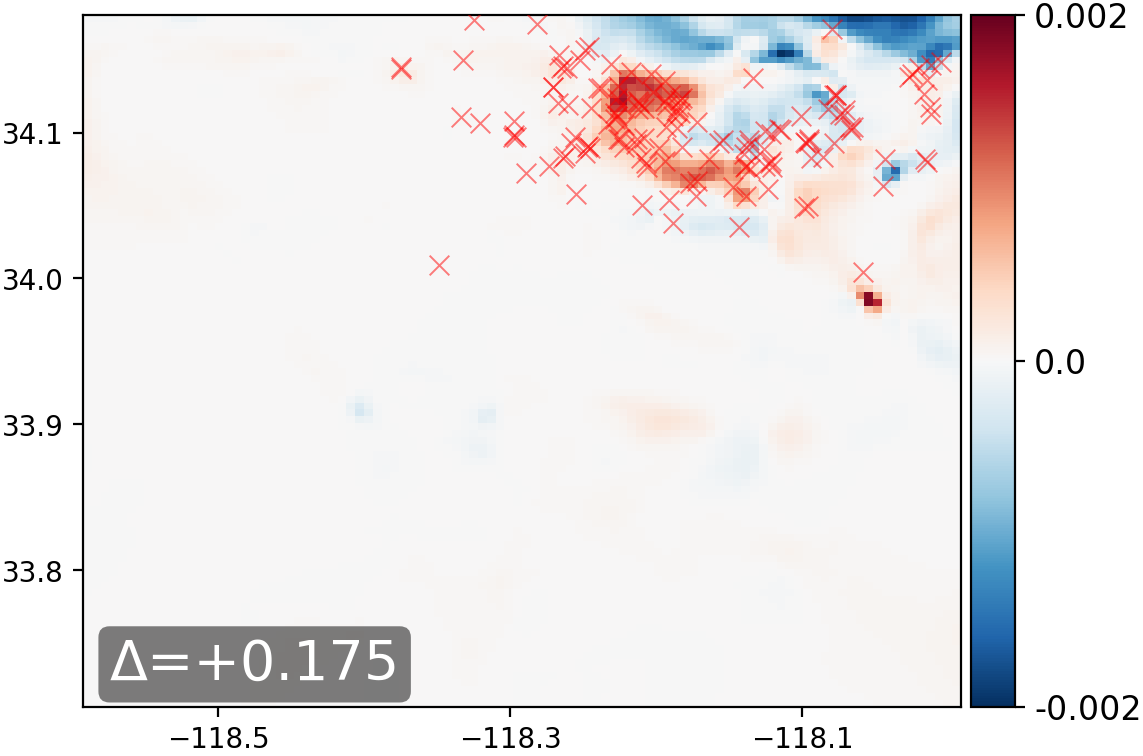}
\\[-1pt]

\querylabel{In DTLA}
&
\groundingmap{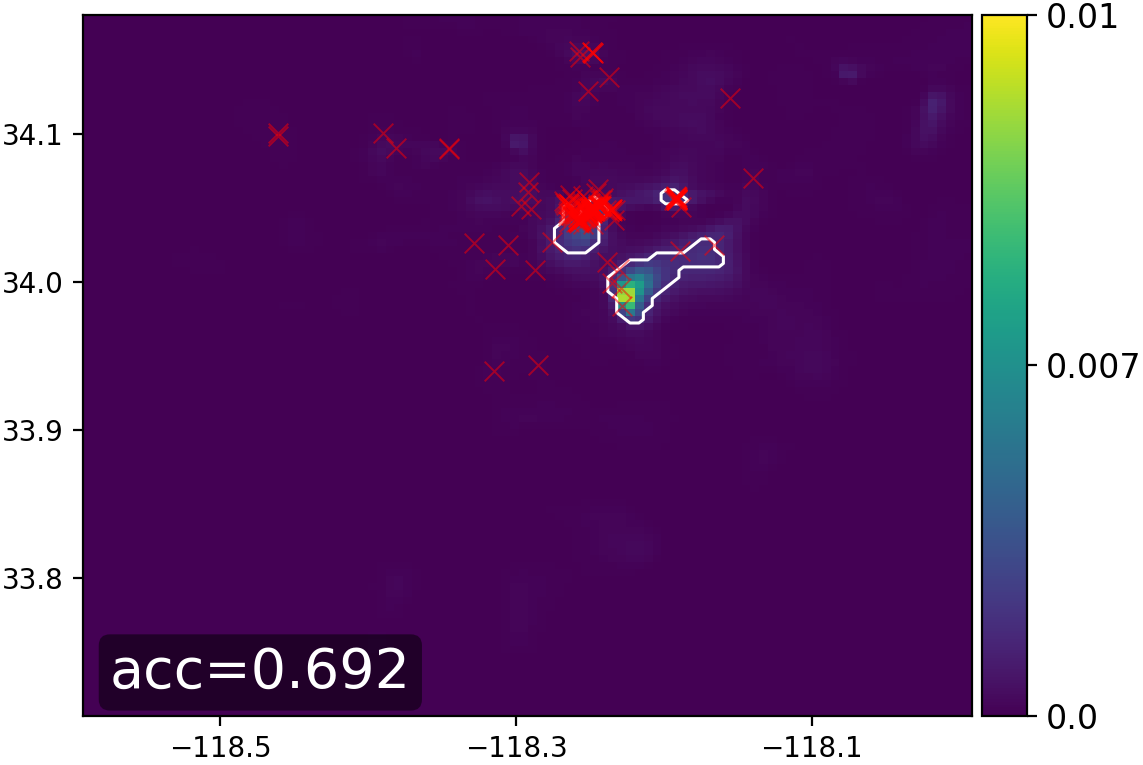}
&
\groundingmap{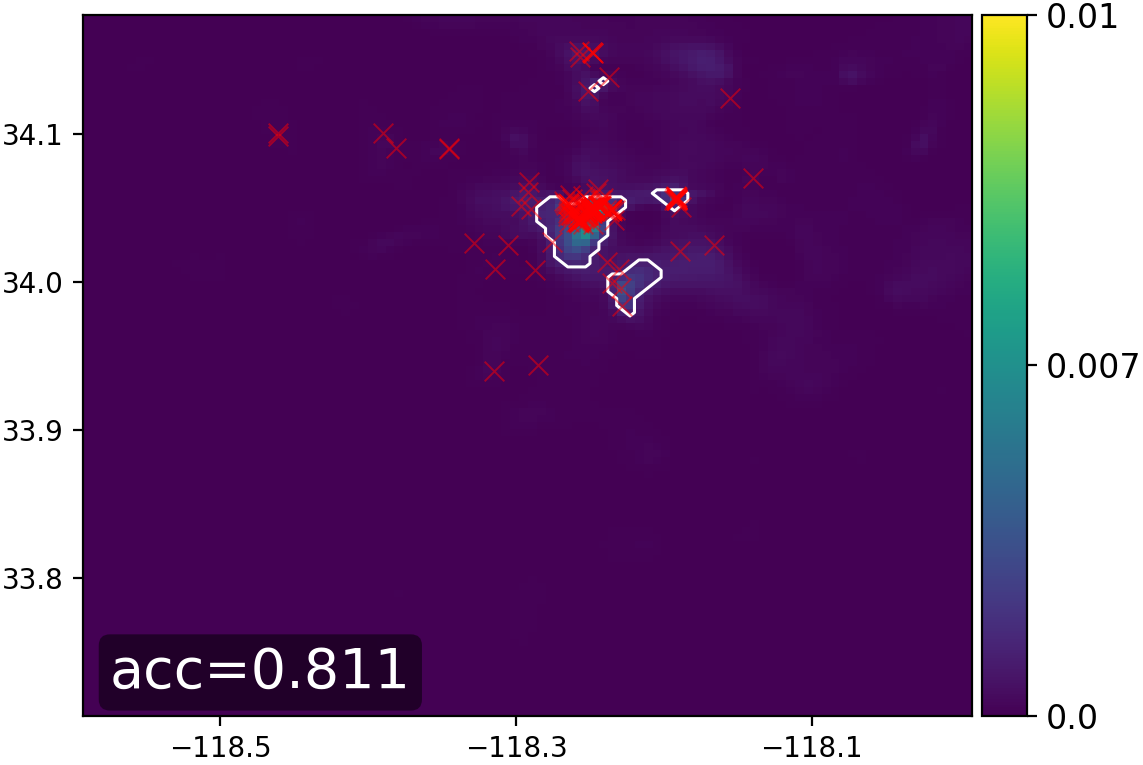}
&
\groundingmap{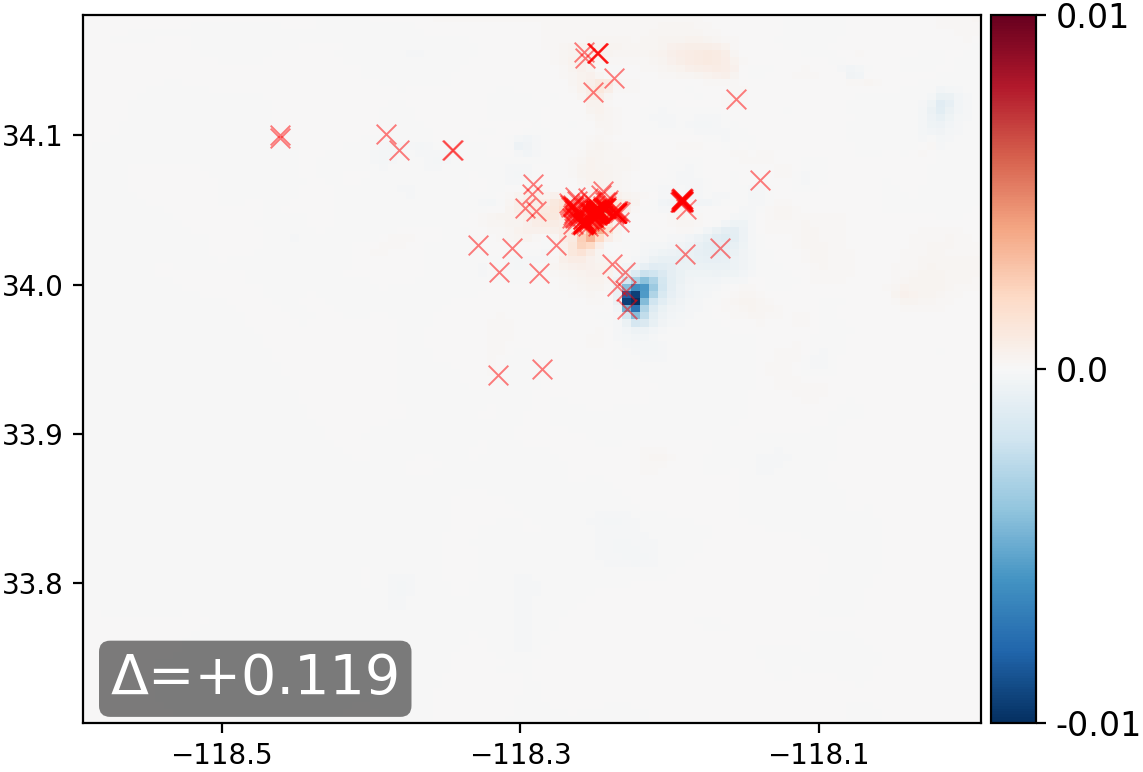}
\\

\end{tabular}
\endgroup

\caption{\textbf{Qualitative examples of geospatial grounding before and after caption augmentation.} Each row corresponds to a region-level grounding query, with red crosses indicating its ground-truth support locations. The first two columns show the predicted spatial similarity fields without and with gemma generated-caption augmentation, respectively, while white contours outline the highest-scoring $1\%$ of the geographic domain. The final column shows their difference, computed as the augmented field minus the baseline field; red indicates increased similarity and blue indicates decreased similarity.}
\label{fig:grounding_case_studies}
\end{figure*}

Figure~\ref{fig:grounding_case_studies} illustrates how caption augmentation changes the spatial field learned by the grounding model. For ``near the ocean,'' the baseline produces several disconnected high-similarity regions, including an incorrectly aligned inland region. After augmentation, the response becomes more continuous along the coastline while the incorrect inland mode is suppressed, increasing Top-$1\%$ coverage from $0.046$ to $0.126$. 

For ``close to Pasadena,'' the second row of Figure~\ref{fig:la_grounding_demos} shows the distribution of another semantically related query ``in Pasadena.'' The two similar expressions have distinct but spatially consistent ground-truth distributions: locations described as ``in Pasadena'' form a compact central region, while those described as ``close to Pasadena'' surround it. This pattern demonstrates that our grounding benchmark preserves fine-grained geographic relations between similar expressions. Without LLM-generated reasoning augmentation, the grounding model incorrectly aligns ``close to Pasadena'' with the central Pasadena region, effectively conflating the two phrases. Augmentation corrects this mismatch by shifting similarity toward the surrounding support locations, improving Top-$1\%$ coverage from $0.007$ to $0.182$.

The third example shows that augmentation can also refine an already meaningful spatial field. For ``in DTLA,'' the baseline identifies the general area but assigns substantial similarity to a displaced region south of the primary support. Augmentation suppresses this incorrect response and strengthens the spatially continuous region around the DTLA observations, increasing coverage from $0.692$ to $0.811$. The difference maps further show that augmentation does not simply increase similarity across the entire domain; instead, it redistributes similarity toward geographically consistent regions while reducing incorrectly aligned responses. These cases provide qualitative evidence that captions inferred at unobserved locations can fill gaps in the grounding data and promote a more continuous and accurately aligned spatial field.

\subsection{Statistic significance}
\label{appendix:statistic}

\paragraph{Significance testing.}
We test whether the grounding differences in Tables~3 and~4 are statistically significant.
Each setting, defined by an LLM family and a city, has $n=5$ grounding runs with different seeds.
For two augmentation methods $a$ and $b$ and a given metric, we average the difference of their
mean scores over the set of settings $\mathcal{S}$:
\begin{equation}
\widehat{\Delta} = \frac{1}{|\mathcal{S}|}\sum_{s\in\mathcal{S}}\left(\bar{x}_{a,s}-\bar{x}_{b,s}\right),
\qquad
\widehat{\mathrm{Var}}\big(\widehat{\Delta}\big) = \frac{1}{|\mathcal{S}|^{2}}\sum_{s\in\mathcal{S}}\frac{\sigma_{a,s}^{2}+\sigma_{b,s}^{2}}{n},
\end{equation}
where $\bar{x}$ and $\sigma^{2}$ are the mean and sample variance over seeds.
We apply a two-sided $t$-test to $\widehat{\Delta}$ with Welch--Satterthwaite degrees of freedom.
When the reference is \texttt{GeoGrounding} without augmentation, its five runs per city are shared
by all LLM families and are counted once per city.
We apply Holm correction over the five metrics within each comparison.
The seeds vary the grounding model only; each captioning model is fine-tuned once.

\begin{table}[t]
\centering
\caption{\textbf{Significance tests for geospatial grounding.}
Each cell reports the mean difference with its two-sided $p$-value in parentheses.
Differences are averaged over the three LLM families and both cities, which are treated as strata
(stratified $t$-test with Welch--Satterthwaite degrees of freedom; five grounding seeds per setting).
The first five rows compare each augmentation with \texttt{GeoGrounding} trained without augmentation;
the last row compares \textbf{LATENT} with \textbf{TEXT} directly.
\textbf{Bold} marks a significant improvement and $^{\dagger}$ a significant degradation
after Holm correction over the five metrics (adjusted $p<0.05$).
$\uparrow$/$\downarrow$ indicate whether higher or lower values are better.}
\label{tab:significance}
\footnotesize
\setlength{\tabcolsep}{3pt}
\resizebox{\textwidth}{!}{%
\begin{tabular}{lccccc}
\toprule
Comparison & Hit Rate@1\% $\uparrow$ & Hit Rate@5\% $\uparrow$ & Coverage@1\% $\uparrow$ & JSD $\downarrow$ & Wasserstein $\downarrow$ \\
\midrule
\multicolumn{6}{l}{\textit{Augmentation vs.\ no augmentation}} \\
w/ GRID   & $-2.68$ ($<$0.001)$^{\dagger}$ & $-2.29$ ($<$0.001)$^{\dagger}$ & $-1.77$ ($<$0.001)$^{\dagger}$ & $+0.0156$ ($<$0.001)$^{\dagger}$ & $+312.2$ (0.009)$^{\dagger}$ \\
w/ HASH   & $+0.12$ (0.487) & $-0.07$ (0.774) & $-0.10$ (0.668) & $-0.0013$ (0.572) & $-135.2$ (0.173) \\
w/ TEXT   & $+0.20$ (0.243) & $-0.09$ (0.686) & $-0.09$ (0.702) & $-0.0012$ (0.588) & $-147.0$ (0.141) \\
w/ HYBRID & $+0.23$ (0.180) & $-0.12$ (0.608) & $-0.52$ (0.055) & $-0.0048$ (0.062) & $-72.0$ (0.442) \\
w/ LATENT & {\bfseries\boldmath $+0.63$ (0.002)} & $+0.27$ (0.235) & $-0.13$ (0.574) & {\bfseries\boldmath $-0.0081$ (0.005)} & $-164.4$ (0.103) \\
\midrule
\multicolumn{6}{l}{\textit{Direct comparison}} \\
LATENT vs.\ TEXT & {\bfseries\boldmath $+0.43$ ($<$0.001)} & {\bfseries\boldmath $+0.37$ (0.006)} & $-0.05$ (0.757) & {\bfseries\boldmath $-0.0069$ ($<$0.001)} & $-17.3$ (0.730) \\
\bottomrule
\end{tabular}}
\end{table}

\paragraph{Results.}
Table~\ref{tab:significance} shows that \textbf{LATENT} is the only augmentation that significantly
improves grounding over no augmentation, on Hit Rate@1\% ($+0.63$, $p=0.002$) and JSD ($-0.0081$, $p=0.005$).
\textbf{GRID} significantly degrades all five metrics, and \textbf{HASH}, \textbf{TEXT} and
\textbf{HYBRID} show no significant change.
Compared directly with \textbf{TEXT}, \textbf{LATENT} is significantly better on Hit Rate@1\%,
Hit Rate@5\% and JSD.
On the remaining metrics the differences are not significant, but \textbf{LATENT} still compares
favorably with the other augmentations: it is the only one with a positive mean difference on
Hit Rate@5\% ($+0.27$, $p=0.235$), and it has the largest reduction and the smallest $p$-value on
Wasserstein distance ($-164.4$, $p=0.103$).
Coverage@1\% shows no reliable difference for any augmentation other than \textbf{GRID}.

\subsection{Location Encoder}
\label{appendix:location_encoder}

The location encoder is not pretrained. The encoding has a closed form, following the definition in Geo2Vec~\citep{geo2vec}, and contains no learned parameters. It therefore captures only the geometry of a location and carries no geographic knowledge. The projection that maps the encoding into the LLM embedding space is trained from scratch together with the LoRA adapters.

To separate what the encoding provides from what the LLM learns, we cluster the feature from the location encoding alone, using the same procedure as in Appendix~\ref{appendix:cluster}. Figure~\ref{fig:encoding_clusters} shows the result for both cities.

\begin{figure}[h]
    \centering
    \begin{subfigure}[b]{0.4\textwidth}
        \centering
        \includegraphics[width=\linewidth]{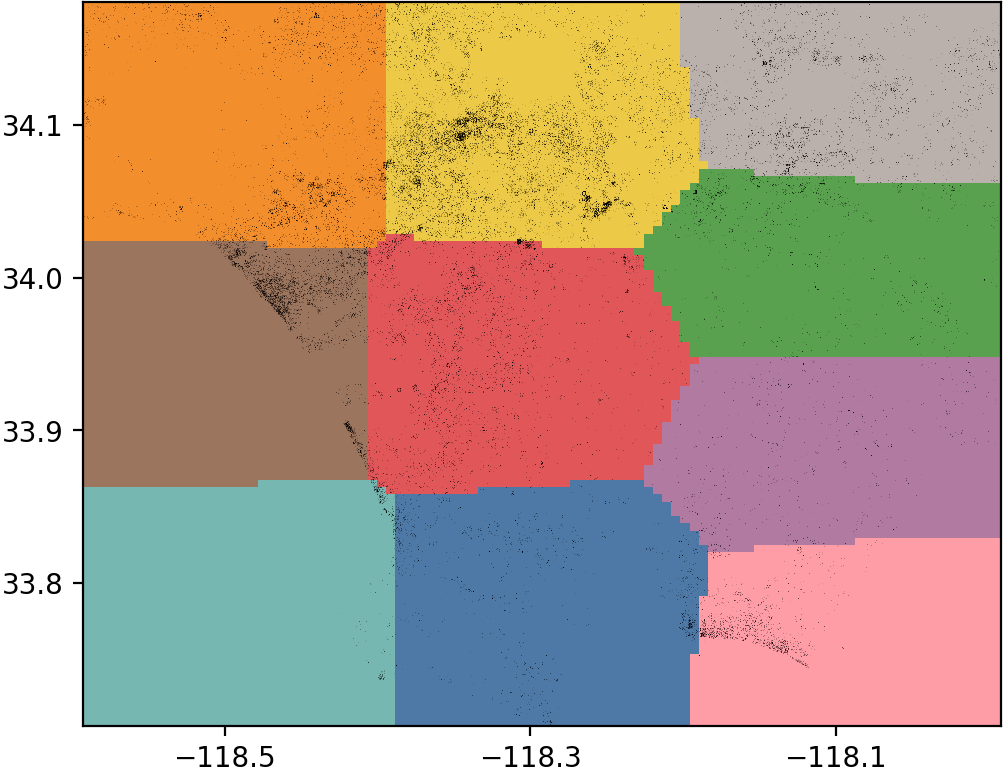}
        \caption{Los Angeles}
        \label{fig:encoding_clusters_la}
    \end{subfigure}
    \hfill
    \begin{subfigure}[b]{0.48\textwidth}
        \centering
        \includegraphics[width=\linewidth]{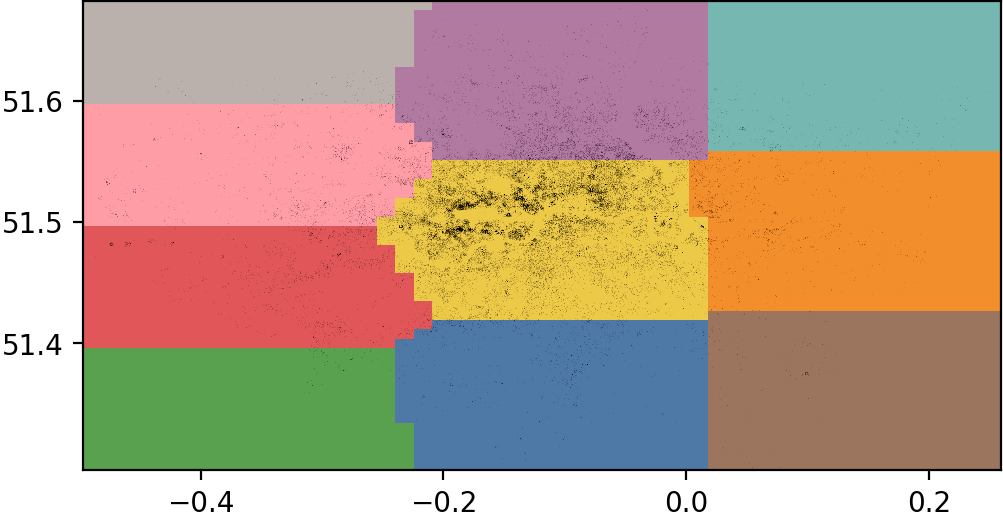}
        \caption{London}
        \label{fig:encoding_clusters_london}
    \end{subfigure}
    \caption{\textbf{Cluster maps of the location encoding alone.}}
    \label{fig:encoding_clusters}
\end{figure}

The encoding alone preserves only spatial proximity: its clusters divide each city into regular regions that do not follow the underlying geography. After the LLM learns from the data, the clusters of its hidden representations reflect the data. For example, in the \textbf{LATENT} maps of Figure~\ref{fig:la_cluster_maps_appendix}, the coastal areas around Santa Monica and Long Beach are grouped together, following the coastline. Geometry alone cannot produce this grouping, so this structure is learned from the descriptions and does not originate from the encoder.

\end{document}